\documentclass{article}

\usepackage{arxiv}
\usepackage{booktabs}

\date{}

\usepackage[utf8]{inputenc} 
\usepackage[T1]{fontenc}    
\usepackage{hyperref}
\usepackage{natbib}
\setcitestyle{authoryear}
\setcitestyle{aysep={}} 
\usepackage{rotating}
\usepackage{multirow}
\usepackage{lscape}
\usepackage{longtable}
\usepackage[flushleft]{threeparttable}
\usepackage{color, colortbl}
\usepackage[labelsep=space]{caption} 
\usepackage{newtxmath}
\definecolor{DarkGray}{rgb}{0.655, 0.655, 0.655}
\definecolor{LightGray}{rgb}{0.847, 0.847, 0.847}
\usepackage{pifont}
\newcommand{\cmark}{\ding{51}}%

\def\sidewaystablefn{\renewcommand\footnotetext[2][]{{\removelastskip\vskip3pt%
\let\tablebodyfont\tablefootnotefont%
\hskip0pt\if!##1!\else{\smash{$^{##1}$}}\fi##2\par}}%
}%

\title{How to keep text private? A systematic review of deep learning methods for privacy-preserving natural language processing}
\renewcommand{\shorttitle}{How to keep text private?}
\hypersetup{
pdftitle={How to keep text private? A systematic review of deep learning methods for privacy-preserving natural language processing},
pdfsubject={cs.AI, cs.CL, cs.CR, cs.LG, cs.GL, cs.NE},
pdfauthor={Samuel Sousa, Roman Kern},
pdfkeywords={Deep learning, Privacy, Natural language processing, Differential privacy, Homomorphic encryption, Searchable encryption, Federated learning},
hidelinks,
}

\author{ \href{https://orcid.org/0000-0001-7196-9095}{\includegraphics[scale=0.06]{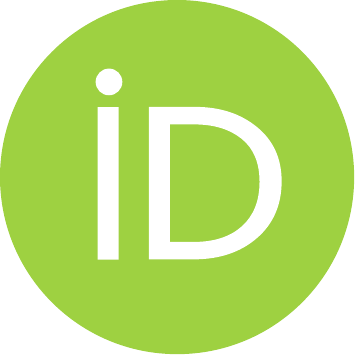}\hspace{1mm}Samuel Sousa}\\ 
	Know-Center GmbH \& \\
	Institute of Interactive Systems and Data Science\\
	Graz University of Technology\\
	Graz, Austria \\
	\texttt{ssousa@know-center.at} \\
	\And
	\href{https://orcid.org/0000-0003-0202-6100}{\includegraphics[scale=0.06]{orcid.pdf}\hspace{1mm}Roman Kern} \\
	Know-Center GmbH \& \\
	Institute of Interactive Systems and Data Science\\
	Graz University of Technology\\
	Graz, Austria \\
	\texttt{rkern@know-center.at} \\
}

\begin{document}
\maketitle

\begin{abstract}
Deep learning (DL) models for natural language processing (NLP) tasks often handle private data, demanding protection against breaches and disclosures.
Data protection laws, such as the European Union's General Data Protection Regulation (GDPR), thereby enforce the need for privacy. 
Although many privacy-preserving NLP methods have been proposed in recent years, no categories to organize them have been introduced yet, making it hard to follow the progress of the literature. 
To close this gap, this article systematically reviews over sixty DL methods for privacy-preserving NLP published between 2016 and 2020, covering theoretical foundations, privacy-enhancing technologies, and analysis of their suitability for real-world scenarios.
First, we introduce a novel taxonomy for classifying the existing methods into three categories: data safeguarding methods, trusted methods, and verification methods.
Second, we present an extensive summary of privacy threats, datasets for applications, and metrics for privacy evaluation.
Third, throughout the review, we describe privacy issues in the NLP pipeline in a holistic view.
Further, we discuss open challenges in privacy-preserving NLP regarding data traceability, computation overhead, dataset size, the prevalence of human biases in embeddings, and the privacy-utility tradeoff.
Finally, this review presents future research directions to guide successive research and development of privacy-preserving NLP models.
\end{abstract}



\section{Introduction}\label{sec:introduction}

Privacy is the ability to control the extent of personal matters an individual wants to reveal~\citep{westin1968privacy}. 
It is protected by regulations in many countries around the planet, like the European Union (EU)'s General Data Protection Regulation (GDPR)~\citep{EUdataregulations2018}, in order to protect the security, integrity, and freedom of people. 
For instance, the GDPR establishes guidelines for the collection, transfer, storage, management, processing, and deletion of personal data within the EU. 
Penalties and fines are also applicable in case of misbehavior or non-compliance with legal terms. 
The approval of data protection laws has driven an increased need for privacy-enhancing technologies (PETs), which control the amount of existing personal data, limiting it or ridding it, as well as its processing, without losing system functionalities~\citep{van2003handbook}.
Therefore, PETs constitute the cornerstone for the privacy preservation of personal data.

\begin{figure}
    \includegraphics[width=\textwidth]{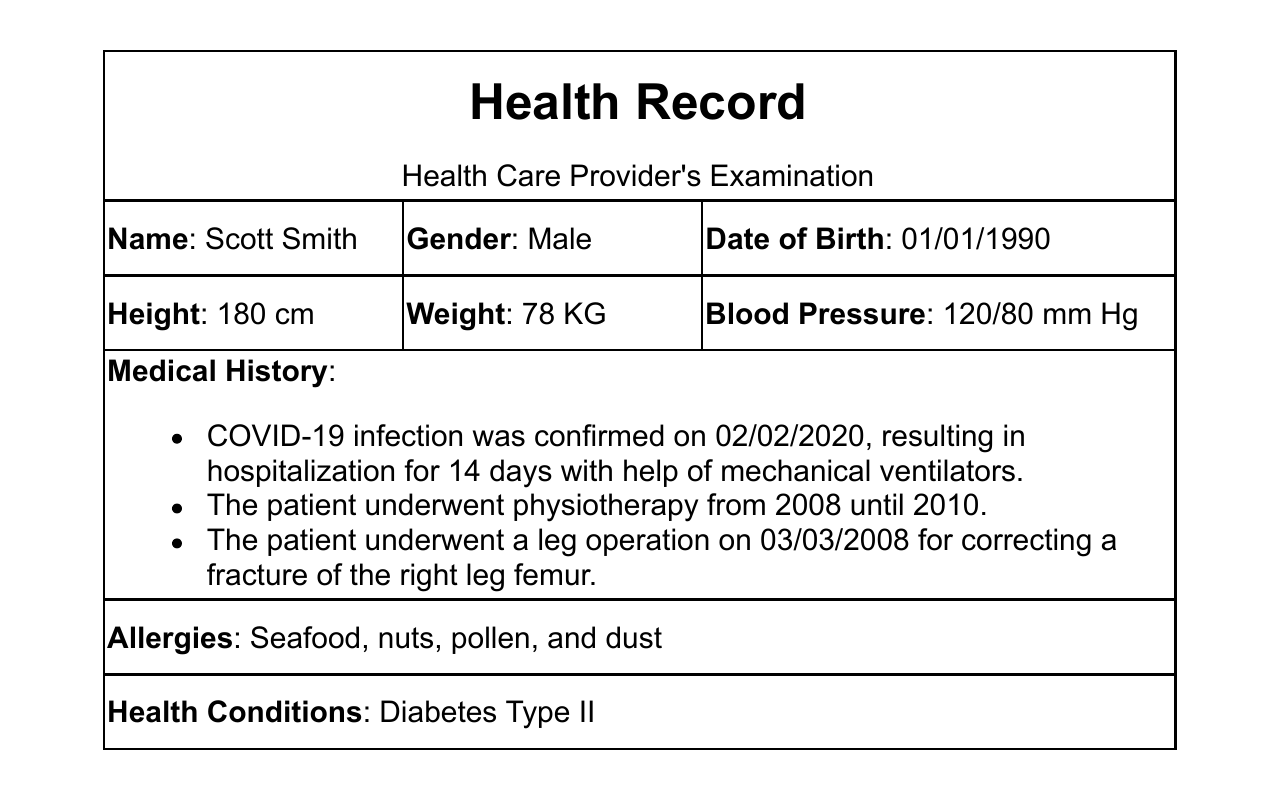}
    \caption{Pieces of private textual information in a health record}
    \label{fig:private-information-in-text}
\end{figure}

Data from several domains, such as finance~\citep{han2019logistic}, documents~\citep{eder2019identification}, bio-medicine~\citep{dernoncourt2017identification}, social media~\citep{blodgett2017racial,salminen2020enriching}, and images~\citep{he2019model,sanchez2018automatic}, inherently present sensitive content which must be protected. 
Such sensitive attributes include a person's identity, age, gender, home address, location, income, marital status, ethnicity, political and societal views, health conditions, pictures, as well as any other traits that allow their identification or have the potential to harm their safety or reputation~\citep{alawad2020privacy}. 
In text data, private information can be found in many features derived from the text content in documents, e-mails, chats, online comments, medical records, and social media platforms.
Figure~\ref{fig:private-information-in-text} depicts some pieces of private information in text data as the fields of a health record.
These fields contain demographic attributes (e.g., gender and age), physical characteristics (e.g., height, weight, and blood pressure), history of medical treatments, allergies, a person's full name, and health conditions.
A model for predicting diseases from such a health record must not reveal the identity and health information of the hospital patient who generated it.

The preservation of privacy is a bottom line for developing new deep learning (DL) methods in scenarios that feature sensitive or personal data alongside threats of breaches. 
In addition, deep neural network models are often a target for attacks that aim at recovering data instances used for training, especially reverse engineering~\citep{li2018towards}, membership inference~\citep{pan2020privacy}, and property inference~\citep{ganju2018property}.
PETs~\citep{van2003handbook}, which are computational techniques to manage privacy risks, emerged thereon, covering all phases of the model pipeline, from data pre-processing up to the validation of results.
In the past few years, PETs have drawn attention in the literature since DL architectures have often been applied to private data combined with them~\citep{boulemtafes2020review}. 
Well-known PETs for DL include fully homomorphic encryption (FHE)~\citep{gentry2009fully}, differential privacy (DP)~\citep{dwork2008differential}, adversarial learning~\citep{goodfellow2014generative}, federated learning (FL)~\citep{yin2020fdc}, multi-party computation (MPC)~\citep{goldreich1998secure}, and secure components~\citep{jia2013human}. 
Although these technologies protect privacy from disclosures and attacks, they come with utility-efficiency tradeoffs. 
For instance, adding noise to a model's training data often degrades its performance.
Running DL models on encrypted data can also be memory-demanding and time-consuming.

Recent advances in the field of natural language processing (NLP) have been provided by DL methods that steadily led to outstanding results for tasks, such as automatic question-answering~\citep{minaee2017automatic}, dependency parsing~\citep{kiperwasser2016simple}, machine translation~\citep{vaswani2018tensor2tensor}, natural language understanding~\citep{sarikaya2014application}, text classification~\citep{liu2017deep}, and word sense disambiguation~\citep{samuel2020wsdijcnn}. 
However, DL models trained or used for inference on sensitive text data may be susceptible to privacy threats where the learning setting features more than a single computation party or the data is outsourced~\citep{boulemtafes2020review}. 
Therefore, privacy is an essential requirement for developing NLP applications for personal and private data.

Preserving the privacy of text data is a tough challenge due to the hardness of identifying sensitive attributes in text alongside the computational costs PETs may demand. 
In some cases, removing private attributes does not hinder their inference from their surrounding context.
For instance, location names can be placed near descriptions that allow their identification.
As a response to a large number of privacy-related issues in NLP, a broad separate literature has developed as an active research topic.
Despite its broadness, the literature on privacy-preserving NLP does not present a precise categorization of approaches regarding DL and PETs to provide an overview of the main topics and technologies to both the scientific community and industry practitioners. 
Table~\ref{tab:abbreviations} lists all the abbreviations we make use of throughout this review.

\begin{table}[h!]
    \begin{center}
    \begin{minipage}{\textwidth}
    \caption{Abbreviations used throughout this review}
    \label{tab:abbreviations}
    \centering
    \begin{tabular}{lc}
    \toprule
        \textbf{Abbreviation} & \textbf{Description}\\
    \toprule
    AA & Authorship attribution \\
    AI & Artificial intelligence \\
    BiLSTM & Bidirectional long short-term memory \\
    CNN & Convolutional neural network\\
    DA & Domain adaptation \\
    DL & Deep learning \\
    DP & Differential privacy \\
    EU & European Union \\
    FHE & Fully-homomorphic encryption \\
    FL & Federated learning \\
    GDPR & General Data Protection Regulation \\
    HIPAA & Health Insurance Portability and Accountability Act \\
    LSTM & Long short-term memory \\
    MPC & Multi-party computation \\
    NLP & Natural language processing \\
    PETs & Privacy-enhancing technologies \\
    PHI & Protected health information \\
    RNN & Recurrent neural network\\
    SE & Searchable encryption \\
    SEAT & Sentence Encoder Association Test \\
    TL & Transfer learning \\
    WEAT & Word Embedding Association Test \\
    
    \bottomrule
    \end{tabular}
    \end{minipage}
    \end{center}
\end{table}

\subsection{Related surveys and reviews}\label{sub:related-works}

\begin{table}[ht!]
    \begin{minipage}{\textwidth}
    \caption{Overview of related surveys on privacy for DL and NLP}
    \label{tab:summary-related-surveys-and-reviews}
    \centering
    \begin{tabular}{ccc}
    \toprule
        \textbf{Work} & \textbf{Year} & \textbf{Contributions}\\
    \toprule
     \parbox[t]{0.375\textwidth}{Technical privacy metrics: A systematic survey~\citep{wagner2018technical}.} & 2018 & \parbox[t]{0.45\textwidth}{Summary and classification of over 80 privacy metrics, theoretical foundations, application domains, examples in context, and a methodology to determine suitable metrics for application scenarios.}\\
     {} \\
    \parbox[t]{0.38\textwidth}{Mitigating gender bias in natural language processing: Literature review~\citep{sun2019mitigating}.} & 2019 & \parbox[t]{0.45\textwidth}{Literature review of recent works on recognition and mitigation of algorithmic gender bias in the NLP field and discussion aiming to point out drawbacks and gaps for future research.}\\
     {} \\
    \parbox[t]{0.38\textwidth}{Text analysis in adversarial settings: Does deception leave a stylistic trace?~\citep{grondahl2019text}.} & 2019 & \parbox[t]{0.45\textwidth}{Study of changes in authors' writing styles to detect deceptive text, analysis of whether authorship identification techniques encompass privacy threats, description of backdrops for attacks, the outline of style obfuscation and imitation methods, and discussion on the traceability of style obfuscation techniques.}\\
     {} \\
    \parbox[t]{0.38\textwidth}{A survey on interdependent privacy~\citep{humbert2019survey}.} & 2019 & \parbox[t]{0.45\textwidth}{Insights on privacy and interdependent privacy, highlighting how different scientific communities approach these topics; review of tools and theoretical frameworks for research; and the categorization of risks, concerns, and solutions.} \\
     {} \\
    \parbox[t]{0.38\textwidth}{A review of privacy-preserving techniques for deep learning~\citep{boulemtafes2020review}.} & 2020 & \parbox[t]{0.45\textwidth}{A multi-level taxonomy of privacy-preserving techniques for DL, followed by their performance analysis.} \\
     {} \\
    \parbox[t]{0.38\textwidth}{The AI-based cyber threat landscape: A survey~\citep{kaloudi2020ai}.} & 2020 & \parbox[t]{0.45\textwidth}{The map of cyber-attacks adopting AI techniques onto a framework for the classification of dishonest AI uses. This work also presents a basis for their detection in future events. An attack backdrop on a smart grid infrastructure is also included to illustrate how to apply the proposed framework.} \\
     {} \\
    \parbox[t]{0.38\textwidth}{Adversarial attacks on deep-learning models in natural language processing: A survey~\citep{zhang2020adversarial}.} & 2020 & \parbox[t]{0.45\textwidth}{The outline, discussion, and taxonomy of works on generating adversarial examples for NLP applications based on DL models.}\\
    \bottomrule
    \end{tabular}
    \end{minipage}
\end{table}

Privacy preservation has multiple perspectives involving privacy metrics, DL models, NLP tasks, threats, computation scenarios, and PETs.
This work lies at the intersection of these perspectives, aiming to bridge the gap between them.
For this reason, we review the literature on privacy, DL, and NLP from 2016 to 2020, to provide an extensive review of the privacy-preserving DL methods for NLP and draw attention to the privacy issues NLP data may face. 
Some recently published survey papers have brought overviews on privacy metrics, privacy-preserving DL, security attacks, security issues, biases in NLP, privacy on social media, and adversarial threats in text data, as Table~\ref{tab:summary-related-surveys-and-reviews} shows.
Each work in the table introduces different perspectives on privacy preservation, yet apart from a broad unified review for NLP applications based on DL models.
Moreover, the papers we review rarely overlap those reviewed by the works in Table~\ref{tab:summary-related-surveys-and-reviews}; hence, highlighting the comprehensive coverage of our review. 

A primary problem for privacy regards its measurement. 
Many metrics have been proposed, so \cite{wagner2018technical} review more than 80 metrics for measuring privacy, introducing a novel classification determined by four determinants. 
First, the capabilities that an adversary model is likely to present, such as information about a probability distribution. 
Second, the data sources, especially public and private data. 
Third, the expected inputs, e.g., parameters. 
Finally, properties privacy metrics measure, including data similarity, error, information gain, or information loss.  
A procedure for choosing the most suitable privacy measures for different backdrops is also provided, followed by the advice on choosing more than a single metric to cover a larger number of privacy aspects. 
In the end, the authors point out the need for further research on privacy metrics aggregation concerning cases in which personal privacy is affected by other parties. 

\cite{humbert2019survey} survey risks and solutions for interdependent privacy, proposing categories for both.
For example, when users of different e-mail providers engage in message exchange, the history of such communication is stored on the servers of both e-mail providers. 
In this case, the privacy of an e-mail user depends on the other's actions, like leaking the exchanged messages or keeping them secret.
The risks are sorted in the survey according to the data type they arise from, such as demographic, genomic, multimedia, location, and aggregated data.  
Similarly, the solutions for interdependent privacy are split into two groups based on their complexity. 
On the one hand, simple manual user actions, like limiting data visibility for other users, compose the group of non-technical solutions (also referred to as social solutions). 
On the other hand, solutions that rely on computational tools, software architectures, encryption, and similar are grouped into technical solutions. 
The authors also argue that research approaches for interdependent privacy should lean mostly on principles rather than data dependency.

In DL, a total of 45 privacy-preserving solutions are reviewed by \cite{boulemtafes2020review} and then organized as a four-level taxonomy. 
The first level regards privacy-preserving tasks, namely model learning, analysis, and model releasing.
Level 2 of the taxonomy refers to the learning paradigm of the reviewed techniques, like individual or collaborative learning. 
The third level draws up differences between server-based and server-assisted approaches. 
Last, the fourth taxonomy level classified privacy-preserving methods based on technological concepts, such as encryption or MPC.
\cite{kaloudi2020ai} also cover DL architectures in their survey of cyber-attacks based on artificial intelligence (AI) techniques, such as password crackers which use self-learning for brute-force attacks. 
However, the authors focus on malicious uses of such neural networks, presenting a classification of attacking threats as a future prediction framework. 
Consequently, this new framework can be helpful for implementing new safeguarding measures against the identified risk scenarios.

NLP data and tasks also encompass privacy-related issues requiring special algorithmic solutions. 
\cite{sun2019mitigating} review studies on identifying and mitigating gender bias in text data tasks. 
The authors propose four categories of representation bias, relying on problems noticed in the datasets, such as derogatory terms, societal stereotypes, and under-representation of some groups, and models, like their inaccuracies for sensitive tasks.
Bias arising from text data can easily be embedded into vector representations of words and consecutively propagated to downstream tasks so that bias removal methods are demanded. 
The conclusions of this review are four-fold. 
First, it is observed that debiasing methods are usually implemented as end-to-end solutions whose interactions between their parts are still unknown. 
Second, debiasing methods' generalization power is questioned since they are mostly tested on narrow NLP tasks. 
Further, the authors have argued that some of these methods may introduce noise into NLP models, leading to drops in performance. 
Last, it is noticed that hand-crafted debiasing techniques may be inadvertently biased by their developers. 

An outstanding security issue in NLP regards deceptive language~\citep{MihalceaS09}. \cite{grondahl2019text} provide an overview of empirical works on the detection of deceptive text.
The authors first target misleading content, i.e., detecting dishonest text based on the author's writing style.
Second, they focus on adversarial stylometry, which is considered a type of deceptive text since truthful information pieces in the data were replaced with anonymized versions.
It consists of a method against deanonymization attacks. 
Finally, they conclude that stylometry analysis is efficient in predicting deceptive text when the training and test domains are adequately related.
However, deanonymization attacks will continue to be a rampant and growing privacy threat.
As a result, manual style obfuscation approaches are expected to solve more efficiently than automatic ones. 

Beyond deanonymization attacks, text data can also be used for adversarial attacks against DL models. In these attacks, data samples are generated with small perturbations but can dupe DL models and prompt false predictions.
\cite{zhang2020adversarial} overview 40 works that addressed adversarial attacks towards DL models for NLP tasks, such as text classification, machine translation, and text summarization. 
To split the works into categories in a taxonomy, the authors consider five viewpoints in both model and semantic perspectives. 
First, knowledge of the attacked model at the time of the attack.
Second, NLP applications.
Third, the target for the attack, e.g., incorrect predictions or specific results.
Further, the granularity level of the attack which ranges from character level to sentence level. 
Finally, the attacked DL model, such as convolutional neural network (CNN), recurrent neural network (RNN), and autoencoders.
State-of-the-art methods, general NLP tasks, defenses, and open challenges are also discussed.
The authors then raise the issue that adversarial examples can be used for membership inference attacks, which reconstruct the original data samples used for training DL models based on perturbed inputs.
Therefore, safeguarding models against these attacks is still an open challenge.

\begin{table}[ht!]
    \begin{center}
    \begin{minipage}{\textwidth}
    \caption{Comparison of this review against related surveys and their content coverage}
    \label{tab:comparison-related-surveys}
    \centering
    \begin{tabular}{cccccccc}
    \toprule
        \textbf{Work} & \textbf{Year} & \textbf{Threats} & \textbf{Solutions} & \textbf{DL} & 
        \textbf{NLP tasks} & 
        \textbf{Metrics} \\
    \toprule
    \begin{tabular}{p{0.17\textwidth}}\cite{wagner2018technical} \end{tabular} & 2018 & \cmark & \cmark & {} & {} & \cmark \\
    \begin{tabular}{p{0.17\textwidth}} \cite{sun2019mitigating} \end{tabular} & 2019 & \cmark & \cmark & \cmark & Specific & {} \\
    \begin{tabular}{p{0.17\textwidth}} \cite{grondahl2019text}  \end{tabular} & 2019 & \cmark & \cmark & \cmark & General & {} \\
    \begin{tabular}{p{0.17\textwidth}} \cite{humbert2019survey} \end{tabular} & 2019 & \cmark & \cmark & \cmark & {} & {}\\
     \begin{tabular}{p{0.17\textwidth}}
           \cite{boulemtafes2020review}\\
     \end{tabular} & 2020 & \cmark & \cmark & \cmark & {} & \cmark\\
    \begin{tabular}{p{0.17\textwidth}} \cite{kaloudi2020ai} \end{tabular} & 2020 & \cmark & {} & \cmark & General & {} \\
    \begin{tabular}{p{0.17\textwidth}} \cite{zhang2020adversarial} \end{tabular} & 2020 & \cmark & {} & \cmark & General & \cmark \\
    \midrule
    \begin{tabular}{p{0.17\textwidth}} This review \end{tabular} & 2022 & \cmark & \cmark & \cmark & General & \cmark\\
    \bottomrule
    \end{tabular}
    \end{minipage}
    \end{center}
\end{table}

Unlike the related surveys above, this work covers a broader range of topics to review risks and solutions, especially PETs, for the preservation of privacy in the NLP field.
This is highlighted in Table~\ref{tab:comparison-related-surveys}, which compares this review against its counterparts in the literature on privacy-preserving DL and NLP. 
For instance, subjects related to solutions or DL were not approached by some of the works in the table.
Therefore, our work covers a broad range of threats, PETs, DL models, NLP tasks, and privacy metrics, bringing a holistic view of privacy preservation for NLP applications.

\subsection{Objectives and contributions}\label{sub:objectives-and-contributions}

Text data can hold private content explicitly, as a user's ID, location, or many demographic attributes, or implicitly as information inferred from the text, like a user's political view~\citep{coavoux-etal-2018-privacy}.
Furthermore, privacy has attracted great attention for developing DL methods for NLP tasks in recent years~\citep{huang2020texthide,zhuetalal2020empiricalstudies}.
Despite the relevance of this topic, there is no extensive review paper that focuses exclusively on privacy-preserving NLP, covering a large number of tasks and PETs.
Therefore, this work aims at providing an overview of recent privacy-preserving DL methods for NLP, shaping the landscape of privacy challenges and solutions in this field.
We cover all steps of the NLP pipeline, from dataset pre-processing to model evaluation, to come up with a holistic view of privacy issues and solutions for text data.
Such a review is needed to help successive scientists and practitioners in the industry have a starting point for the research in privacy-preserving NLP.

The major contribution of this work regards a taxonomy for classifying the existing works in the literature of privacy-preserving NLP, bearing in mind the target for privacy preservation, PETs, the NLP task itself, and the computation scenario. 
This taxonomy can be easily extended to aggregate future approaches and incorporate new categories of methods. 
Additional contributions of this review are summarized as follows.
\begin{itemize}
    \item First, we bring a review of PETs and point out the directions for their efficient integration into NLP models.
    \item Second, we describe several threats that put the privacy of text data at risk. To provide defenses against these threats, a model has to meet functional requirements related to data types and PETs.
    Thus, this review helps ease the efforts to find these requirements.
    \item Third, we introduce and discuss the open challenges to developing privacy-preserving NLP models, taking into account five criteria: traceability, computation overhead, dataset size, bias prevalence in embedding models, and privacy-utility tradeoffs.
    These criteria affect the suitability of PETs for real-world scenarios.
    \item Further, we bring an extensive list of benchmark datasets for privacy-preserving NLP tasks so that interested researchers can easily find out baselines for their works.
    \item Finally, we list metrics to measure the extent privacy can be protected and evaluated in NLP.
\end{itemize}

\subsection{Paper structure}\label{sub:paper-structure}

The remainder of this paper is organized as follows. 
Section~\ref{sec:research-method} outlines the methods for searching and selecting works for this review.
Section~\ref{sec:background} overviews the theoretical foundations of DL, NLP, and privacy preservation. Section~\ref{sec:literature-review} introduces a taxonomy to categorize privacy-preserving NLP methods. 
Section~\ref{sec:applications} gives a summary of applications and datasets for privacy-preserving NLP.
Section~\ref{sec:privacy-metrics} lists metrics for assessing privacy in the NLP field.
Section~\ref{sec:discussion} discusses the findings of the review and presents open problems for successive research.
Last, Section~\ref{sec:conclusions} brings the concluding remarks.

\section{Research Method}\label{sec:research-method}

This review of DL methods for privacy-preserving NLP follows a systematic literature review methodology. 
We follow the procedure proposed by~\cite{kitchenham2004procedures} in order to retrieve research papers from the existing literature, select relevant works out of the results, and summarize them afterward.
Therefore, the systematic review process is reproducible and mitigates selection biases toward the works in the literature.
Sections~\ref{sub:research-questions}, \ref{sub:search-strategy}, and \ref{sub:study-selection} outline research questions, the search strategy, and the study selection for the making of this review.

\subsection{Research questions}\label{sub:research-questions}

Research questions are the cornerstone of a literature review since every step of the review process relies on them~\citep{kitchenham2004procedures}.
To come up with this review, we answer the following research question: What are the current DL methods for privacy-preserving NLP, which provide solutions against privacy attacks and threats arising from DL models, computation scenarios, and pieces of private information in text data, such as a person's full name, demographic attributes, health status, and location?
For completeness and broader coverage of the privacy-preserving NLP topic, we split the main research question into a row of sub-questions as follows.
\begin{itemize}
    \item Which PETs have recently been implemented along DL for NLP?
    \item How can the literature on privacy-preserving NLP be organized into a taxonomy which categorizes similar approaches based on the type of data, NLP task, DL model, and PET?
    \item How can each PET influence the performance of an NLP model?
    \item How to select the most suitable PET for an NLP task?
    \item What are the tradeoffs between privacy preservation and performance for utility tasks in NLP?
    \item Which privacy metrics can be used for evaluating privacy-preserving NLP models?
    \item Which benchmark datasets are available for privacy-preserving NLP applications?
    \item What are the open challenges regarding privacy preservation in the NLP domain?
\end{itemize}

\subsection{Search strategy}\label{sub:search-strategy}

\begin{table}[h!]
    \begin{center}
    \begin{minipage}{\textwidth}
    \caption{Privacy and NLP terms used to create search expressions}
    \label{tab:search-expressions}
    \centering
    \begin{tabular}{lll}
    \toprule
    \textbf{Term 1} & \textbf{Term 2} & \textbf{NLP Terms}\\
    \toprule
     Concealment & Algorithm & Computational linguistics \\
     Confidentiality & Approach & Natural language processing \\
     Privacy & Concept & NLP \\
     Private & Framework & Text analytics \\
     Retreat & Hazard  & Text mining \\
     {} & Idea & {} \\
     {} & Manner & {} \\
     {} & Means & {} \\
     {} & Menace & {} \\
     {} & Method & {} \\
     {} & Mode & {} \\
     {} & Model & {} \\
     {} & Path & {} \\
     {} & Peril & {} \\
     {} & Preservation\slash Preserving & {} \\
     {} & Procedure & {} \\
     {} & Protocol & {} \\
     {} & Risk & {} \\
     {} & Scheme & {} \\
     {} & Solution & {} \\
     {} & Strategy & {} \\
     {} & Technique & {} \\
     {} & Threat & {} \\
     {} & Way & {} \\
    \bottomrule
    \end{tabular}
    \end{minipage}
\end{center}
\end{table}

The works we review in this article were retrieved from top venues for NLP, machine learning, AI, data security, and privacy, which are indexed by either ACL Anthology\footnote{\url{https://www.aclweb.org/anthology/}.}, ACM Digital Library\footnote{\url{https://dl.acm.org/}.}, IEEE Xplore\footnote{\url{https://ieeexplore.ieee.org/Xplore/home.jsp}.}, ScienceDirect\footnote{\url{https://www.sciencedirect.com/}.}, Scopus\footnote{\url{https://scopus.com/search/}.}, SpringerLink\footnote{\url{https://link.springer.com/}.}, or Web of Science\footnote{\url{https://apps.webofknowledge.com/}.}. 
In addition to papers indexed by the aforementioned electronic scientific libraries, we have also included valuable works in e-Print's archive\footnote{\url{https://arxiv.org/}.} since this repository stores the most up-to-date research results~\citep{zhang2020adversarial}. 
Thus, once the list of libraries was defined, we came up with search terms derived from the main research question to create search strings~\citep{kitchenham2004procedures}.
Such strings have been used to conduct the searches on the electronic scientific libraries and retrieve published works afterward.

Table~\ref{tab:search-expressions} lists the search terms we derived from the main research question in three columns. 
Firstly, column ``Term 1'' holds the privacy-related terms.
Secondly, column ``Term 2'' encompasses terms that suggest how privacy is approached.
Finally, column ``NLP Terms'' contains NLP-related terms.
In total, the table lists 34 different terms which were combined to create search strings for the electronic scientific libraries (Section~\ref{subsub:logical-operators}).

\subsubsection{The use of ``OR'' and ``AND'' Boolean operators}\label{subsub:logical-operators}

In order to construct sophisticated search strings, we have combined the terms listed in Table~\ref{tab:search-expressions} using ``OR'' and ``AND'' Boolean operators. 
First, we selected each term from column ``Term 1'' and placed it alongside each term from column ``Term 2''.
Second, we repeated this same step but used the plural form of the term from ``Term 2'' to replace the original one.
For instance, we replaced ``risk'' with ``risks'', ``threat'' with ``threats'', and so forth.
Third, we combined the outcomes from the past two steps using the ``OR'' Boolean operator to come up with the first half of each search string, such as ``((``privacy risk'' OR ``privacy risks''))''.
Similarly, we joined all terms from ``NLP Terms'' using the ``OR'' Boolean operator to construct the second half for all search strings, as ``(((``natural language processing'') OR (``NLP'')) OR ((``text mining'') OR (``text analytics'') OR (``computational linguistics'')))''.
Finally, we coupled both search string halves using the ``AND'' Boolean operator and came up with 120 different search strings to guarantee a wide-stretching coverage of the privacy-preserving NLP literature during the search step.

\subsection{Study selection}\label{sub:study-selection}

\begin{figure}[h!]
    \centering
    \includegraphics[width=\textwidth]{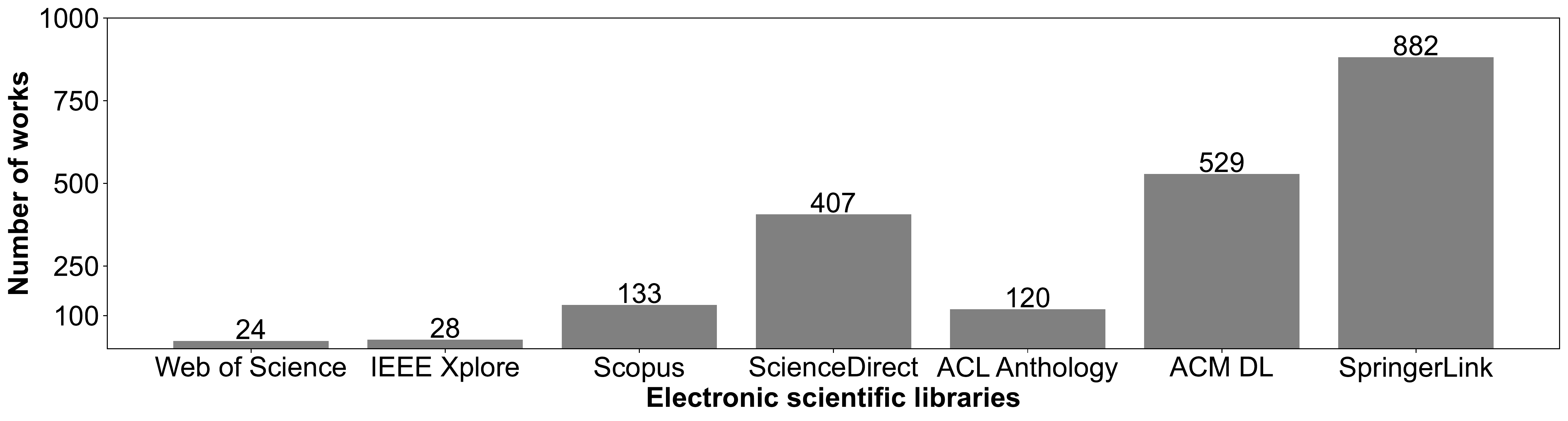}
    \caption{Literature search results}
    \label{fig:literature-search-results}
\end{figure}

The literature search on the electronic scientific libraries took place after constructing the search strings, as detailed in Section~\ref{subsub:logical-operators}.
So we applied each of the 120 search strings on the electronic libraries, retrieving 2,123 works in total.
Figure~\ref{fig:literature-search-results} shows the number of works collected from each electronic library at the end of the searches.
Searches on SpringerLink, ACM DL, and ScienceDirect returned most of the results, which account for 1,818 papers combined.
Therefore, we needed to apply some inclusion and exclusion criteria to select the most relevant works out of this plethora of results.

Published papers that satisfied all the following inclusion criteria were selected for this review.
\begin{itemize}
    \item I1. Works which employed at least one neural network model in the experimental evaluation, such as CNNs~\citep{lecun1990handwritten}, RNNs, BERT~\citep{devlin-etal-2019-bert}.
    For the sake of more extensive coverage, we also included works that reported the use of word embedding models, such as word2vec~\citep{mikolov2013distributed}, GloVe~\citep{pennington2014glove}, and fasttext~\citep{joulin2017bag}, since these models are broadly applied to NLP tasks and the privacy threats they may face are similar to those faced by DL architectures.
    \item I2. Works published from the year 2016 onward.
    Since this review aims at bringing the most recent developments of privacy-preserving NLP models based on DL, we limited the time range for publications from the past five years.
    \item I3. Long papers which report the development of privacy-preserving NLP models.
    These works were preferred over short papers since the surveyed papers were expected to present the complete results for their proposed approaches.
    However, short papers which demonstrated high impact given the number of citations were also selected.
    \item I4. Works published by top-tier venues. Many of the papers we review were published at renowned NLP, DL, and privacy conferences, such as ACL, NAACL, EACL, EMNLP, ACM SIGIR, ACM SIGKDD, NEURIPS, IEEE ICDM, and USENIX Conference, or journals, as IEEE Transactions on Pattern Analysis and Machine Intelligence and IEEE Transactions on Information Forensics and Security.
 \end{itemize}

To select e-Prints for this review, we followed three criteria applied by~\cite{zhang2020adversarial}: paper quality, method novelty, and the number of citations.
However, we used an additional criterion, namely publication date. 
If an e-Print was published before 2018, we have applied a citation number threshold of forty citations.
Otherwise, e-Prints published since 2018 were selected if they presented novel and promising approaches for privacy-preserving NLP.

Published works that satisfied any of the following exclusion criteria were removed from this review.
\begin{itemize}
    \item E.1. Published works that did not report the use of neural network models
    \item E.2. Works that did not focus on NLP or text data privacy.
    \item E.3. Works whose datasets used for the experiments did not include text data.
    \item E.4. Works published before 2016.
    \item E.5. Duplicated works. 
    \item E.6. Works which consisted of either title page or abstract only.
\end{itemize}

\begin{figure}[h!]
    \centering
    \includegraphics[width=\textwidth]{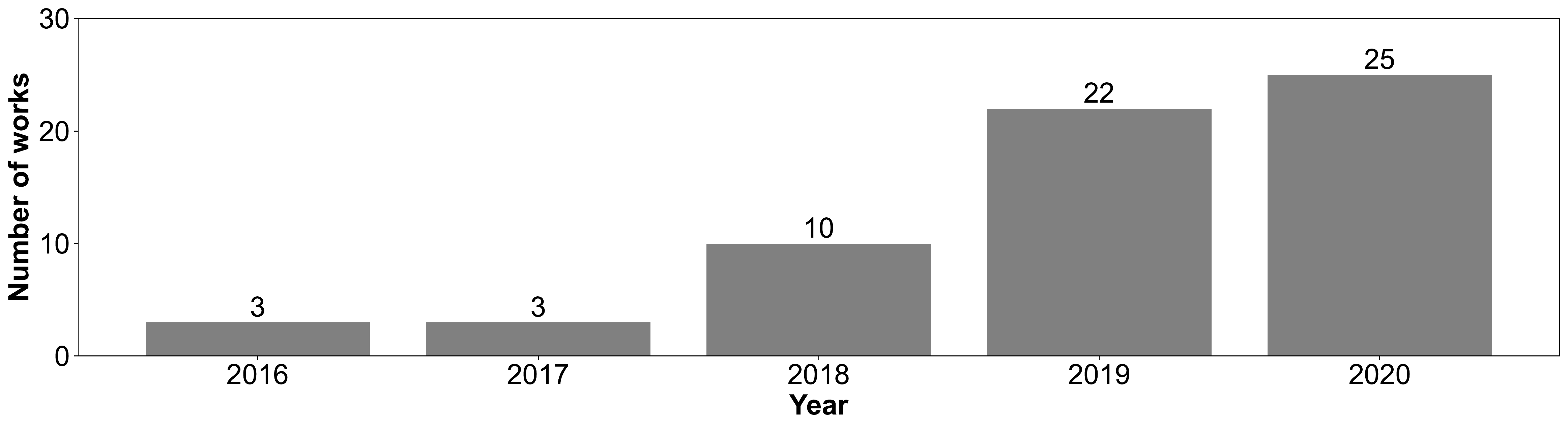}    \caption{Publication years of the selected works}
    \label{fig:number-of-works}
\end{figure}

The selection of works from the literature on privacy-preserving NLP took place between January and February 2021.
Consequently, published works indexed by the electronic scientific libraries after February 2021 were not included in this review.
After applying the criteria for inclusion and exclusion of works, 63 papers remained to be reviewed.
Figure~\ref{fig:number-of-works} shows the number of papers selected for the review by year from 2016 to 2020.
In the figure, it is possible to notice that most of the selected works were published from 2019 onward.
In contrast, the number of selected papers published in 2016 and 2017 corresponds to fewer than 10\% of the total number of selected works.
These results demonstrate the increased interest in privacy for the NLP domain in the past two years.
Therefore, this review and its main research question are justified by this ever-increasing interest.

\section{Background}\label{sec:background}

Prior to introducing the taxonomy of DL methods for privacy-preserving NLP, we need to present the theoretical background of DL and privacy for NLP.
Firstly, we briefly overview DL and its applications for NLP in Section~\ref{sub:dl}.
Secondly, we introduce the terminology related to privacy in the reviewed papers in Section~\ref{sub:ppdl}.
Finally, we list the privacy issues for NLP in Section~\ref{sub:summary-of-threats}.

\subsection{Deep learning for natural language processing}\label{sub:dl}

DL is a field of machine learning that typically includes neural network architectures featuring multiple layers, a vast number of parameters, and the ability to learn data representations whose abstraction levels are manifold~\citep{lecun2015deep}. 
These neural network models can receive enormous amounts of data as inputs and perform inference with high generalization power to reach outstanding performance results~\citep{neyshabur2017exploring}. 
The training step of a deep neural network encompasses two phases. 
Firstly, a forward pass over the data is performed~\citep{boulemtafes2020review}. 
Each network layer is initialized with random weights, so bias signals and activation functions, like ReLU~\citep{nair2010rectified}, are computed.
Then, the model outputs labels for the input data instances in case a supervised task is envisaged. 
The predicted labels are compared against the ground truth afterward~\citep{boulemtafes2020review}. 
A loss function, as binary cross-entropy~\citep{goodfellow2016deep}, computes the error rate, which will be passed backward through the network layers, consequently updating their weights and navigating downward the error gradient~\citep{lecun2015deep,boulemtafes2020review}. 
The inference step occurs after the architecture finds a minimum value for the error rate. 
For this reason, only the forward pass is computed for the testing data, predicting the labels~\citep{boulemtafes2020review}. 
DL architectures can also be trained following unsupervised and semi-supervised settings, such as autoencoders~\citep{hinton2006reducing} and ladder networks~\citep{rasmus2015semi}.

In the NLP domain, deep neural networks have brought groundbreaking results to many tasks in the past few years~\citep{feng2020securenlp,vaswani2018tensor2tensor,duarte2021deep,sarikaya2014application,liu2017deep,mozhgan2019categorizing}. 
Text data has demanded the creation of methods specially designed for its representation, such as word embedding and general-purpose language models.
We briefly introduce the main DL architectures used in the reviewed privacy-preserving NLP works, such as CNNs and RNNs, and word embedding models, which are frequently implemented alongside DL architectures.
The DL models we describe are intrinsically linked to the content of this survey.
However, we assume the reader has prior knowledge of such models.
Therefore, we omit detailed technical aspects and recommend the reader refer to seminal articles cited next to the architecture's names for complete information if needed. 

\paragraph{Convolutional neural networks} 
CNNs~\citep{lecun1990handwritten} are a family of DL architectures originally designed for computer vision tasks and applicable to data types other than images, such as text and tabular data. 
These models learn feature maps from complex and multi-dimensional inputs using multiple stacked network layers~\citep{lecun1998gradient}. 
A convolution operation for text consists on the application of a filter $\mathcal{F}$ over a window of words $w'_{i:i+j-1}$ with size $j$ for yielding a new feature $c_i$ in the form
\begin{equation}
    c_i = f'(\mathcal{F} \cdot w'_{i:i+j-1}+\mathcal{B}),
\end{equation}
in which $f'$ is a non-linear function and $\mathcal{B}$ is a bias term~\citep{kim-2014-convolutional}.
CNNs are computationally efficient methods that require fewer parameters than architectures solely relying on fully-connected layers. 
Moreover, the convolution operations are easily suitable for parallelization settings.

\paragraph{Recurrent neural networks}
RNNs are DL architectures that work with sequential data, such as text, DNA sequences, speech, and time series.
The outputs of such models depend on previous computations, which are stored in the model's internal memory, hence exploiting current and past inputs to make a prediction for new data instances~\citep{boulemtafes2020review}.
Long short-term memory (LSTM)~\citep{hochreiter1997long} is a popular RNN variant that deals with the vanishing gradient problem triggered by long input sequences~\citep{feng2020securenlp,boulemtafes2020review,hochreiter1997long}.
LSTM takes an input sequence $\{x_1,x_2,\dots,x_\mathcal{T}\}$ and transforms it into another sequence $\{y_1,y_2,\dots,y_\mathcal{T}\}$ by a series of layers, which comprise hidden cells and hidden parts of state memory~\citep{feng2020securenlp}.
Each layer is composed of gates, namely the input gate, forget gate, and output gate, which control the amount of information to be stored or forgotten by the model in the hidden states, update the hidden states, and decide on the final output.
Another popular RNN variant is bidirectional LSTM (BiLSTM)~\citep{graves2005framewise}, which encompasses two LSTM models for processing sequence data in both forward and backward directions, hence improving the amount of information and the data context for the architecture~\citep{cornegruta-etal-2016-modelling}.

\paragraph{General-purpose language models}

In the past few years, a new paradigm to yield language representations has been noticed.
The so-called general-purpose language models comprise giant pre-trained language models which are built upon multiple layers of transformer~\citep{vaswani2017attention} blocks and feature millions of parameters learned during the pre-training step on billions of sentences~\citep{pan2020privacy}.
These models are designed to encode whole sentences as vectors (embeddings) and, after the pre-training step, are publicly released to be optionally adapted (fine-tuned) for NLP tasks.
Another outstanding feature of such models regards their ability to learn new tasks from a few data instances.
BERT~\citep{devlin-etal-2019-bert} is a general-purpose model which yields language representations by conditioning the context on both sides of target words. 
It relies on a loss function based on masking some tokens and trying to predict the word id of the masked words solely based on the surrounding context. 
There are two standard model sizes for BERT concerning the number of layers ($L$), hidden size ($\mathcal{H}$), self-attention heads ($\mathcal{A}$), and parameters ($\Theta$) used to build the model architectures. 
The first one is BERT$_{BASE}$($L = 12$, $\mathcal{H} = 768$, $\mathcal{A} = 12$, $\Theta = 110M$), and BERT$_{LARGE}$($L = 24$, $\mathcal{H} = 1024$, $\mathcal{A} = 16$, $\Theta = 340M$) is the second architecture.
BERT variants have pushed state-of-the-art results forward in many NLP tasks. 
However, BERT-based models require a high memory footprint due to the enormous parameter sizes.
Additional huge models in the same category as BERT are GPT~\citep{radford2018improving}, GPT-2~\citep{radford2019language}, and GPT-3~\citep{NEURIPS2020_1457c0d6}.

\paragraph{Word embedding models}
Word embeddings are distributed representations of words yielded by shallow neural networks in order to reduce the number of dimensions of such vector representations, whereas semantic features of words, such as context, are preserved~\citep{camacho2018word}.
These representations are extensively used across NLP tasks, mostly combined with DL architectures.
Therefore, based on the relatedness between word embeddings and DL, we also include privacy-related topics arising from embedding models in this review.
Popular word embedding models are word2vec~\citep{mikolov2013distributed}, GloVe~\citep{pennington2014glove}, and fasttext~\citep{joulin2017bag}.
All these models are able to capture semantic properties of words~\citep{camacho2018word}, hence representing related terms as close coordinates in a vector space.

\subsection{Privacy terminology}\label{sub:ppdl}

\begin{table}[ht!]
    \begin{center}
    \begin{minipage}{\textwidth}
    \caption{Topics and terminology for privacy preservation for NLP}
    \label{tab:nomenclature}
    \centering
    \begin{tabular}{ll}
    \toprule
    \textbf{Topic} & \textbf{Key terms} \\
    \toprule
     Bias and fairness in NLP & \parbox[t]{0.45\textwidth}{Demographic attributes, adversarial training, adversarial components, debiasing, gender-neutral word embeddings, approximate fairness risk, attributes obfuscation, fair representation learning, neural machine translation} \\ 
     \midrule
     Privacy-preserving NLP & \parbox[t]{0.45\textwidth}{Private training, private inference, identifiable words, nontransferable data, laws and regulations, private information, lack of trust, non-traceability, data safeguarding, PETs} \\ 
     \midrule
     Disentangled representations & \parbox[t]{0.45\textwidth}{Factors of variation, optimization, separate latent spaces, adversarial objective functions, transfer learning, style transfer}\\
    \bottomrule
    \end{tabular}
    \end{minipage}
\end{center}
\end{table}

Text data may contain pieces of private information explicitly or implicitly integrated into its content, such as precise key phrases or demographic attributes inferred from the context~\citep{coavoux-etal-2018-privacy}. 
Furthermore, privacy-related issues may occur at any phase of the NLP pipeline, from data pre-processing to downstream applications.
Privacy preservation, for this reason, involves the perspectives of data~\citep{chen2019federated}, DL model~\citep{song2020information}, PETs~\citep{melamud2019towards}, fairness~\citep{ekstrand2018privacy}, computation scenario~\citep{feng2020securenlp}, downstream tasks~\citep{sousa2021privacy}, and the interplay between these perspectives~\citep{ekstrand2018privacy}.
These multiple perspectives contribute to ever-increasing literature from which many privacy-related topics develop.
In Table~\ref{tab:nomenclature}, we list the three major topics for privacy preservation for NLP beside their related key terms.
First, bias and fairness in NLP is a topic regarding automated decision-making systems which are prone to biases arising from pieces of private information, like gender~\citep{elazar-goldberg-2018-adversarial}, yet wished not to prompt discriminatory outputs~\citep{ekstrand2018privacy}.
The efforts to promote fair decision-making often interact with those for privacy protection~\citep{ekstrand2018privacy}.
Second, privacy-preserving NLP is a catchphrase for approaches that protect privacy for models combined with PETs~\citep{feng2020securenlp,alawad2020privacy}.
Finally, disentangled representations play a key role in integrating privacy premises into learning data representations used for downstream tasks~\citep{lee2021private}.
We further describe each of these topics in the paragraphs below.

\textbf{Bias} in machine learning is a concept related to unfair or discriminatory decisions made by models towards or against specific individuals or groups~\citep{ntoutsi2020bias}. 
It becomes a severe problem in real-world scenarios since those decisions may directly impact a person's life or society as a whole. 
For instance, gender-biased job advertising tools were found to suggest lower-paying job positions to women more often than men~\citep{datta2015automated}. 
Human biases are long-lasting research topics in philosophy, social sciences, psychology, and law~\citep{ntoutsi2020bias}. 
There are three types of bias: preexisting bias (from the data), technical bias (from models with computational or mathematical constraints), and emergent bias (from the evaluation of results and their applicability)~\citep{papakyriakopoulos2020bias}.
Recently, this subject has also been gaining attention in the NLP field since word representations, like the widely used word embeddings~\citep{mikolov2013distributed}, can also be under threat of encoding human bias~\citep{kaneko2019gender} regarding gender, ethnicity, or social stereotypes from text corpora used for training. 
Therefore, applications built on such biased NLP models have the potential to amplify such misuses of language and propagate them to inference steps. 
\textbf{Fairness} is then typically associated with trustworthy AI~\citep{floridi2019establishing} and defined as an assurance against discriminatory decisions by AI models based on sensitive features~\citep{zhang2018fairness}.
Privacy technologies and policies often go hand in hand with concepts of fairness~\citep{ekstrand2018privacy}.
Additionally, works approaching bias and fairness in NLP follow the premise that the input data present sensitive features, e.g., gender~\citep{elazar-goldberg-2018-adversarial,zhao2018learning}, which may lead to unfair predictions by downstream systems or be recovered from representations. Therefore, debiasing and attribute removal are techniques to mitigate the undesired discriminatory effects arising from unfair models.
Examples of use case tasks covering bias and fairness encompass learning gender-neutral word embeddings~\citep{zhao2018learning,bolukbasi2016man}, analysis and reduction of gender bias in multi-lingual word embeddings~\citep{zhao2020gender,font2019equalizing}, text rewriting~\citep{xu-etal-2019-privacy}, analysis of biases in contextualized word representations~\citep{tan2019assessing,hutchinson2020social,gonen2019lipstick,Basta2020ExtensiveSO}, detection, reduction and evaluation of biases for demographic attributes in word embeddings~\citep{papakyriakopoulos2020bias,sweeney2020reducing,sweeney2019transparent,kaneko2019gender}, analogy detection~\citep{nissim2020fair}, cyberbullying text detection~\citep{gencoglu2020cyberbullying}, fair representation learning~\citep{friedrich-etal-2019-adversarial}, protected attributes removal~\citep{elazar-goldberg-2018-adversarial,barrett-etal-2019-adversarial}, analysis of racial disparity in NLP~\citep{blodgett2017racial}, and prediction of scientific papers authorship during double-blind review~\citep{caragea2019myth}.

\textbf{Privacy-preserving NLP} is an expression that refers to language models trained or used for inference on private data without putting privacy at risk. 
Some assumptions are considered for the development of such methods: (i) encoded sensitive information about the input must be kept private~\citep{coavoux-etal-2018-privacy,mosallanezhad-etal-2019-deep,feyisetan2019leveraging} (e.g., personal attributes, demographic features, location, etc.); (ii) the model's vocabulary may contain words that easily identify people in the data~\citep{alawad2020privacy,li2018towards}; (iii) personal data should never leave their owner's devices~\citep{chen2019federated,alawad2020privacy,hard2018federated}; (iv) the data are subject to legal terms and regulations~\citep{clinchant2016transductive,melamud2019towards,belli2020privacy,battaglia2020towards,martinelli2020nlpijcnn}; (v) private information may be correlated with the labels of the model outputs with high likelihood and learned thereby~\citep{coavoux-etal-2018-privacy,li2018towards,song2020information,carlini2019secret}; (vi) the computation scenario is not trusted against privacy attacks and threats, such as eavesdropping, breaches, leaks, or disclosures~\citep{feng2020securenlp,coavoux-etal-2018-privacy,dai2019efficient,feyisetan2020privacy,liu2020mitigating}; (vii) the input data must be untraceable to any users but the data owner~\citep{oak2016generating}. 
Among the PETs for NLP, there are anonymization~\citep{oak2016generating}, data sanitization~\citep{feyisetan2019leveraging}, data obfuscation~\citep{martinelli2020nlpijcnn}, text categorization~\citep{battaglia2020towards}, transfer learning~\citep{alawad2020privacy,song2020information}, FL~\citep{chen2019federated,hard2018federated}, black box model adaptation~\citep{clinchant2016transductive}, encryption~\citep{dai2019efficient,liu2020mitigating}, MPC~\citep{feng2020securenlp}, DP~\citep{feyisetan2020privacy,melamud2019towards}, adversarial learning~\citep{li2018towards}, deep reinforcement learning~\citep{mosallanezhad-etal-2019-deep}, and generative models~\citep{carlini2019secret}.

\textbf{Disentangled representations} regard the premise that good data representations capture factors of variation from the input feature space and represent these factors separately~\citep{bengio2009learning}. 
Latent spaces of neural networks can be, thus, disentangled as to different features, e.g., adding terms to the model's objective functions as an adversarial training setting~\citep{john2019disentangled}. 
Images in computer vision tasks are popular targets for learning disentangled representations~\citep{NIPS2016_6051,lee2021private}. Moreover, recent NLP approaches helped yield representations for language features, such as style and content, as separate latent spaces for style transfer tasks to be performed afterward~\citep{john2019disentangled}. 
Therefore, sensitive features are preserved while the representations for the remaining ones can be used for applications without putting privacy at risk.  

\subsection{Privacy issues for NLP}\label{sub:summary-of-threats}

Privacy issues arise in situations where an attacker can successfully associate a record owner to a sensitive attribute in a published database~\citep{MENZIES2015165}, disclose model inputs~\citep{li2018towards,song2020information,huang2020texthide}, obtain information that should be kept private~\citep{lyu2020differentially,coavoux-etal-2018-privacy,feng2020securenlp}, among other harmful activities.
Consequently, unintended data breaches may occur and lead to problems, such as social exposure, documents leakage, and damages to an individual's or organization's reputation.
Furthermore, data protection laws establish penalties and fines if a data breach happens.
Therefore, when a DL model is designed to process personal data, it is crucial to consider the privacy threats that put this model at risk of data breaches.

We list over fifteen different privacy threats and attacks to ease their identification for designing DL models for privacy-preserving NLP.
We take into account three perspectives to grouping privacy threats in NLP.
Firstly, the threats arising from datasets that are made public and therefore can have their original content disclosed.
Secondly, the threats related to how DL models can violate data privacy. 
For instance, a model can memorize protected attributes and allow their disclosure later on~\citep{kumar-etal-2019-topics}. 
Another model threat regards how language models address human discriminatory biases from their training text corpora~\citep{Basta2020ExtensiveSO,sweeney2020reducing}.
Moreover, the computation scenario, such as centralized cloud servers or distributed processing architectures, plays an important role in the existence of privacy threats since many of them are related to the misbehavior of components.
So we present the most common threats from the computation scenario in the reviewed papers.
Finally, we overview privacy attacks that target DL models for NLP.

\subsubsection{Threats from data}\label{subsub:threats-from-data}

\begin{figure}[t!]
    \centering
    \includegraphics[width=\textwidth]{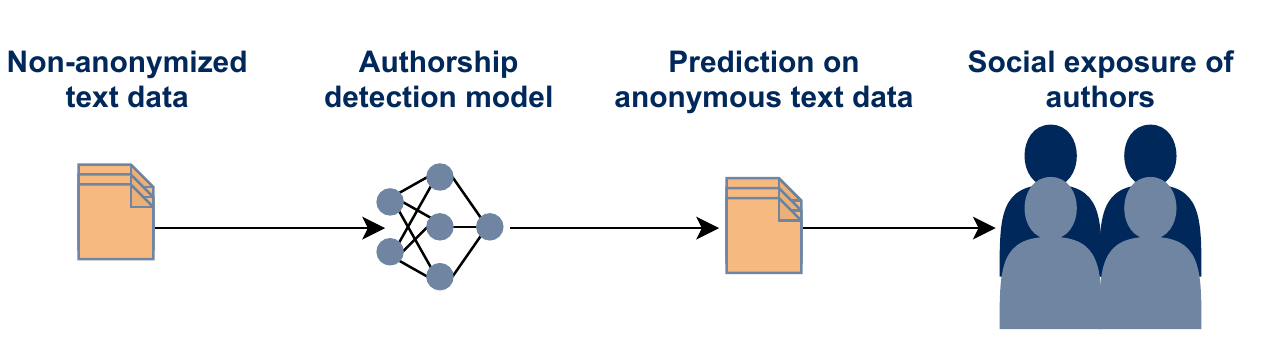}
    \caption{Re-identification of anonymous texts. In this setting, an authorship detection model can threaten the identities of anonymous text data authors from social exposure.}
    \label{fig:re-identification-of-documents}
\end{figure}

From the data perspective, the most common privacy threats for text data are related to its content~\citep{clinchant2016transductive}, which encompasses pieces of private information like identities of authors, health status, sentiment polarities, and demographic attributes.
In the reviewed works, we identified the following privacy threats arising from the data perspective.
\begin{itemize}
    \item \textit{Hardness to tag sensitive information}. Data privacy frequently relies on the obfuscation of sensitive information of documents, which is a hard task since all direct and indirect informational clues that may identify a person should be obfuscated~\citep{martinelli2020nlpijcnn}. Furthermore, the sensitive content may be expressed by words that are not sensitive themselves, as describing sensitive bank transactions using the same vocabulary as descriptions of non-sensitive ones, but with different natural language expressions~\citep{neerbek2018detecting}.
    Therefore, information taggers based on keyword lists are susceptible to failure.
    \item \textit{Re-identification of documents and anonymous text}. Documents such as electronic medical records and government reports are usually publicly released for the sake of leveraging research on their respective domains or government transparency~\citep{fernandes2019generalised}.
    Although these documents are required to be sanitized by the removal of their authorship information to protect their authors, re-identification by malicious models can still take place~\citep{fernandes2019generalised}.
    Online anonymous texts, such as e-mails, comments, and blog posts, can also be re-identified by authorship detection models trained on non-anonymized data~\citep{seroussi2014authorship}, as depicted by Figure~\ref{fig:re-identification-of-documents}.
    Additionally, authorship identification models can have beneficial applications to intellectual property management~\citep{boumber2018experiments}, as in plagiarism detection.
    Thus, formal guarantees against re-identification of documents have to be provided by efficient de-identification methods, for instance, via DP~\citep{dwork2008differential,fernandes2019generalised}.
    \item \textit{Re-identification of anonymous source code}. Source code of open source projects can be used to identify the developers based on their coding style~\citep{abuhamad2018large,abuhamad2019code}. This threat is particularly dangerous when developers do not wish to expose their identities. 
    \item \textit{Self-disclosure of emotions and personal information}. Social media posts often carry private information voluntarily released by users, such as gender, location, career, and feelings towards things and people~\citep{akiti-etal-2020-semantics,battaglia2020towards}.
    These users are frequently not aware of the sensitivity of the information they post so that models trained on such data can be input with private information without notice. 
    \end{itemize}
    
\subsubsection{Threats from models}\label{subsub:threats-from-models}

Model properties, such as vocabulary and neural network layers, can be used by an adversary to perform attacks that end up disclosing private data used for training~\citep{alawad2020privacy}. 
Additional privacy-related issues from NLP models concern biases~\citep{sweeney2020reducing,gencoglu2020cyberbullying,tan2019assessing} and the unfair decisions made by biased models~\citep{sweeney2020reducing,xu-etal-2019-privacy}.
Therefore, from the perspective of NLP models, privacy threats are as follows.
\begin{itemize}
    \item \textit{Bias in word embedding models}. The encoding, amplification, and propagation of human discriminatory biases are noticeable privacy-related issues for word embedding models. The most common types of encoded biases regard gender~\citep{Basta2020ExtensiveSO,bolukbasi2016man,font2019equalizing,gencoglu2020cyberbullying,kaneko2019gender,nissim2020fair,papakyriakopoulos2020bias,sweeney2020reducing,tan2019assessing,vig2020causal,zhao2018learning}, race~\citep{sweeney2020reducing,tan2019assessing}, professions~\citep{papakyriakopoulos2020bias},
    religion~\citep{sweeney2020reducing},
    intersectional identities~\citep{tan2019assessing},
    language~\citep{gencoglu2020cyberbullying}, and disabilities~\citep{hutchinson2020social}, to name a few.
    The removal or lessening of such issues is challenging since the semantics of the representations yielded by the models should be preserved, whereas the discriminatory biases should be removed to the largest extent possible.
    Given the hardness of performing de-biasing of embedding models, biased or unfair decisions towards demographic attributes can be made~\citep{xu-etal-2019-privacy,sweeney2020reducing,bolukbasi2016man}.
    Furthermore, it was found that bias may present some prevalence on debiased embedding models~\citep{gonen2019lipstick}, hardening its complete removal.
    Finally, the transfer of gender bias across languages in multilingual word embeddings is another threat to be taken into account~\citep{zhao2020gender}. For instance, word embeddings generated for a neural machine translation task~\citep{feng2020securenlp}, which translates sentences from Spanish into English, may capture gender-related bias from Spanish and integrate it into the embeddings of English words.
    
    \item \textit{Disclosure of protected health information}. Patient data is inherently private since it holds attributes related to a person's identity, health status, diagnosis, medication, and demographic information.
    It is protected by regulations, such as the Health Insurance Portability and Accountability Act (HIPAA)~\citep{act1996health} in the United States, hence demanding de-identification prior to the public release of such data for research activities~\citep{liu2017identification,dernoncourt2017identification,obeid2019impact}.
    NLP models for healthcare data often suffer threats from sharing vocabulary dictionaries, which encompass entries related to patient identities, for the embedding layer of neural networks~\citep{alawad2020privacy}.
    
    \begin{figure}[ht!]
    \centering
    \includegraphics[width=\textwidth]{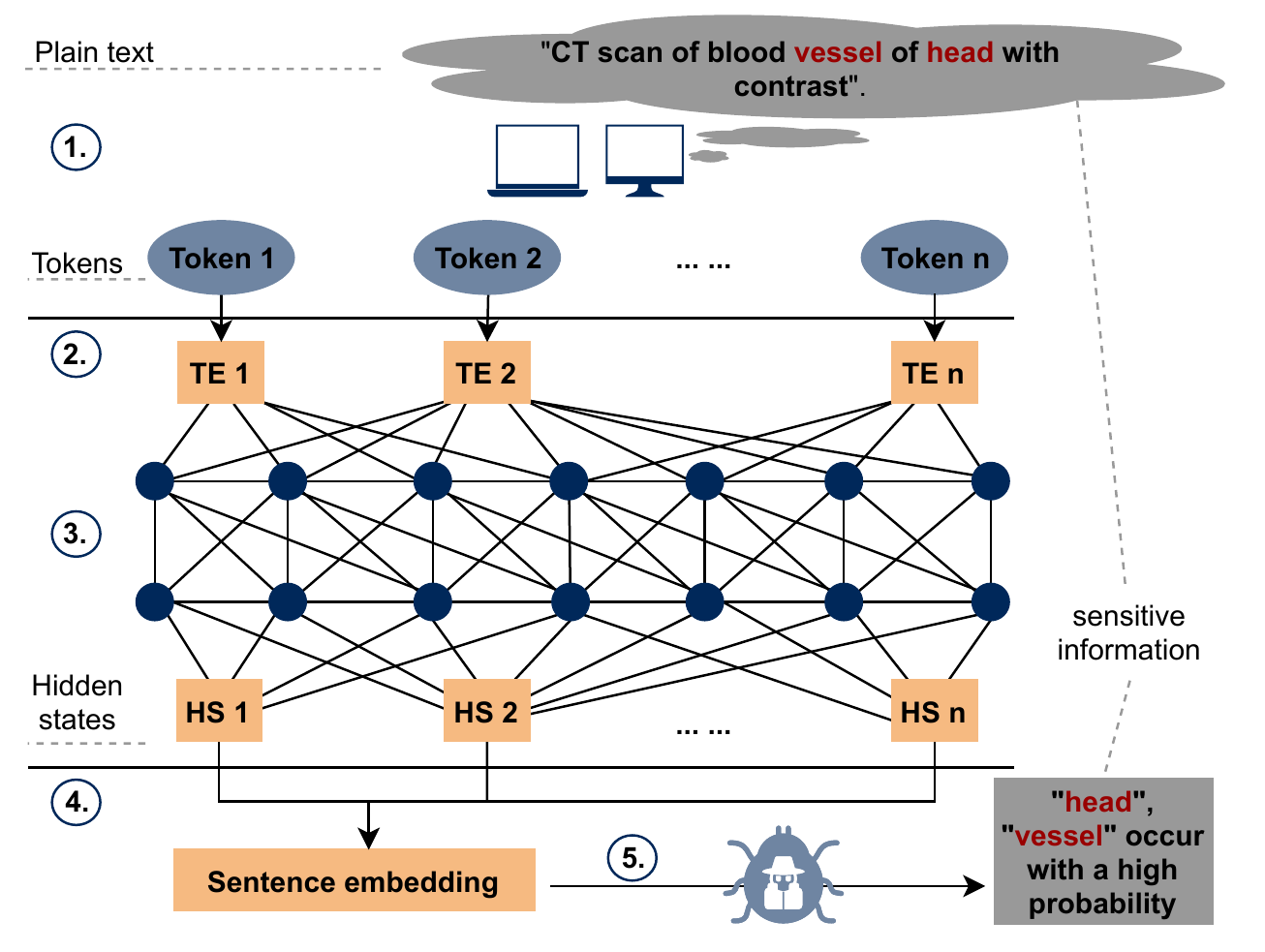}
    \caption{Unintended feature memorization~\citep{pan2020privacy}. The bold red text represents privacy-sensitive features memorized by the general-purpose language model. The unintended memorization and successive reconstruction of such features from sentence embeddings can occur as follows. \textbf{1.} Plain text tokenization. \textbf{2.} Token embedding (TE). \textbf{3.} Propagation across transformer layers. \textbf{4.} Pooling over hidden states (HS) to yield a sentence embedding. \textbf{5.} Inference of sensitive information from sentence embeddings.}
    \label{fig:unintended-memorization}
\end{figure}

    \item \textit{Unintended feature memorization}. Firstly, deep neural NLP models can learn topical features containing spurious correlations from the training data, which damage the model performance on prediction tasks~\citep{kumar-etal-2019-topics}. 
    For instance, the prediction of an author's native language from texts written in a second language can be biased towards geographical terms frequently occurring in their text, like country names, yet not related to the task target.
    Consequently, these models are likely to fail to generalize on new data instances because of such learned features~\citep{kumar-etal-2019-topics}, which may also include pieces of private information.
    Secondly, general-purpose language models with millions of learnable parameters, like BERT and GPT, also capture sensitive information from the training data, hence being at risk from model inversion attacks, which retrieve this information from the trained model~\citep{pan2020privacy}.
    Figure~\ref{fig:unintended-memorization} depicts the unintended memorization of privacy-sensitive attributes from the text while yielding sentence embeddings.
    In the figure, an attacker infers these attributes from the sentence embeddings afterward.
    Finally, DL models were also found to memorize unique or rare training data sequences~\citep{carlini2019secret}. Therefore, first-line defenses like removing such pieces of private information can hinder this issue.
\end{itemize}

\subsubsection{Threats from the computation scenario}\label{subsub:threats-from-the-compuattion-scenario}

The computation scenario plays an important role in putting privacy at risk.
For instance, communication channels, servers~\citep{dai2019efficient}, or computation parties~\citep{feng2020securenlp} may behave in unreliable manners.
Therefore, the following threats can take place.
\begin{itemize}
    \item \textit{Disclosure of data from user devices}. In FL scenarios, model parameters may allow adversaries to learn about the original training data~\citep{mcmahan2018learning}. 
    For instance, model gradients computed locally on users' devices can carry implicit private information from the locally stored data like user's behavior~\citep{qi-etal-2020-privacy}.
    Therefore, DP methods can be used to provide privacy guarantees at the user level~\citep{mcmahan2018learning}.
    \item \textit{Honest-but-curious server}. Another example of a semi-honest security model regards cloud or central servers that are not fully trustworthy~\citep{dai2019efficient,zhuetalal2020empiricalstudies}.
    This threat requires encryption or DP noise to be applied to the data or model parameters before outsourcing to the server.
    \item \textit{Semi-honest security model}. This threat, also referred to as the honest-but-curious security model, frequently happens in MPC settings, in which a corrupted party follows the MPC protocols exactly, but this party tries to learn more information than expected during the iterations~\citep{feng2020securenlp}. 
\end{itemize}

\subsubsection{Privacy attacks}\label{subsub:attacks}

In DL-based NLP, privacy attacks aim at leaking data samples used for model training, exposing individuals and their private information, such as identity, gender, and location.
Many factors influence the likelihood of success an attacker model can obtain, e.g., the DL model itself~\citep{mosallanezhad-etal-2019-deep}, the computation scenario~\citep{lyu2020differentially}, and the data properties~\citep{oak2016generating}.
Therefore, there is a large number of privacy attacks that target text data used to train DL models.
In the surveyed works, we identified nine different privacy attacks, which we describe as follows.
\begin{itemize}
    \item \textit{Adversarial attacks}. Malicious modifications of texts with a small impact on the readability by humans, yet able to make DL models output wrong labels, constitute adversarial attacks~\citep{liu2020mitigating,zhang2020adversarial}.
    Examples of such modifications include character-level perturbations and the replacement of words by semantic similarity or probability.
    
    \item \textit{Membership inference attacks}.
    Membership inference attacks comprise a widely researched class of attacks against DL models.
    In these attacks, an adversary attempts to disclose the `is-in' relation between a data sample and the original private training set of a model~\citep{pan2020privacy}.
    
     \item \textit{Attribute inference attacks}.
     Text data encompasses a large number of private attributes which can be leaked through embeddings for words, sentences, or texts, which are trained without efficient anonymization.
     These attributes include gender, age, location, political views, and sexual orientation~\citep{mosallanezhad-etal-2019-deep}. 
     For instance, this attack can take place as a classifier that predicts private information, like location, of real-time system users from the embeddings of their texts~\citep{elazar-goldberg-2018-adversarial,mosallanezhad-etal-2019-deep,barrett-etal-2019-adversarial}.
     Thus, language representations shared over different tasks or computation parties have to be robust against those attacks.
     
     \item \textit{Re-identification attacks}. 
     Anonymized documents can still hold information that can be used to trace back the individuals who generated it, using auxiliary data sources~\citep{oak2016generating}.
     An example of such an attack was the re-identification of the Netflix Prize dataset, a database composed of 100,480,507 movie ratings of 480,189 Netflix users in the years between 1999 and 2005, using the IMDB dataset as a surrogate for underlying knowledge about the attack targets~\citep{narayanan2008robust}.
     
     \item \textit{Eavesdropping attacks}.
     An eavesdropping attack happens in scenarios in which the computation is distributed across many devices~\citep{lyu2020differentially}, e.g., FL.
     Thus, one of the devices would try to infer private information from, for instance, latent representations sent to a cloud server by other devices in the setting~\citep{coavoux-etal-2018-privacy}.
     
     \item \textit{File injection attacks}. 
     Searchable encryption~\citep{cash2015leakage} can have its privacy guarantees broken by file injection attacks, which can be seen as a more general class of adversarial attack.
     In such attacks, an adversary injects files composed of keywords into a client of a cloud server, which will encrypt the injected files and store them on the server~\citep{liu2020mitigating}. 
     Therefore, the attacker will observe the patterns of the encrypted files, threatening query files, and disclose user keywords~\citep{liu2020mitigating}.
    
    \item \textit{Reverse engineering attacks for language models}.
    DL models for NLP can be easily reverted by an adversary that has prior knowledge of the model~\citep{li2018towards}.
    Consequently, this adversary may be able to reverse engineer the input data sampled and leak private information from the training examples.
    Embedding models are prone to these attacks since word vectors also leak information about the input data~\citep{song2020information}.
    Therefore, preventing reverse engineering attacks for NLP is a tricky challenge, especially for FL settings, in which the training cannot be slowed down, and the model accuracy should fall short.
    In FL scenarios, the adversaries can be corrupted devices that have access to information communicated by all parties in the computation like parameters of the model during the training~\citep{huang2020texthide}.
    Potential solutions include DP~\citep{dwork2008differential}, FHE~\citep{gentry2009fully}, or the combination of both~\citep{huang2020texthide}.
    
    \item \textit{Pattern reconstruction attacks}. 
    For pattern reconstruction attacks, the text in its original format presents a fixed structure, like a genome sequence, and the adversary tries to recover a specific part of this sequence that contains a piece of sensitive information~\citep{pan2020privacy}.
    A gene expression related to an illness can be an example of such sensitive information.
    
    \item \textit{Keyword inference attacks}.
    Sometimes the adversary is solely interested in probing whether a plaintext includes a given sensitive keyword~\citep{pan2020privacy}.
    For instance, the plaintext can be a clinical note, and the sensitive keyword can be a disease location.
    Therefore, the adversary tries to recover her/his interested keywords.
    
    \item \textit{Property inference attacks}. 
    Unlike membership inference attacks, property inference attacks regard the attempts of an adversary to discover global properties of the original training set, such as the class distribution~\citep{pan2020privacy}.
    These attacks pose privacy threats for NLP models since the predicted properties may not be shared by the model producer in a consenting manner~\citep{ganju2018property}.
    For instance, an adversary may intend to disclose the male-female ratio of a given population related to the electors of a political party.
    In order to mask this sensitive information against property inference attacks, a model can apply DP noise to the dataset.
    
\end{itemize}

\section{Deep learning methods for privacy-preserving NLP}\label{sec:literature-review}

The protection of privacy in NLP is a challenge whose solution depends on many factors, such as computation scenario, utility performance, memory footprint, dataset size, data properties, NLP task, and DL model.
Consequently, choosing a suitable PET is not a problem of solely protecting the greatest extent of privacy as possible since privacy-utility tradeoffs can either turn a solution feasible for a real-world application or impractical otherwise.
In the past few years, privacy has been attracting significant attention.
So many works have been addressing it for DL and NLP, yet making it hard to follow the progress of the literature due to lack of categorization.
Therefore, we propose a taxonomy that organizes this literature and shapes the landscape of DL methods for privacy-preserving NLP.

\subsection{Categories of DL methods for privacy-preserving NLP}\label{sub:summary-of-methods}

\begin{figure}[t!]
    \centering
    \includegraphics[width=\textwidth]{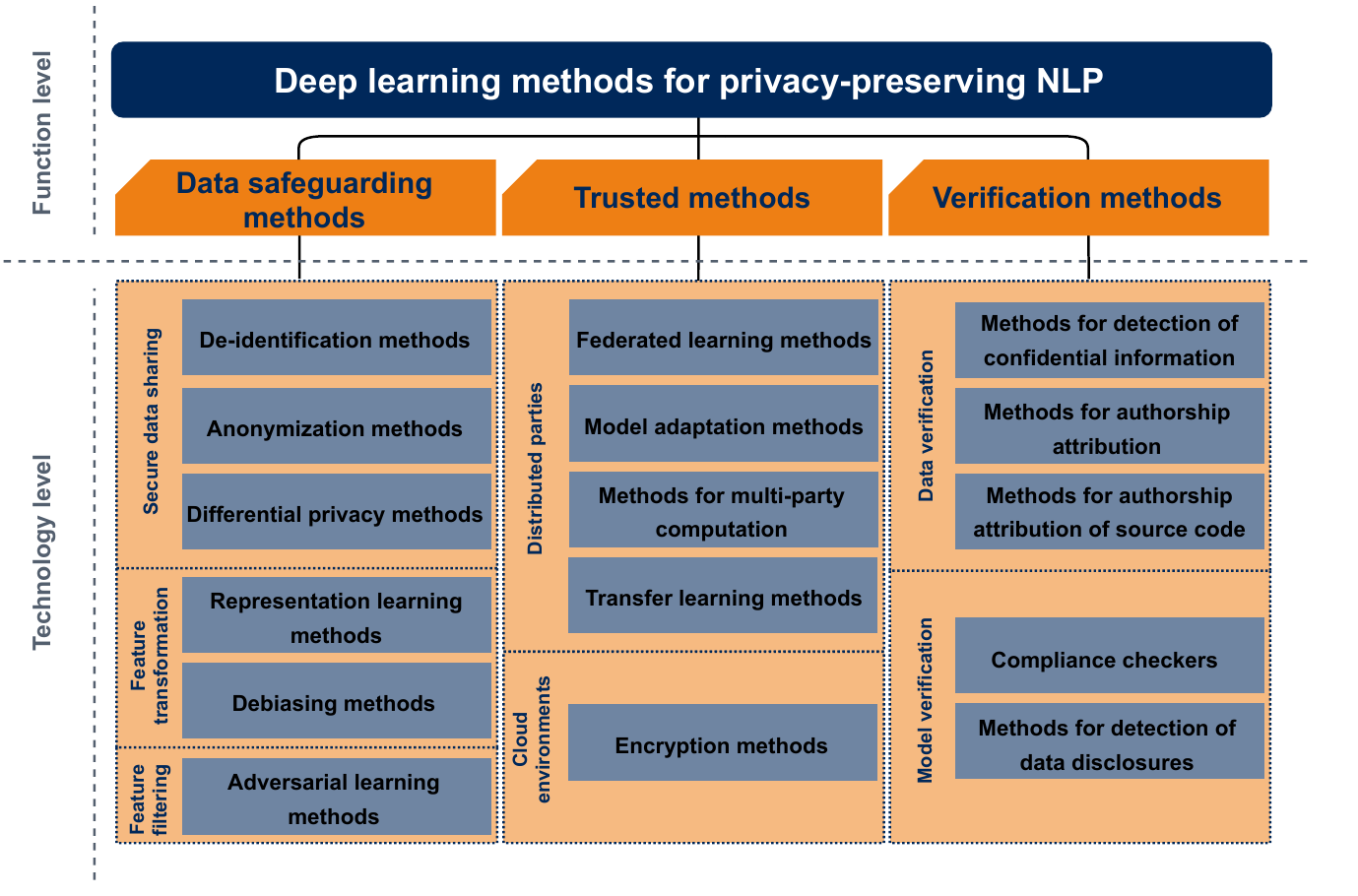}
    \caption{Taxonomy of DL methods for privacy-preserving NLP}
    \label{fig:taxonomy}
\end{figure}

When it comes to privacy-preserving NLP approaches based on DL, we can find similarities between methods considering two major factors: the target of privacy preservation and the PETs specifically.
The former determines where privacy is assured, such as in the dataset prior to training and inference, model components during the learning phase, or post-processing routines.
The latter specifies which existing PETs are appropriate for each privacy scenario.
For instance, encryption is recommended when the server where the data is stored, or another computation party, is no longer trusted.
So we gather the methods which implement encryption schemes for utility tasks of NLP into a group of encryption methods.
Additionally, since encryption methods are commonly implemented alongside a DL model and remain in place during model training and inference, we insert them into the category of methods whose privacy focus is on the model side, namely trusted methods.  
This category is divided into two sub-categories according to the computation scenarios for which the trusted methods are implemented.
Similarly, we followed this insight to come up with a taxonomy (Figure~\ref{fig:taxonomy}) which is composed of two levels, three categories, seven sub-categories, and sixteen groups for the surveyed methods and their respective PETs.

In Figure~\ref{fig:taxonomy}, we depict the full proposed taxonomy in two levels: function level and technology level.
The former refers to the target of privacy preservation throughout the NLP pipeline, and the latter separates the methods into groups based on the PETs they implement and the computation scenarios they approach.
In the taxonomy structure, there are three categories of methods.
First, the category of data safeguarding methods is the most extensive in the structure.
It aggregates methods divided into six groups, accounting for twenty-seven works in total, covering PETs that are run before the model training and inference, such as debiasing of word embeddings.
Second, methods for privacy preservation during model training or inference constitute the category of trusted methods.
This category encompasses five groups which account for fourteen works in total. Finally, the category of verification methods includes the remaining five groups with twenty-two works, which aim to detect confidential information in text documents or even assess how susceptible to privacy threats DL models for privacy-preserving NLP are.
This taxonomy serves as a framework for categorizing the existing literature on privacy-preserving NLP, easily extendable for aggregating successive works.
Moreover, it helps researchers and practitioners identify the most suitable PET for their needs.

\subsection{Data safeguarding methods}\label{subsub:data-safeguard-methods}

\begin{figure}[ht!]
    \centering
    \includegraphics[width=\textwidth]{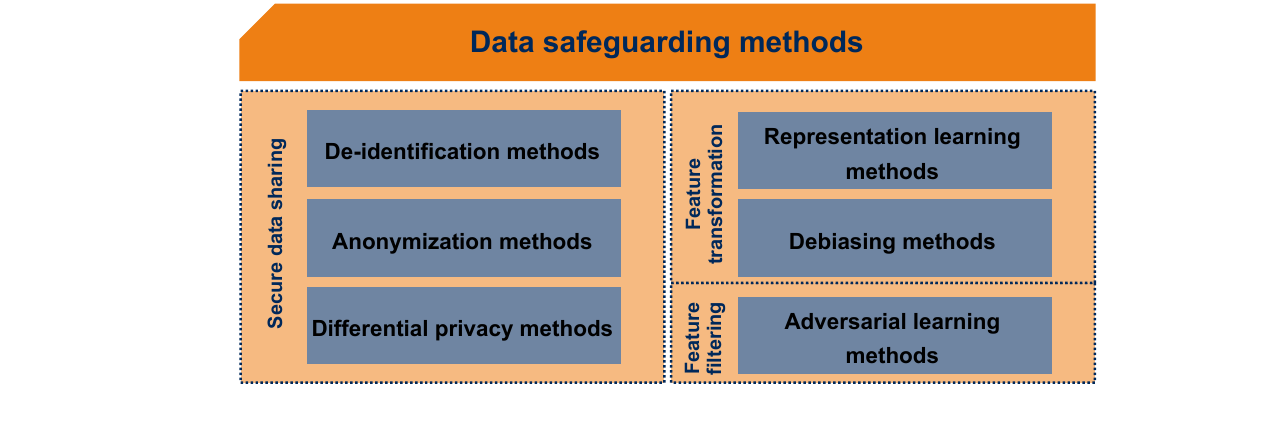}    \caption{Sub-categories and groups for data safeguarding methods}
    \label{fig:data-safeguarding-methods}
\end{figure}

Data safeguarding methods are applied over datasets shared between NLP tasks or used for downstream applications.
Figure~\ref{fig:data-safeguarding-methods} depicts these groups, which approach secure data sharing, feature transformation, and feature filtering.
Subsequently, Table~\ref{tab:summary-of-ds-methods} summarizes the groups of works for data safeguarding, including their neural network models, PETs, and computation scenarios.

\subsubsection{De-identification methods}\label{subsub:de-identification-m}

De-identification of data instances can be seen as a named-entity recognition problem~\citep{liu2017identification} to scrub private content from the text by detecting and replacing it with synthetic instances of text, mostly from domains that deal with protected health information (PHI)~\citep{el2009globally}. 
For instance, a patient's name can be supplanted by a generic tag (e.g., `PHI'), a private information class descriptor (e.g., `patient'), or a randomly generated surrogate word from the same information class (e.g., a pseudonym)~\citep{obeid2019impact,eder2020code}. Furthermore, regulations like the HIPAA~\citep{act1996health} in the United States enforce any PHI to be safeguarded from disclosures without the patient's permission or awareness. 
However, given the number of medical records generated every day, the de-identification task demands methods that are able to handle big databases. 
\cite{liu2017identification} develop an ensemble model for this task using three different methods combined with a regular expression-based sub-system to identify PHI. 
These methods were a conditional random field and two BiLSTMs. 
Hence, each method was able to identify PHI unidentified by the other two. 
Similarly, \cite{dernoncourt2017identification} introduce a de-identification system for medical records based on a BiLSTM architecture without relying on handcrafted features. 
Finally, \cite{obeid2019impact} analyze how de-identification of clinical text impacts the performance of machine learning and DL models. 
The authors take into account the tradeoff between privacy protection and electronic health records utility degradation for tasks of machine learning and information extraction by comparing the performance of machine learning classifiers and CNN models on both original and de-identified versions of a medical dataset. 
In this scenario, PHI attributes, such as dates, may improve classification outcomes but simultaneously breach private information of patients. 
Thus, the de-identification process gets rid of these attributes in order to comply with data protection regulations. 

Another source of private information that should be de-identified to safeguard privacy is user-generated content, such as e-mails, social media posts, online comments, chats, and SMS. 
This content poses privacy threats to people, locations, and entities described in the piece of text besides its writer. 
\cite{eder2020code} investigate the automatic recognition of privacy-sensitive attributes in e-mails and come up with a solution for privacy protection using BiLSTMs and sub-word embedding models for recognition and pseudonymization of identifying information of people in the messages. 
The authors conducted their experimental evaluation on two e-mail datasets for the German language, starting with the manual annotation of named entities related to attributes to be protected. 
Further, four neural models for the recognition of private entities are tested and benchmarked. 
Finally, the private entities are replaced with pseudonym forms that preserve the information type. 
Therefore, at the end of the experiments, the authors also provided a pseudonymized German e-mail corpus for additional research.   

The protection of user-generated text also influences research areas that use examples directly from data to bring examples of key findings, such as behavioral research, since those examples can lead to the re-identification of private data. 
\cite{oak2016generating} produce synthetic data from users' discourse about life-changing events on social media. 
To do so, the authors use an LSTM model fed with tweets concerning the events of birth, death, marriage, and divorce. 
The model is first trained to predict the most probable data item (character or word) given a data item used for input and, finally, used for language generation. 
The synthetic data generated by the model are compared against real tweets held out from training for the evaluation step. 
Subsequently, human annotators indicate if they thought a tweet was human- or machine-generated. 
Besides yielding realistic-looking text with similar statistical features as the training data, the authors bring insights on downstream applications, e.g., the utility for software developers who dismiss access to data in raw form for software development activities. 

\begin{sidewaystable}
\sidewaystablefn%
\begin{center}
\begin{minipage}{\textheight}
\begin{threeparttable}
\caption{Summary of data safeguarding methods}\label{tab:summary-of-ds-methods}
    \begin{tabular}{l|lllll}
    \toprule
    \textbf{Group} & \textbf{Work} & \textbf{Neural models} & \textbf{PET} & \boldmath{$T$} & \boldmath{$S$} \\
    \toprule
    \multirow{ 6}{*}{DI methods} & \cite{liu2017identification} & BiLSTM & RE & $A$ & MR \\
    {} & \cite{dernoncourt2017identification} & GloVe, word2vec, BiLSTM & PHI detection & $A$ & MR\\
    {} & \cite{obeid2019impact} & Word2vec, CNN & BoB & $D$ & MR\\
    {} & \multirow{2}{*}{\cite{eder2020code}} & 
    	GERMANER, NEURONER, & EAR & $A$ & E-mails\\
    	{} & {} & GERMAN NER, BPEMB & \\
    {} & \cite{oak2016generating} & LSTM & EAR & $A$ & Tweets \\
    \midrule
    \multirow{ 3}{*}{AM} & \cite{mosallanezhad-etal-2019-deep} & GloVe, BiLSTM + attention & DRL & $A$ & PA\\ 
    {} & \cite{sanchez2018automatic} & CNN & SA & $D$ & PA\\ 
    {} & \cite{pablos2020sensitive} & BERT & EAR & $D$ & CD \\
    \midrule 
    \multirow{ 2}{*}{DP methods} & \cite{feyisetan2020privacy} & GloVe, fasttext, BiLSTM & $d_\mathcal{X}$-privacy & $W$ & PA \\ 
    {} & \cite{fernandes2019generalised} & Word2vec, fasttext & $d_\mathcal{X}$-privacy & $A$ & PA \\ 
    {} & \cite{melamud2019towards} & Word2vec, LSTM & S-PDTP & $A$ & CN \\
    {} & \cite{lyu2020differentially} & BERT, MLP & $\epsilon$-DP & $W$ & Cloud \\
    \midrule 
    \multirow{4}{*}{RL methods} & \cite{li2018towards} & BiLSTM, word2vec, CNN & AL & $A$ & PA \\ 
    {} & \multirow{2}{*}{\cite{feyisetan2019leveraging}} & GloVe, SkipThought,
    & $d_\mathcal{X}$-privacy & $W$ & PA\\ 
    {} & {} & Fasttext, InferSent & \\
    {} & \cite{john2019disentangled} & DAE, VAE, word2vec, CNN & Auxiliary losses & $A$ & ST \\
    \bottomrule
    \end{tabular}
    \textit{Continues on next page...}
    \end{threeparttable}
    \end{minipage}
    \end{center}
    \end{sidewaystable}
    
    \begin{sidewaystable}
    \sidewaystablefn%
    \begin{center}
    \begin{minipage}{\textheight}
    \begin{threeparttable}
    \ContinuedFloat  
    \caption{Summary of data safeguarding methods (continued)}
    \begin{tabular}{l|lllll}
    \toprule
    \textbf{Group} & \textbf{Work} & \textbf{Neural models} & \textbf{PET} & \boldmath{$T$} & \boldmath{$S$} \\
    \toprule
    \multirow{ 7}{*}{Debiasing methods} & \cite{bolukbasi2016man} & Word2vec & Debiasing & $W$ & TL \\ 
    {} & \multirow{2}{*}{\cite{kaneko2019gender}} & GloVe, Hard-GloVe, & Debiasing & $W$ & TL\\
    {} & {} & GN-GloVe, autoencoders & \\
    {} & \multirow{2}{*}{\cite{font2019equalizing}} & 
        GloVe, GN-GloVe, transformer &  Debiasing & $W$ & NMT \\
    {} & {} & Hard-Debiased GN-GloVe & \\
    {} & \cite{papakyriakopoulos2020bias} & GloVe, LSTM & VST & $W$ & SM\\ 
    {} & \cite{gencoglu2020cyberbullying} & sentence-DistilBERT & FC & $W$ & OT \\ 
    \midrule 
    \multirow{ 9}{*}{AL methods} & \cite{friedrich-etal-2019-adversarial} & Fasttext, GloVe, BiLSTM-CRF & RL, DIM & $R$ & MR\\
    {} & \cite{elazar-goldberg-2018-adversarial} & LSTM, MLP & AT & $A$ & PA \\
    {} & \cite{barrett-etal-2019-adversarial} & LSTM, MLP & AT & $A$ & PA \\
    {} & \cite{xu-etal-2019-privacy} & Transformer & RL, AT & $A$ & PA \\
    {} & \cite{coavoux-etal-2018-privacy} & LSTM & RL, AT & $A$ & PA \\ 
    {} & \cite{kumar-etal-2019-topics} & BiLSTM + attention & RL, AT & $W$ & PA \\
    {} & \cite{sweeney2020reducing} & Word2vec, GloVe, LSTM, CNN & RL, AT & $A$ & PA \\ 
    \bottomrule
    \end{tabular}
    \footnotetext{\small
      \boldmath{$A$} stands for a set of protected attributes, 
      \textbf{AL} stands for `adversarial learning',
      \textbf{AM} stands for `anonymization methods',
      \textbf{AT} stands for `adversarial training',
      \textbf{BoB} stands for `best-of-breed clinical text de-identification application'~\citep{ferrandez2013bob},
      \textbf{CD} stands for `clinical data',
      \textbf{CN} stands for `clinical notes',
      \textbf{CRF} stands for `conditional random field',
      \boldmath{$D$} stands for a set of documents,
      \textbf{DAE} stands for `deterministic autoencoder',
      \textbf{DI} stands for `de-identification',
      \textbf{DIM} stands for `de-identification model',
      \textbf{DP} stands for `differential privacy',
      \textbf{DRL} stands for `deep reinforcement learning',
      \textbf{EAR} stands for `entity annotation and recognition',
      \textbf{FC} stands for `fairness constraints',
      \textbf{MR} stands for `medical records',
      \textbf{NMT} stands for `neural machine translation',
      \textbf{OT} stands for `online text', 
      \textbf{PA} stands for `private attributes',
      \textbf{PET} stands for `privacy-enhancing technology',
      \textbf{PHI} stands for `protected health information',
      \boldmath{$T$} stands for `target', 
      \boldmath{$S$} stands for `scenario', 
      \boldmath{$R$} stands for a set of records,
      \textbf{RE} stands for `regular expressions',
      \textbf{RL} stands for `representation learning',
      \textbf{SA} stands for `standard anonymization',
      \textbf{SM} stands for `social media',
      \textbf{S-PDTP} stands for `Sequential-PDTP',
      \textbf{ST} stands for `style transfer',
      \textbf{TL} stands for `transfer learning',
      \textbf{VAE} stands for `variational autoencoder',
      \textbf{VST} stands for `vector space transformation',
      \boldmath{$W$} stands for a set of target words.
    }
\end{threeparttable}
\end{minipage}
\end{center}
\end{sidewaystable}

\subsubsection{Anonymization methods}\label{subsub:anonymization-m} 

In privacy-preserving data publishing, the data must be anonymized prior to its release, aiming at averting attacks that put privacy at risk~\citep{zhou2008brief}. 
Data anonymization consists of the removal of all pieces of sensitive information, such as names, dates, and locations, that may lead to the re-identification of a document collection, followed by the replacement of this information with artificial codes (e.g., `xxxx')~\citep{sanchez2018automatic,narayanan2008robust,eder2020code}. 
Since text data is a rich source of private attributes, such as gender, ethnicity, location, and political views, text anonymization is a well-known challenge in the literature on privacy-preserving NLP. 
\cite{mosallanezhad-etal-2019-deep} propose an anonymizator based on reinforcement learning that extracts a latent representation of text and manipulates this representation to mask all private information it may hold.
Simultaneously, the utility of the representation is preserved by changing the reinforcement learning agent's loss function and assessing the quality of the embedded representation. 
\cite{sanchez2018automatic} describe a system for anonymizing images of printed documents whose text encompasses private information, such as names, addresses, dates, and financial content. 
To do so, the authors train a CNN model to strip private information out of images of invoices written in the Spanish language. 
\cite{pablos2020sensitive} also conducted experiments for the anonymization of documents in the Spanish language but focused on clinical records, which were anonymized by variations of BERT. 
One of the biggest advantages of this language model is the prospect of outstanding performances without demanding feature engineering for the specific task. 
Therefore, the authors use this model as the basis for a sequence labeling approach that detects if each token in a sentence is related to a private feature or not. 
Finally, the results also demonstrate that BERT is robust to drops in the size of the training data, solely resulting in small performance reductions.

\subsubsection{Differential privacy (DP) methods}\label{subsub:dp-based-methods}

\begin{figure}[t!]
    \centering
    \includegraphics[width=\textwidth]{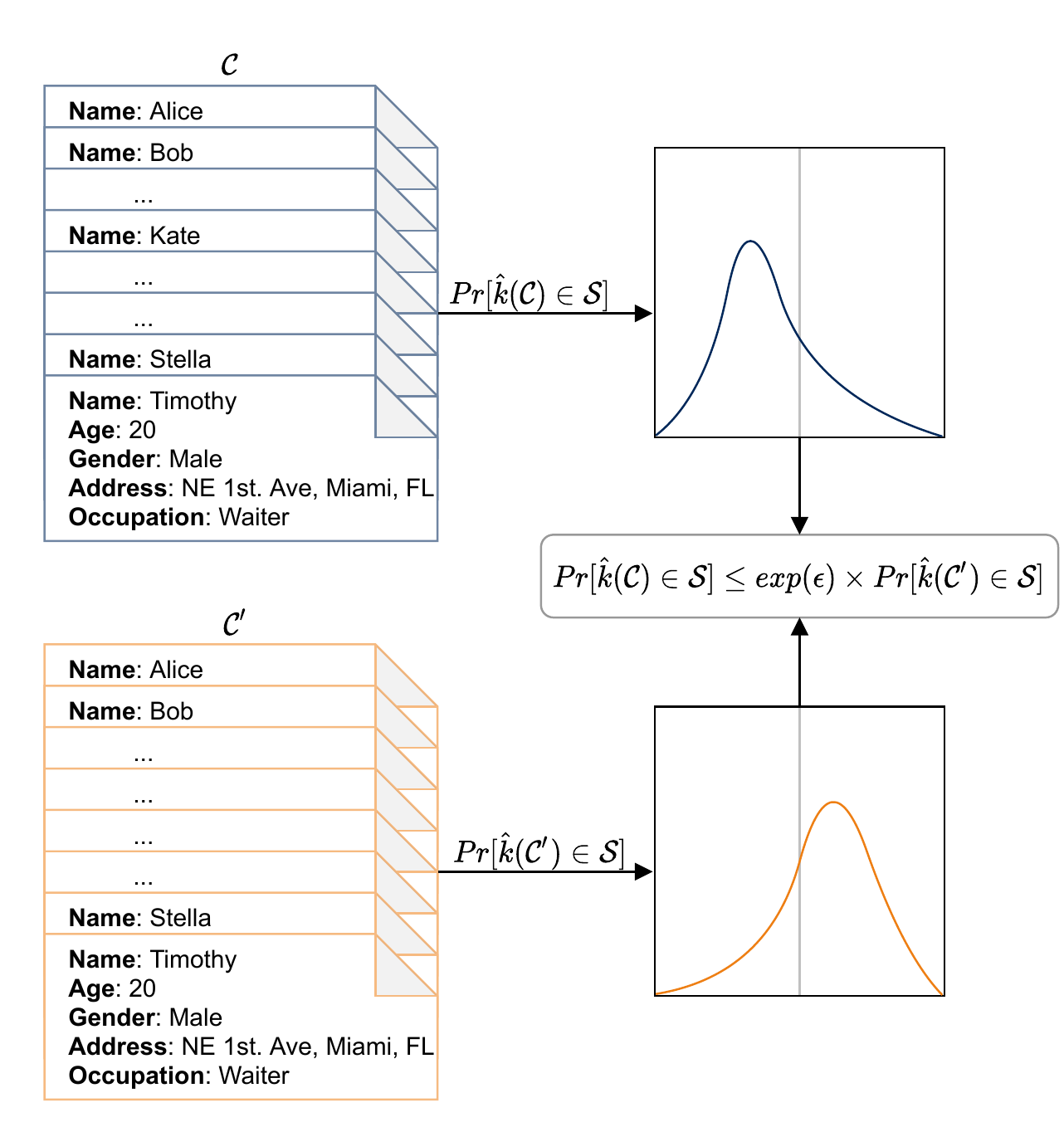}    \caption{Example of $\epsilon$-DP for two collections $\mathcal{C}$ and $\mathcal{C}'$. The collections differ by at most one element (i.e., the document for Kate).}
    \label{fig:differential-privacy}
\end{figure}

Data instances can be individually protected by DP~\citep{dwork2008differential,fernandes2019generalised}, which grants theoretical bounds to the protection of personal information in each data instance within a database, even though the aggregated statistical information of the whole database is revealed~\citep{melamud2019towards}.
DP takes into account the assumption of plausible deniability~\citep{fernandes2019generalised}, in which the output of a query may arise from a database that does not contain personal information as possible as from one that does.
Ideally, there should be no way to distinguish between these two possibilities~\citep{fernandes2019generalised}.
For instance, models trained on documents featuring personal information will provide stronger DP guarantees the less their outputs rely on individual documents in the collection~\citep{melamud2019towards}.
Formally, a randomized function $\hat{k}$ gives $\epsilon$-DP in case for two collections $\mathcal{C}$ and $\mathcal{C}'$, which differ by at most one element, and all $\mathcal{S} \subseteq Range(\hat{k})$:
\begin{equation}
    Pr[\hat{k}(\mathcal{C}) \in \mathcal{S}] \leq exp(\epsilon) \times Pr[\hat{k}(\mathcal{C}') \in \mathcal{S}].
\end{equation}
Every mechanism that satisfies this definition, which is depicted by Figure~\ref{fig:differential-privacy}, will address worries about leakages of personal information from any individual element since its inclusion or removal would not turn the output significantly more or less likely~\citep{dwork2008differential}.
On the one hand, the efficiency of DP at protecting personal information often comes along with overheads in complexity and running time~\citep{melamud2019towards}.
On the other hand, this mechanism presents noticeable flexibility in finding a balance between performance degradation and privacy budget.

DP provides theoretical bounds that preserve privacy, whereas regular data sanitization approaches may fail to demonstrate that privacy issues are formally averted. 
Further advantages of DP regard its ability to be blended with either utility task or NLP model. 
\cite{feyisetan2020privacy} perturb text data using a $d_\mathcal{X}$-privacy method, which is similar to a local DP setting when it comes to perturbing each data record independently.
The $d_\mathcal{X}$-privacy grants privacy bounds onto location data, generalizing DP across distance metrics, such as Hamming distance, Euclidean, Manhattan, and Chebyshev metrics~\citep{feyisetan2019leveraging}.
Then, calibrated noise is added to word representations from the word embedding models GloVe and fasttext.
The method takes a string $s$ of length $\text{\textbar}s\text{\textbar}$ as input and outputs a perturbed string $s'$ of same length, which is privatized by a $d_\mathcal{X}$-privacy mechanism $\mathcal{M}: \boldsymbol{\mathcal{X}} \rightarrow \boldsymbol{\mathcal{X}}$, where $\boldsymbol{\mathcal{X}} = \boldsymbol{\mathcal{D}}^l$ represents the space of all strings with length $l$ whose words are in a dictionary $\boldsymbol{\mathcal{D}}$.
Then $\mathcal{M}$ computes the embedding $\phi(w)$ for each word $w \in s$, adding calibrated random noise $\mathcal{N}$ to yield a perturbed embedding $\phi'=\phi(w)+\mathcal{N}$.
Later on, $w$ is replaced with a word $w'$ whose embedding is the closest to $\phi'$, according to a Euclidean metric $d$.
The authors consider that a randomized algorithm satisfies DP if its output distribution is similar to those when the algorithm is applied to two adjacent databases.
They argue that the notion of similarity is managed by a parameter $\epsilon$, which governs the extent privacy is preserved from full privacy, when it assumes the value of 0, to null privacy when it approaches $\infty$.
For instance, $\mathcal{N}$ is sampled from a distribution $z$ with density $\mathcal{P}_{\mathcal{N}}(z) \propto exp(-\epsilon ||z||)$.
By varying the values of $\epsilon$, the tradeoff between privacy and utility is demonstrated. 

\cite{fernandes2019generalised} obfuscate the writing style of texts without losing content, addressing the threats of unintended authorship identification. 
The authors assume that an author's attributes can be predicted from the writing style, such as identity, age, gender, and mother tongue.
A DP mechanism inspired by $d_\mathcal{X}$-privacy then perturbs bag-of-words representations of texts, preserving topic classification but disturbing clues that lead to authorship information.
Firstly, a randomized function $\hat{k}$ receives $b,b'$ bag-of-words as inputs and outputs noisy bag-of-words $\hat{k}(b),\hat{k}(b')$.
If $b$ and $b'$ are classified as similar in topic, their perturbed versions $\hat{k}(b),\hat{k}(b')$ should also be similar to each other, depending on the privacy budget $\epsilon$, regardless of authorship.
Finally, $\hat{k}(b)$ should be distributed in agreement with a Laplace probability density function calculated according to a metric for semantic similarities, such as the Earth Mover's distance.

Clinical notes are unstructured text data that encompass information input by doctors, nurses, or other patient care staff members.  
This kind of data requires de-identification before sharing activities, so patient privacy is not put at risk.
\cite{melamud2019towards} propose a method relying on DP to generate synthetic clinical notes, which can be safely shared.
The authors define a setup for the task in three steps.
Firstly, real de-identified datasets of clinical notes are used to train neural models that output synthetic notes.
Secondly, privacy measures assess the privacy safeguarding properties of the synthetic notes.
Finally, the utility of the generated notes is estimated using benchmarks.

DP also helps produce fair text representations as to demographic attributes.
\cite{lyu2020differentially} provide a framework for learning deferentially private representations of texts, which masks private content words, whereas guarantees fairness by reducing discrimination towards age, gender, and five `person' entities.
The framework assumes data exchange between client and server parties and takes into account the threat of an eavesdropping attack which discloses private information from text representations yielded by a feature extractor from the client's side and sent to a classifier on the server's side.
In order to protect the text representation from the eavesdropper, the training algorithm adds noise to the representations generated by the feature extractor.
The same level of noise is added for both training and test phases, escalating the model robustness to noisy representations.
Later on, the feature extractor $f''$ is also given to the client.
Another mechanism proposed by the authors for their framework consists of a word dropout which masks words before the DP noise injection.
Let $x_i$ be a sensitive input composed of $g$ words, and $\boldsymbol{\vec{I}}$ a dropout vector $\boldsymbol{\vec{I}} \in \{0,1\}^g$.
Therefore, dropout will be a word-wise multiplication of $x_i$ with $\boldsymbol{\vec{I}}$.
The number of zeroes in $\boldsymbol{\vec{I}}$ is defined by the dropout rate $\mu$ as $g \cdot \mu$.
Additionally, combining word dropout with the $\epsilon$-differentially private mechanism is useful to lower the privacy budget without drastically degrading the inference performance.

\subsubsection{Representation learning methods}\label{subsub:repr-learning-m}

\begin{figure}[h!]
    \centering
    \includegraphics[width=\textwidth]{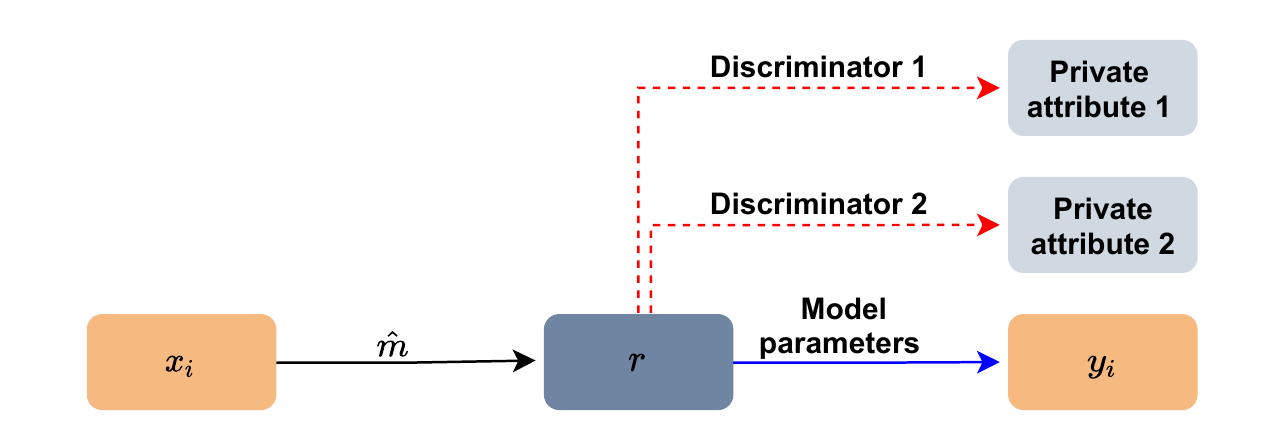}
    \caption{Adversarial learning for representation learning \citep{li2018towards}. The method takes an input instance $x_i$ and outputs a label $y_i$ from the hidden representation $r$. The dashed red and blue lines represent the adversarial and standard losses.}
    \label{fig:adversarial-learning-for-rl}
\end{figure}

Representation learning allows data to be represented as an information format conveniently used by DL models, whereas the efforts for manual feature engineering are no longer required~\citep{bengio2013representation}.
Even text representations may still enclose private information and, for this reason, be prone to privacy threats.
To address privacy issues, \cite{li2018towards} train deep neural models using adversarial learning to yield unbiased representations and safeguard individuals' private information, namely age, gender, and location.
The proposed method takes inputs $x$ to compute a hidden representation $r$, which is used to form the parametrization of a model $\hat{m}$ for predicting a target $y$, as depicted by Figure~\ref{fig:adversarial-learning-for-rl}.
During the training process, a loss function like cross-entropy is minimized to determine the model parameters $\theta_m$.
AL is based on learning a discriminator model $\hat{q}$ and $\hat{m}$ jointly.
The discriminator model tries to predict a private attribute from each instance of $r$, so that the adversarial training can be seen as joint estimating the parameters $\theta_m$ and $\theta_q$ for $\hat{m}$ and $\hat{q}$ respectively.
In order to safeguard privacy, $r$ must lead to efficient predictions of $y$ and deficient representations of the private attributes.
Therefore, the objective function $\hat{\theta}$ of the method has the form
\begin{equation}
    \hat{\theta} = \min_{\theta_m} \max_{\{{\theta_q}_i\}^N_{i=1}} \mathcal{E}(\hat{y}(x,\theta_m),y) - \sum^N_{i=1} (
    \lambda_i \cdot \mathcal{E}(\hat{a}(x,{\theta_c}_i),a_i)),
\end{equation}
in which $\mathcal{E}$ represents the cross-entropy function, $\hat{y}$ denotes the predicted labels, $a$ denotes a private attribute, $\hat{a}$ denotes the prediction of the discriminator model, and $N$ is the number of private attributes.
In the end, the learned text representations can be transferred for downstream applications as well as the discriminator.

\cite{feyisetan2019leveraging} provide privacy preservation guarantees by perturbing word representations in Hyperbolic space, satisfying $d_{\mathcal{X}}$-privacy.
The proposed method first transforms target words into vector representations.
Second, these vectors are perturbed with noise sampled from the same Hyperbolic space where the vectors lie.
The extent of this noise added to the vectors is proportional to the guarantees of privacy preservation.
Finally, a post-processing step maps the vectors perturbed by noise into their closest words in the embedding vocabulary, hence preserving the semantic information.
For instance, a query like `how is the weather like in Los Angeles right now?' would have the location name replaced with the term `city', based on similarity.
Therefore, both user's query intent and privacy are preserved without semantic losses in the sense of the query.

Model interpretability can also be enhanced by representation learning at the pace of safeguarding private attributes.
\cite{john2019disentangled} address the problem of disentangling the latent feature space of a neural model for text generation.
The authors follow the premise that neural networks yield latent representations for the original feature set, which are not interpretable and do not present their meaning explicitly.
So they came up with a method that is based on an autoencoder for encoding sentences into a latent space representation and learning to reconstruct these sentences to their original form.
The representations produced by this method are then disentangled, i.e., divided into two parts with regard to different features (style and content) by a combination of adversarial and multi-task objectives.
Sentiment associated with a sentence is considered as its style.
Finally, the authors designed adversary losses to force the separation between the latent spaces for both style and content features, which can be used for text style transfer tasks later on.

\subsubsection{Debiasing methods}\label{subsub:debiasing-m}

Recent studies have brought evidence that pre-trained embedding models, ranging from word2vec to BERT, exhibit human biases towards demographic attributes, such as gender, race, nationality, and location.
These biases play an influential role in downstream applications, which will be prone to making unfair decisions.
NLP datasets may also encompass biases that weaken model generalization abilities when applied to unbiased datasets for transfer learning.
For the sake of fairness in algorithmic decisions, debiasing consists of removing or lessening human biases that have the potential to compromise model decisions in NLP tasks. 
Among the types of bias, the one towards gender is broadly studied, mostly due to the popularity of word embedding models and the need for gender equality in systems relying on embeddings, which influence the everyday life of a huge number of people.

\cite{bolukbasi2016man} show that word2vec embeddings trained on the Google News dataset noticeably feature gender stereotypes like the association between the words `receptionist' and `female'.
In order to get rid of such discriminatory associations while preserving the embedding power of solving analogy tasks and clustering similar concepts, the authors come up with an approach consisting of two phases.
Firstly, evaluating whether gender stereotypes are present in vectors for occupation words, using crowd workers in the validation process, and generating analogy tasks where a word pair like \{he, she\} is used to predict new pair of words whose first term should be related to `he', and the second one should be related to `she'. 
So the results of this task are validated by the human workers to check if those analogies make sense and express gender stereotypes. 
A gender subspace is captured by the top component of a principal component analysis computed on ten gender pairs of difference vectors.
Later on, the authors identify words that should be neutral with regard to gender, taking a set of 327 occupation terms.
Finally, two debiasing algorithms are developed to two options: Soften or Neutralize and Equalize gender bias. 
Neutralize secures the gender-neutral words to be zero in the gender subspace, whereas Equalize balances the word sets outside the subspace and keeps the gender-neutral ones equidistant to both equality sets. 
For instance, given the equality sets \{grandmother, grandfather\} and \{guy, gal\}, introduced by the authors, the representation for the term `babysit' should be equidistant to the terms in both equality sets after equalization.
However, this representation should also be closer to those in the first equality set for the purpose of preserving the semantic relatedness of the terms.

Similarly, \cite{kaneko2019gender} propose a debiasing method for pre-trained word embeddings that is able to differentiate between non-discriminatory gender information and discriminatory gender bias. 
The authors argue that associations between words like `bikini' and feminine nouns, or `beard' and masculine nouns, would be expected and, then, capable of enhancing applications like recommender systems without prompting unfair model outcomes.
On the other hand, profession titles such as `doctor', `developer', `plumber', and `professor' have frequently been stereotypically male-biased, but `nurse', `homemaker', `babysitter', and `secretary' have been stereotypically female-based.
Therefore, they consider four information types, namely, feminine, masculine, gender-neutral, and stereotypical, to get rid of biases from stereotypical words, whereas gender information in feminine and masculine words and neutrality in gender neutral-words are maintained.
Given a feminine regressor $\hat{u}: \mathbb{R}^e \rightarrow [0,1]$, which has $\theta_u$ parameters for predicting the extent of femininity the word $w$ presents.
In this sense, highly feminine words are assigned to femininity values nearing 1.
In a similar manner, a masculine regressor $\hat{v}: \mathbb{R}^e \rightarrow [0,1]$ with parameters $\theta_v$ estimates the masculinity degree of $w$.
Therefore, the debiasing function will be learned as the encoder component of an autoencoder $E: \mathbb{R}^d \rightarrow \mathbb{R}^e$ with parameters $\theta_E$, whereas the decoder component is defined as $\mathcal{Z}: \mathbb{R}^d \rightarrow \mathbb{R}^e$
with parameters $\theta_\mathcal{Z}$.
The number of dimensions of the original vector space is denoted by $d$, whereas the number of dimensions of the debiased vector space is given by $e$.

Word embedding models can pass gender biases on from training corpora to downstream applications. 
\cite{font2019equalizing} come up with a method to equalize gender bias in the task of neural machine translation using these word representations. The authors detect biases toward terms originally in English, which are translated into masculine forms in Spanish. 
For instance, the word `friend' would be translated into a masculine Spanish word if it came along in a sentence with the term `doctor', whereas it would be translated into the feminine form in case it was used in the same sentence as the term `nurse'.
The authors use a state-of-the-art transformer model for neural machine translation input with word embedding yielded by three embedding models, namely, GloVe, GN-GloVe, and Hard-Debiased GloVe.
Additionally, they compare two different scenarios, featuring no pre-trained embeddings or using pre-trained embeddings from the same corpus on which the model is trained.
The transformer models for the second scenario have three distinct cases regarding the use of pre-trained embeddings: only on the model encoder's side, only on the model decoder's side, or on both model sides.
The experimental results show that debiased embedding models do not slash the translation performance.

Further applications for debiasing approaches include the detection of biases in text data with content related to politics and bullying online. 
\cite{papakyriakopoulos2020bias} develop a method for bias detection in the German language and compare bias in embeddings from Wikipedia and political-social data, proving that biases are diffused into machine learning models. 
The authors test two methodologies to debias word embedding yielded by GloVe, and employ biased word representations to detect biases in new data samples.
\cite{gencoglu2020cyberbullying} proposes a debiasing model for cyberbullying detection on different online media platforms, employing fairness constraints in the training step.
The author conducts experiments on gender bias, language bias, date bias (e.g., drop in performance on recently created insult terms), and bias towards religion, race, and nationality.
A sentence-DistilBERT is used to extract representations for posts and comments in the datasets.
The objective function of the neural model was adapted to implement fairness measures.

\subsubsection{Adversarial learning methods}\label{subsub:adv-le-m}

Health data, like patient notes, is a widely known source of protected attributes to be taken into obliviousness by adversarial learning. 
Automatic de-identification approaches for PHI data are costly since massive datasets for model training are barely available due to regulations that hinder the sharing of medical records. 
\cite{friedrich-etal-2019-adversarial} present a method to yield shareable representations of medical text, without putting privacy at risk, by PHI removal that does not demand manual pseudonymization efforts. 
Firstly, adversarial learning-based word representations are learned from publicly available datasets and shared among medical institutions afterward. 
Secondly, the medical institutions convert their PHI raw data into these representations (e.g., a vector space) that will be pulled into a new dataset for de-identification, avoiding the leakage of any protected attributes.
Finally, the approach is argued to provide defenses against plain-text and model inversion attacks. 

Recent approaches of adversarial learning to safeguard training data include the removal of demographic attributes, such as gender, ethnicity, age, location, nationality, social status, and education level. 
Language models that encode these attributes are prone to a series of privacy issues that compromise their safety and fairness. 
\cite{elazar-goldberg-2018-adversarial} demonstrate that demographic information can be encoded by neural network-based classifiers. 
Later on, an attacker network that predicts protected attributes above chance level is used to retrieve the demographic ones from the latent representation of the classification models. 
The authors use an adversarial component in order to pull out the attributes of ethnicity (race), gender, and age from tweets collected for the tasks of binary tweet-mention prediction and binary emoji-based sentiment prediction. 
In their configuration, a classifier was trained to predict the protected attributes alongside a one-layer LSTM meant to encode a sequence of tokens and undermine the classifier, namely an MLP. 
Therefore, the learned representations have their information with regards to the tasks maximized, while it is minimized to the protected attributes. 
As a result, the adversarial learning method demonstrates efficiency in avoiding leakages but fails to completely remove protected attributes from the text. 

\cite{barrett-etal-2019-adversarial} revisit the experiments of \cite{elazar-goldberg-2018-adversarial}, analyzing correlations between the yielded representations on the models and the demographic attributes of age and gender.
They introduce three correlation types.
First, prevalent correlation arises from features associated with gender in most contexts, like in the sentence fragments including expressions like `as a mother', `my girlfriend', `as a guy', `the mailman', etc.
Second, sample-specific correlation is related to features tied up with different demographic attributes depending on the domain or sample, such as the word `bling' related to different ranges of ages if it is used to describe jewelry items, movies, or rap songs.
Finally, accidental correlation demonstrates the relationship between text features and protected attributes in a particular dataset, although their uncommon relation. 
The experimental evaluation suggests that the model relies on spurious or accidental correlations limited to a specific sample of data since they fail on new data samples or domains. 

\cite{xu-etal-2019-privacy} come up with a privacy-aware text rewriting method for obfuscating sensitive information prior to data release to promote fair decisions which do not take into account demographic attributes. 
The authors defined this task as protecting the sensitive information of data providers by text rephrasing, which lessens the leakage of protected attributes, maintains the semantics of the original text, and preserves the grammatical fluency. 
Formally, this task assumes a set of inputs $X = \{x_1,x_2,\dots,x_n\}$, in which each input $x_i$ represents a word sequence $\langle w_1,w_2,\dots,w_n \rangle$ associated with a sensitive attribute $a \in A$. 
It aims at finding a function $\hat{f}(x_i): x_i \to y_i$, which translates $x_i$ into a different word sequence $y_i \in Y$ that halts an attacker from detecting the values of $a$ given the translated text. 
Since there is no parallel corpus to recognize patterns of privacy-preserving text rewriting, the authors approached the task as a monolingual machine translation problem, using back-translation. 
Here, a text is translated from English to French and later back to English.
This task aims at minimizing the reconstruction loss between $\hat{f}(x_i)$ and $y_i$ along with the risk loss towards privacy $R(X,Y,A)$. 
Two different obfuscation methods are proposed: one based on adversarial learning and the other based on fairness risk measurement. 
Adversarial learning is employed to yield representations that ease reconstructing the input texts while slashing the prediction of sensitive attributes by a linear classifier, which receives the yielded latent representations of the word sequences as inputs. 
In other words, the text reconstruction performance is maximized, whereas that of the linear classifier is minimized.
On the other hand, fairness risk measurement concerns the discrepancy between the privacy-preserving translator and a subgroup translator that relies on a sensitive group attribute $a$. 
The lower the discrepancy, the better the obfuscation.
A transformer~\citep{vaswani2017attention} architecture is used for translation in the experimental evaluation, which aims at confounding the attributes of gender, race, and political leaning.
Additionally, the evaluation of the leakage risk is estimated by logistic regression with L2 regularization~\citep{pedregosa2011scikit}. 
Finally, the authors propose metrics for privacy-aware text rewriting to assure the requirements of fluency, obfuscation of sensitive information, and semantic relevance.
Of the two proposed methods for the task, the one based on fairness risk preserves fluency and relevance to a greater degree than the adversarial one.

Other NLP tasks, such as sentiment analysis and topic classification, also pose privacy risks regarding adversarial attacks that have the potential to recover sensitive information from language representations. \cite{coavoux-etal-2018-privacy} study this kind of privacy attack, propose privacy measures that gauge the leakage risk of private attributes from hidden representations learned by neural networks, discuss the privacy-utility tradeoff, and propose safeguarding methods by adding terms to the objective functions of the models. Both tasks of sentiment analysis and topic classification are approached in the experiments by an LSTM model. 
Moreover, \cite{kumar-etal-2019-topics} bring evidence that language classification models can learn topical features which are confounds for an inference task of native language identification. Hence, the authors propose an adversarial learning setting for representing the latent confounds.
At the same time, a BiLSTM model with attention obfuscates these features by predicting them along with the actual labels for each input. This method is argued to be less prone to using a smaller amount of information related to confounds, besides better generalization abilities, and enhanced for learning writing style features instead of content ones.

\cite{sweeney2020reducing} use adversarial learning to remove correlations between demographic attributes and sentiments in word vectors, referring to this problem as sentiment bias.  
For the authors, word vectors for demographic identity terms, such as those related to nationality, religion, gender, and names, should retain neutrality with regard to sentiments. 
In the experiments, they, firstly, retrieve the vectors for positive words from the Sentiment Lexicon dataset~\citep{hu2004mining} to create a matrix for taking the most significant component from the principal component analysis. 
The same process is done for the negative words in the dataset afterward since there is no pre-defined pairwise mapping between positive and negative terms in a similar fashion as the data for tackling gender bias.
Another reason for this two-step process relates to the semantics for drawing differences between positive and negative perceptions, which is considerably looser than that for gender.
Secondly, the signed difference between both negative and positive components is taken and named the directional sentiment vector.
Further, the sentiment polarity of all remaining vectors is assessed by projecting these against the directional sentiment ones.
In the following stage, a classification model is used to check if the directional sentiment vectors are able to hold sentiment polarity.
Adversarial learning computes two distinct objectives. 
The first finds the least square distance between the input word vector and its debiased version.
Simultaneously, the adversarial one predicts the sentiment polarity based on the input vector.  
Finally, the embedding models word2vec and GloVe are debiased and later tested for the downstream tasks of sentiment valence (intensity) regression and toxicity classification, leading to semantic preservation and fair decisions.

\begin{figure}[h!]
    \centering
    \includegraphics[width=\textwidth]{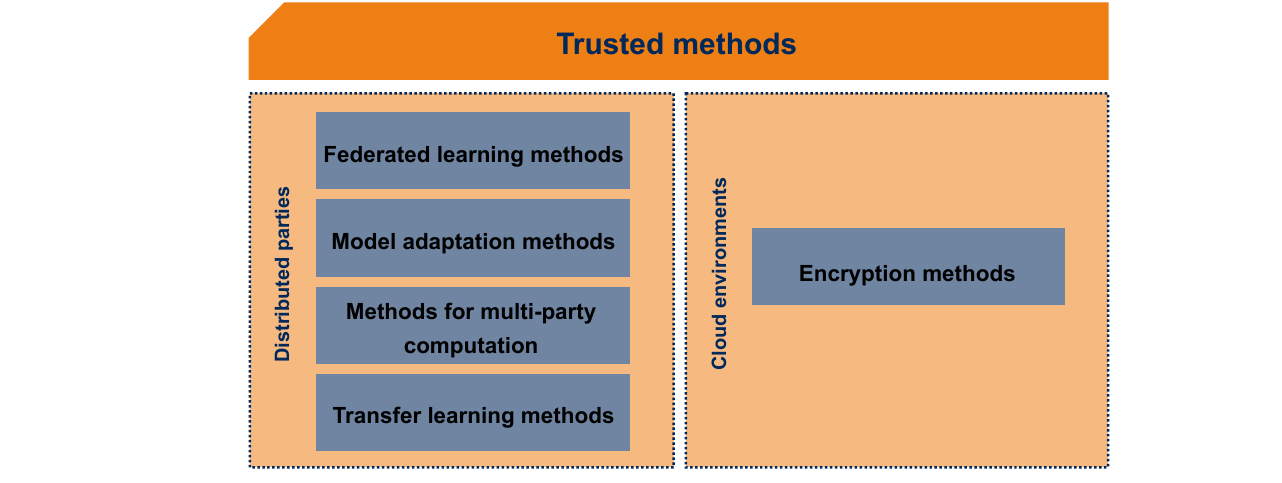}
    \caption{Sub-categories and groups for trusted methods}
    \label{fig:trusted-methods}
\end{figure}

\subsection{Trusted methods}\label{subsub:trusted-methods}

When DL models are designed for learning over data or untrusted computation scenarios, trusted methods appear as solutions.
We have identified trusted methods for scenarios involving cloud environments and distributed parties, as depicted by Figure~\ref{fig:trusted-methods} and summarized in Table~\ref{tab:summary-of-trusted-methods}.
The sub-categories and groups of trusted methods are populated with well-known PETs, often implemented in real-time systems, such as federated models, transfer learning, and encryption.

\subsubsection{Federated learning (FL) methods}\label{subsub:federated-learning-methods}

\begin{figure}[h!]
    \centering
    \includegraphics[width=\textwidth]{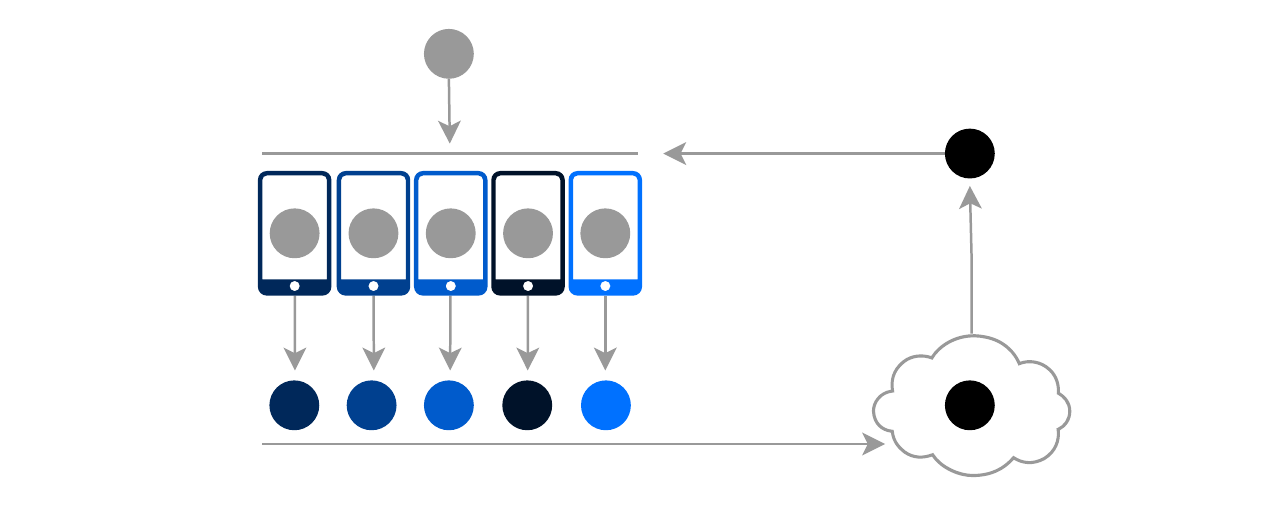}
    \caption{Federated learning setting for mobile devices~\citep{mcmahan2017federatedblog}. From the initial global model at the top, each mobile device locally computes its model update and sends it to the server (lower right corner), which aggregates all the received local updates and distributes the updated global model to the mobile devices.}
    \label{fig:federated-learning}
\end{figure}

FL is a technique proposed by Google in 2016~\citep{konevcny2016federated,pmlr-v54-mcmahan17a} which enables centralized models to train on data distributed over a huge number clients~\citep{konevcny2016federated}.
For instance, a model for spell-checking in virtual keyboards can be hosted on a central server and trained in a decentralized manner across smartphones, without any data exchanges between the client devices and the server or between client devices.
FL does not allow data to ever leave its owner's device~\citep{chen2019federated}.
This learning paradigm consists of training rounds where every client updates the model it receives from the central server with computations on its local data and passes this update on to the server, which computes an improved global model upon the aggregation of all client-side updates~\citep{konevcny2016federated}.
Figure~\ref{fig:federated-learning} illustrates an example of an FL setting for mobile devices.

\begin{sidewaystable}
\sidewaystablefn%
\begin{center}
\begin{minipage}{\textheight}
\begin{threeparttable}
    \caption{Summary of trusted methods}\label{tab:summary-of-trusted-methods}
    \begin{tabular}{l|lllll}
    \toprule
    \textbf{Group} & \textbf{Work} & \textbf{Neural models} & \textbf{PET} & \boldmath{$T$} & \boldmath{$S$} \\
    \toprule
    \multirow{6}{*}{FL methods} & \cite{chen2019federated} & CIFG LSTM, GLSTM & FL & $W$ & VK \\
    {} & \cite{mcmahan2018learning} & LSTM &
         $(\epsilon,\delta)$-DP,
         FL & $W$ & VK \\ 
    {} & \cite{hard2018federated} & CIFG LSTM & FL & $W$ & VK\\
    {} & \cite{huang2020texthide} & BERT, MLP, RoBERTa & TextHide & $W$ & DC \\
    {} & \cite{zhuetalal2020empiricalstudies} & TextCNN &$(\epsilon,\delta)$-DP, FL & $W$ & DC \\
    {} & \cite{qi-etal-2020-privacy} & GloVe, CNN, GRU & $(\epsilon,\delta)$-DP, FL & $W$ & NR\\
    \midrule 
    \multirow{ 2}{*}{MA methods} & \cite{clinchant2016transductive} & MD autoencoders & Transductive adaptation & $D$ & PI\\
    {} & \cite{zhao2018learning} & GN-GloVe & Private training & $W$ & PA \\
    \midrule 
    Methods for MPC & \cite{feng2020securenlp} & Seq2seq + attention & MPC protocols & $W$ & MSD \\
    \midrule
    \multirow{ 3}{*}{TL methods} & \cite{alawad2020privacy} & MT-CNN & Vocabulary restriction & $W$ & CT \\
    {} & \cite{martinelli2020nlpijcnn} & Fasttext & SID & $D$ & PI  \\
    {} & \cite{hu2020privnet} & PrivNet & Adapted loss & $D$ & PI\\
    \midrule
    \multirow{ 2}{*}{Encryption Methods} & \cite{dai2019efficient} & Doc2Vec & Searchable encryption & $D$ & Cloud \\
    {} & \cite{liu2020mitigating} & LSTM & Searchable encryption & $K$ & Cloud \\
    \bottomrule
    \end{tabular}
    \footnotetext{\small
      \textbf{CT} stands for `collaborative training',
      \boldmath{$D$} stands for a set of documents,
      \textbf{DC} stands for `document collection',
      \textbf{DP} stands for `differential privacy',
      \textbf{FL} stands for `federated learning',
      \textbf{GRU} stands for `Gated recurrent units',
      \boldmath{$K$} stands for a set of keywords,
      \textbf{MA} stands for `model adaptation',
      \textbf{MD} stands for `marginalized denoising',
      \textbf{MPC} stands for `multi-party computation',
      \textbf{MSD} stands for `multi-source data',
      \textbf{NR} stands for `news recommendation',
      \textbf{PA} stands for `private attributes',
      \textbf{PET} stands for `privacy-enhancing technology',
      \textbf{PI} stands for `private inference',
      \boldmath{$S$} stands for `scenario', 
      \textbf{SID} stands for `sensitive information detection'
      \boldmath{$T$} stands for `target', 
      \textbf{TL} stands for `transfer learning',
      \textbf{VK} stands for `virtual keyboards',
      \boldmath{$W$} stands for a set of target words.}
\end{threeparttable}
\end{minipage}
\end{center}
\end{sidewaystable}

Moreover, FL is hoped for text datasets that encompass user-generated text, financial transactions, medical records, personal preferences, trajectories, and so on~\citep{zhuetalal2020empiricalstudies}.
Formally, FL learns a model whose parameters are held by a matrix $\boldsymbol{M}$ from data samples stored by many different clients, sharing the current model $\boldsymbol{M}_{t'}$ with a set $T$ of $n'$ clients, which updates the model with their local data at each training round $t'\geq 0$~\citep{konevcny2016federated}.
Each selected client $T_i$ sends its update $\boldsymbol{H}^{T_i}_{t'} := \boldsymbol{M}^{T_i}_{t'} - \boldsymbol{M}_{t'}$ back to the server, which aggregates all the client-side updates and comes up with the global update in the form:
\begin{equation}
    \begin{split}
    \boldsymbol{M}_{t'+1} = \boldsymbol{M}_{t'} + \eta_{t'} \boldsymbol{H}_{t'}, \\
    \boldsymbol{H}_{t'} := \frac{1}{n'} \sum_{T_i \in T} \boldsymbol{H}^{T_i}_{t'}
    \end{split}
\end{equation}
in which $\boldsymbol{M}^{T_1}_{t'},\boldsymbol{M}^{T_2}_{t'},\dots,\boldsymbol{M}^{T_{n'}}_{t'}$ are the updated local models, and $\eta_{t'}$ is the learning rate~\citep{konevcny2016federated}.
FL is especially suitable for scenarios in which the client devices may not feature a high-speed bandwidth, such as smartphones or internet of things sensors, but reducing the communication cost, whereas preserving data privacy is still an open challenge for federated models.

FL methods became popular for smartphone applications, like virtual keyboards.
\cite{chen2019federated} propose a federated training for an RNN  using the FederatedAveraging~\citep{pmlr-v54-mcmahan17a} algorithm and approximate this server-side model with an n-gram language model, which allows faster inference on the client's side.
For reasons concerning memory and latency, language models for virtual keyboards are based on n-grams and do not surpass ten megabytes in size.
Given previously typed words $w_1, w_2, \dots,w_{n-1}$, the language model will assign a probability to predict the next word as
\begin{equation}
    Pr(w_n\mid	w_{n-1},\dots,w_1).
\end{equation}
The authors follow the assumption that n-gram language models are Markovian distributions of order $o-1$, in which $o$ represents the order of the n-gram, in the form
\begin{equation}
    Pr(w_n\mid w_{n-1},\dots,w_{n-o+1}).
\end{equation}
FederatedAveraging collects unigrams on each client device, returning counting statistics to the server instead of gradients.
A unigram distribution is counted based on a white-list vocabulary.
Later on, the unigram part of an n-gram is replaced with its distribution, producing the final language model.
Then, a modified SampleApprox. algorithm~\citep{suresh2019distilling} approximates the RNN.
Tests are conducted on text data from virtual keyboards for two languages, namely American English and Brazilian Portuguese.
Finally, the results demonstrate that federated n-gram models present high quality for faster inference than server-based models, with the advantage of keeping private user-generated data on their owner's smartphone.

Additional methods for next word prediction in virtual keyboards include \cite{mcmahan2018learning}, which apply DP alongside FL training.
The authors present a version of the FederatedAveraging algorithm that is perturbed by noise, satisfying user-adjacent DP~\citep{abadi2016deep}.
This approach presents strong privacy guarantees without losses of utility since it does not decrease the performance of the target task drastically.
\cite{hard2018federated} also explore this task, comparing a server-based training using stochastic gradient descent against client-side training that uses the FederatedAveraging algorithm.
The results demonstrate the efficiency of the federated training alongside gains in prediction recall.

NLP tasks also feature privacy risks that should be tackled by NLP methods, such as eavesdropping attacks or the inversion of general-purpose language models like BERT.
\cite{huang2020texthide} create the framework TextHide for addressing these challenges in natural language understanding by protecting the training data privacy at a minimum cost concerning both training time and utility performance.
This framework requires each client in a federated training setting to add a simple encryption step to hide the BERT representations of its stored text.
Therefore, an attacker would have to pay a huge computational cost to break the encryption scheme and recover the training samples from the model.

Recent challenges in FL include the protection of the parties that are involved in the decentralized model training for sentence intent classification~\citep{li2008learning}.
Taking into account the numerous applications that rely on this task, such as review categorization and intelligent customer services, \cite{zhuetalal2020empiricalstudies} show how to adapt the NLP model TextCNN~\citep{kim-2014-convolutional} for federated training, adding Gaussian noise to the model gradients before updating the model parameters.
TextCNN is built upon a CNN for classification tasks at the sentence level.
For FL model training, the central server receives the gradients, computed on the local data stored by the clients, at the end of each epoch.
Firstly, given the values for the parameters on the central server's model, each client samples its data in a batch by batch manner.
Secondly, the local parameters are updated based on the gradients computed for each sample batch.
At the end of the iterations over all the sample batches, the cumulative difference of the parameter values is then sent to the central server for cross-client aggregation and updating the parameters of the global model.
Before sending the locally computed gradients to the server, the privacy accountant is computed, and controlled noise is added to these gradients.
This procedure protects the per-sample privacy of each client involved in the federated training.
Therefore, each client controls its privacy budget instead of the central model and stops its updates to the server once the privacy threshold is reached.
The authors argue that their method is convenient for scenarios in which the clients trust the communication channels, e.g., by encryption.
However, the central server is an honest-but-curious one.
Sensitive information, for this reason, should not be exposed to the server.

FL can also be blended with local DP for news recommendations without storing user data in a central server.
\cite{qi-etal-2020-privacy} approach this task, proposing a framework that first keeps a local copy of the news recommendation model on each user's device and computes gradients locally based on the behavior of the users in their devices.
Second, the gradients of a randomly selected group of users are uploaded to the server for aggregation and subsequent updating of the global model that it stores.
Prior to the gradients' upload, the framework applies local DP to perturb implicit private information they may hold.
Finally, the updated global model is shared with the user devices, which will compute their updates locally.

\subsubsection{Model adaptation methods}\label{subsub:model-adaptation}

Data is often a target of legal provisions and technical constraints that require the adaptation of machine learning and DL methods to meet privacy preservation requirements that may lead to penalties in case of noncompliance. 
For instance, some domains present huge amounts of data alongside high costs to acquire labels to perform classification tasks. 
In such backdrops, domain adaptation (DA) methodologies can be employed but sometimes result in privacy issues.  
\cite{clinchant2016transductive} apply a marginalized denoising autoencoder in a transductive manner on text data that suffered from domain shift during DA. 
When the source and target domains differ, performance downsides on the latter domain can be noticed, especially if there is no known label information. 
Therefore, this autoencoder aims to minimize the following reconstruction loss:
\begin{equation}
    \mathcal{L}(\digamma) = \sum_{i=1}^{n} \sum_{c=1}^{C} ||x - \digamma\tilde{x}_{ic}||^2,
\end{equation}
where $\digamma$ is a linear mapping between the two domains, $n$ is the number of inputs $x$, $\tilde{x}$ denotes an input that was corrupted $C$ times by random dropout of features. 
The transductive adaptation mechanism proposed by the authors consists of leveraging the feature space for the target data using the class scores generated by the model trained on the source domain, exploiting correlations, and protecting the source data's content. Experiments were conducted on two DA datasets and showed the effectiveness of their adapted autoencoder over a standard classifier baseline.

Word embeddings are also prone to hazards from sensitive data, such as biases that play a role in discriminatory decisions made by application systems like resume filters. Hence adaptation requirements for these neural models are also approached in the literature. 
\cite{zhao2018learning} adapt GloVe embeddings to cope with protected attributes during the training pace, yielding gender-neutral word representations without undermining their functionality. 
The proposed method safeguards sensitive attributes in specific dimensions that can be easily stripped away afterward. 
Masculine and feminine words are used as seeds for restricting gender information in the learned vector spaces, as demonstrated by a list of profession titles that are gender-neutral by definition.
However, their original GloVe representations exhibited gender stereotypes. 
Cosine similarity is used to quantify the gender tendency of each word vector $\boldsymbol{\vec{w}}$ and the gender direction $v'$ in the form
\begin{equation}
    \frac{\boldsymbol{\vec{w}} \cdot v'}{||\boldsymbol{\vec{w}}|| \ ||v'||}.
\end{equation}
The gender direction variable $v'$ averages the differences in the representations for feminine words and their masculine counterparts in a predefined set of word pairs $\Omega'$ as
\begin{equation}
    v' = \frac{1}{||\Omega'||} \sum_{(w_v,w_u) \in \Omega'} (\boldsymbol{\vec{w_v}}-\boldsymbol{\vec{w_u}}),
\end{equation}
where $w_v$ and $w_u$ are respectively a masculine and a feminine word for which the vector representations $\boldsymbol{\vec{w_v}}$ and $\boldsymbol{\vec{w_u}}$ were generated. 
A smaller similarity to gender direction suggests diminished gender information in the vector space. 
This method enables unbiased representations of sensitive words as to binary genders and has the potential to be extended to additional features such as sentiments and demographic information.

\subsubsection{Methods for multi-party computation (MPC)}\label{subsub:methods-for-mpc}

\begin{figure}
    \centering
    \includegraphics[width=\textwidth]{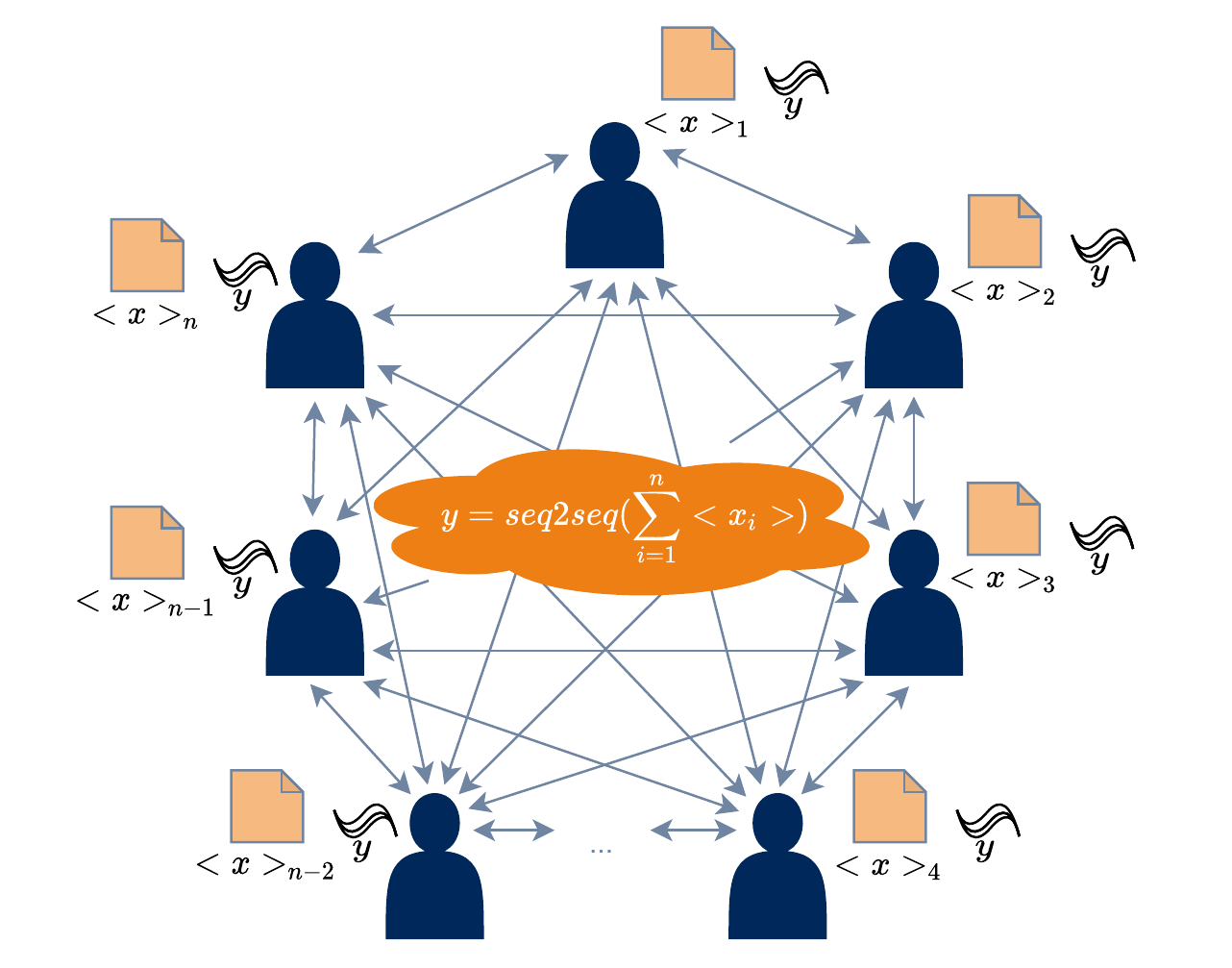}
    \caption{MPC framework designed by~\cite{feng2020securenlp}. The framework involves multiple parties which can obtain the publicly available trained seq2sec model. Secret sharing is used to compute the NLP task. $<x>_i$ denotes the knowledge held by each party $p_i$, and $y$ represents the model output.}
    \label{fig:multi-party-computation}
\end{figure}

Private data often arises from multiple providers, such as distributed databases, users, companies, and devices in internet of things. 
MPC is a PET applicable to such scenarios defined as a generic cryptographic primitive for secure computations of agreed functions over different private data sources, which should not be revealed~\citep{zhao2019secure,feng2020securenlp}. 
In contrast, the computation results hold an interest for all the parties and can be made available therein, assuring correctness and privacy properties~\citep{cramer_damgard_nielsen_2015}. 
For instance, when the exchange of plaintext is not allowed, MPC can be used to share keywords extracted from many text file sets to enrich NLP applications~\citep{feng2020securenlp}. 
Formally, there will be a set of inputs $\{x_1,x_2,\dots,x_n\}$ so that each party $p_i$ will hold $x_i$ and agree to compute $y = \hat{f}(x_1,x_2,\dots,x_n)$, in which $y$ is the output information for release, and $\hat{f}$ is the agreed function on the whole set of inputs ~\citep{cramer_damgard_nielsen_2015}. 
The inputs may include keywords, messages, and medical records.

Privacy in NLP models can also be achieved by MPC protocols that enable separate computations on secret inputs hailing from diverse parties, such as users, devices, and service providers. 
The parties involved in the computations should not learn from each other's inputs but their outputs instead since there should be no plaintext exchange. 
\cite{feng2020securenlp} design new MPC protocols aiming to preserve the privacy of every computation party in the NLP task of neural machine translation, simultaneously computing non-linear functions for deep neural networks quicker. 
Figure~\ref{fig:multi-party-computation} depicts the framework proposed by the authors.
The authors come up with interactive MPC protocols, using both additive and multiplicative secret sharing, for the non-linear activation functions of sigmoid $\sigma$ and tanh $\tau$, which are defined for any input value $x$, respectively, as
\begin{equation}
	\sigma(x) = \frac{e^{x}}{e^{x}-1}
\end{equation}
and
\begin{equation}
	\tau(x) = \frac{e^{2x}-1}{e^{2x}+1}.
\end{equation}
The protocols are implemented on an RNN-based seq2seq with an attention model, which performs predictions on multi-source data, keeping parties and attackers from learning the secret knowledge of another party during the model inference step. 

\subsubsection{Transfer learning methods}\label{subsub:transfer-learning-methods}

In domains that lack data for specific tasks and prevent data sharing, transfer learning (TL) is a straightforward solution relying on models that allow knowledge transfer by applying neural networks to tasks that differ from those targeted by previous training~\citep{weiss2016survey}. 
TL approaches can be adjusted for preserving private or sensitive information arising from medical text, documents, social media posts, etc. 
Taking the medical domain as an example, the vocabulary of NLP models may contain specific terms that breach the anonymity of documents used in the training step, so protective measures have to be set prior to the release of these models for further fine-tuning. 
\cite{alawad2020privacy} implement a multi-task CNN for TL on cancer registries aiming at information extraction from pathology reports concerning six characteristics (i.e., tumor site, subsite, literality, behavior, histology, and grade). 
Regarding privacy-preserving NLP, the authors come up with a restriction for the word embedding vocabulary to keep out PHI, such as patient names, therefore allowing the model to be shared between different registries observing data security. 
This vocabulary restriction consists of restraining the shareable tokens in the embedding vocabulary to those obtained from word embeddings pre-trained on corpora with no protected textual information like vectors trained on PubMed and MIMIC-III datasets~\citep{zhang2019biowordvec}.
Alongside the PET, the authors test two transfer learning settings: acyclic TL and cyclic TL. 
The former regards a usual TL approach in which a model is trained on a registry and forwarded to the next one for adjustments, whereas the latter embodies iterations between all the involved registries during the training step until the model converges.

Recognition of sensitive information is another use case suitable for TL approaches since each application field imposes its categories of sensitive content whose identification mostly relies on efforts spent by humans. 
\cite{martinelli2020nlpijcnn} propose a methodology to improve NLP models for knowledge transfer across general and specific domain tasks in steps prior to data obfuscation. 
From completely unlabeled corpora of documents, the authors are able to yield annotated versions whose sensitive content has been recognized and tagged. 
In the approach, word vectors extracted from fasttext make up the document representations to be combined with a sensitive content detection technique based on named-entity recognition and topics extraction. 
Therefore, the approach's outcomes may lighten the workloads of human officers that perform sensitive content detection manually.

\subsubsection{Encryption methods}\label{subsub:methods-for-encryption}

\begin{figure}
    \centering
    \includegraphics[width=\textwidth]{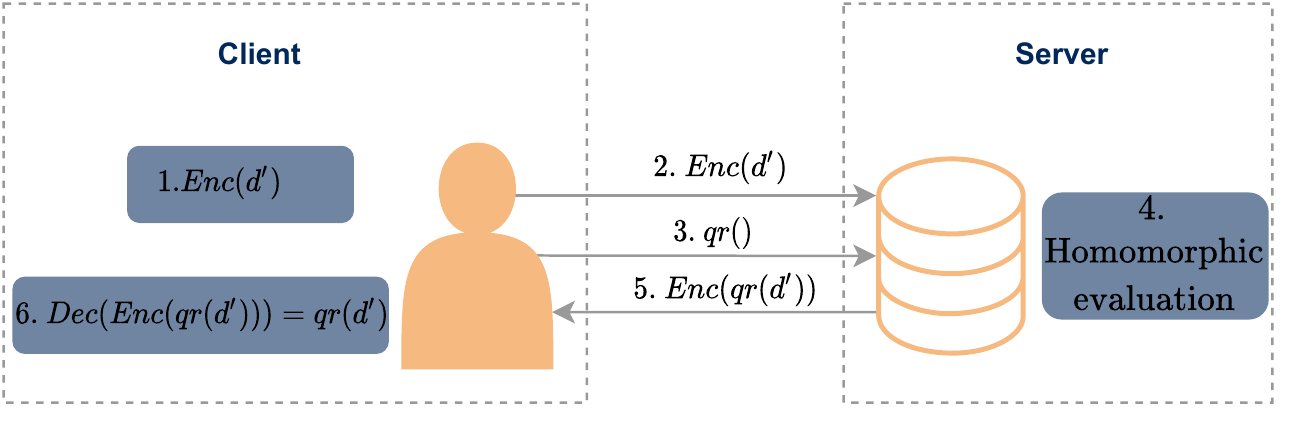}
    \caption{Simple homomorphic encryption scenario~\citep{acar2018survey}. In this scenario, $d'$ denotes a document. \textbf{1.} The client encrypts $d'$.
    \textbf{2.} The client sends its encrypted document $Enc(d')$ to the server.
    \textbf{3.} The client sends a query function $qr()$ to the server for querying its data.
    \textbf{4.} The server performs a homomorphic operation over the data without decryption.
    \textbf{5.} The server sends the encrypted results to the client.
    \textbf{6.} The client recovers the data with its private key and retrieves the query function results $qr(d')$.
    }\label{fig:simple-homomorphic-encryption}
\end{figure}

Encryption is a broadly used PET to encode information when sharing data in raw format is not a secure option. 
So a ciphertext is yielded by applying an encryption function over a data instance in the so-called plaintext. 
FHE has emerged as an encryption scheme that allows computations on encrypted data with no requirements of decryption~\citep{gentry2009fully}. 
In other words, it allows mathematical operations to be computed on encrypted data while it is still in its encrypted form~\citep{acar2018survey}, as depicted by Figure~\ref{fig:simple-homomorphic-encryption}.
Consequently, outsourced models hosted on untrusted cloud environments are able to perform training or inference on encrypted datasets without disclosing any clues about their content. 
However, FHE frequently leads to computation overheads due to the amount of noise used added to the ciphertexts, although its efficiency for privacy protection~\citep{li2020faster}. 
Another drawback of this encryption scheme is the need for approximating activation functions, such as relu, sigmoid, and tanh, since only addition and multiplication operations are supported.

\begin{figure}[h!]
    \centering
    \includegraphics[width=\textwidth]{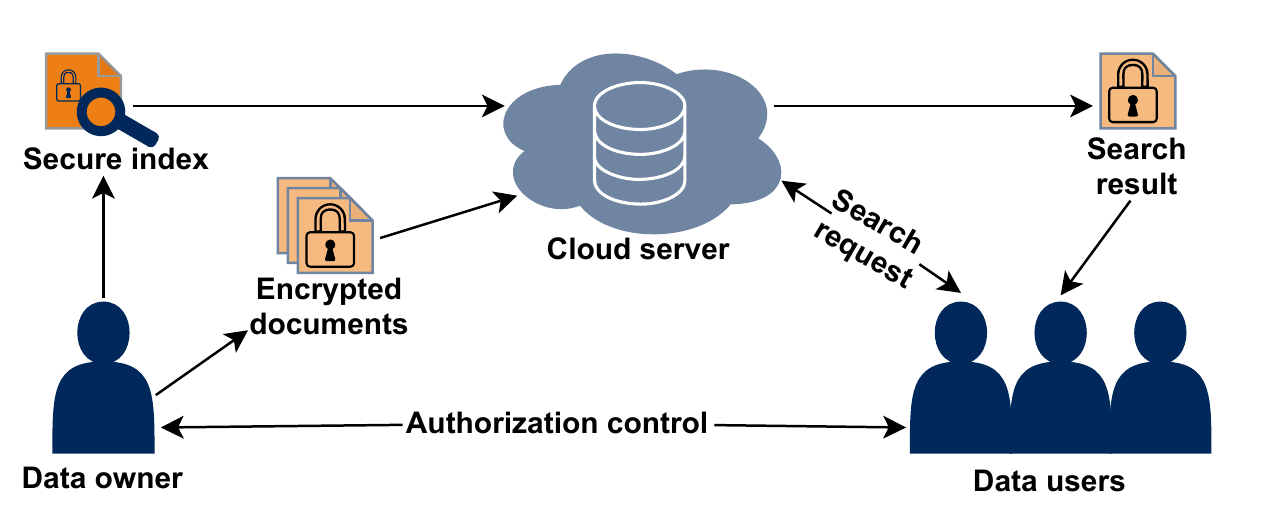}
    \caption{Searchable encryption system proposed by~\cite{dai2019efficient}.
    First, the data owner encrypts its documents and sends them to the cloud server in this system.
    Second, the data owner also encrypts document vectors and uses them as the secure index, which is also sent to the cloud server.
    Third, data users who have the authorization to query the encrypted documents encrypt their queries and send them to the cloud server.
    Further, the cloud server returns the encrypted document results to the data users.
    Afterward, the data users decrypt the document results using the secret key shared by the data owner.
    Finally, the query is finished.}
    \label{fig:searchable-encryption-system}
\end{figure}

Another encryption scheme that enables utility tasks is searchable encryption (SE), which encrypts document collections in such a manner that search capabilities can be delegated by the data owner without the need for decryption by the server or service provider~\citep{cash2015leakage}. 
Hence, an untrusted server can provide searches for uses without disclosing the content of both data and queries~\citep{liu2020mitigating}. 
Figure~\ref{fig:searchable-encryption-system} shows an example of a SE system.
Moreover, a common challenge on SE includes preserving the semantic relations between words and documents, which may undermine search results due to ambiguities in the keywords~\citep{dai2019efficient}. 
Generally, SE involves four algorithms for, respectively, key generation, encryption, token generation, and search~\citep{liu2020mitigating}. 
Nonetheless, this encryption scheme is sensitive to attacks that aim at recovering encrypted keywords.

\cite{dai2019efficient} propose two privacy-preserving keyword search schemes on data protected by SE, namely DMSRE and EDMRSE. 
Both schemes rely on the embedding model doc2vec~\citep{le2014distributed} for document representation. 
DMSRE is composed of five procedures. 
Firstly, it generates a secure key. 
Secondly, it pre-trains the embedding model and extracts feature vectors for every document $d'$ in the set of documents $D$. 
It also processes $d'$ and extracted vectors at the same time that it obtains the encrypted documents $\boldsymbol{\tilde{D}}$ and the encrypted document indexes $\boldsymbol{\tilde{I}}$ for later searches.
Subsequently, the trapdoor $\boldsymbol{\tilde{V_Q}}$ is produced for the queried keywords $\boldsymbol{Q}$ from a data user. 
Finally, it conducts the inner product between every document index in $\boldsymbol{\tilde{I}}$ and $\boldsymbol{\tilde{V_Q}}$ in order to retrieve the $k$ most semantically related results as:
\begin{equation}
    \mid \boldsymbol{RList}\mid = k \forall \tilde{d_i}, \tilde{d_j} (\tilde{d_i} \in \boldsymbol{RList} \land \tilde{d_j} \in (\boldsymbol{\tilde{D}} - \boldsymbol{RList})) \rightarrow \boldsymbol{\tilde{I}}_i \cdot	 \boldsymbol{\tilde{V_Q}} > \boldsymbol{\tilde{I}}_j \cdot	 \boldsymbol{\tilde{V_Q}}.
\end{equation}
EDMRSE has its security enhanced by adding phantom terms on both document vectors and trapdoors to confound the search results and keep the cloud model from gathering statistical information about the documents. 
Except for the procedure of pre-training the embedding model and extracting the document's vector, all the remaining ones have different definitions to deal with the introduced phantom terms. 
As a result, EDMRSE increases privacy protection but decreases search accuracy.

Although providing efficient schemes for privacy protection of text data, SE is prone to file injection attacks in which a malicious adversary injects customized files into the encrypted search system and distinguishes specific keywords by observing the patterns on the encrypted versions of the injected files. 
\cite{liu2020mitigating} investigate the susceptibility of leakages from data protected by a SE scheme using an LSTM model to generate text files to be injected into the encryption scheme. 
The authors conducted extensive experiments finding that automatically generated texts present low quality that could be manually identified by humans and also diminish the feasibility likelihood of attacks that use them as injected files. 
Automatic file injection attack detection was performed by three ensemble methods, i.e., random forest, Adaboost based on support vector machines, and Adaboost based on random forest. 
Among those methods, the third one provided the highest accuracy rates in the study. 
The main takeaway from this work regards the practicability to identify automatically generated files for injection attacks, although semantically meaningful files injected in an ad-hoc manner cannot be easily detected and attain successful attacks.

\begin{figure}[ht!]
    \centering
    \includegraphics[width=\textwidth]{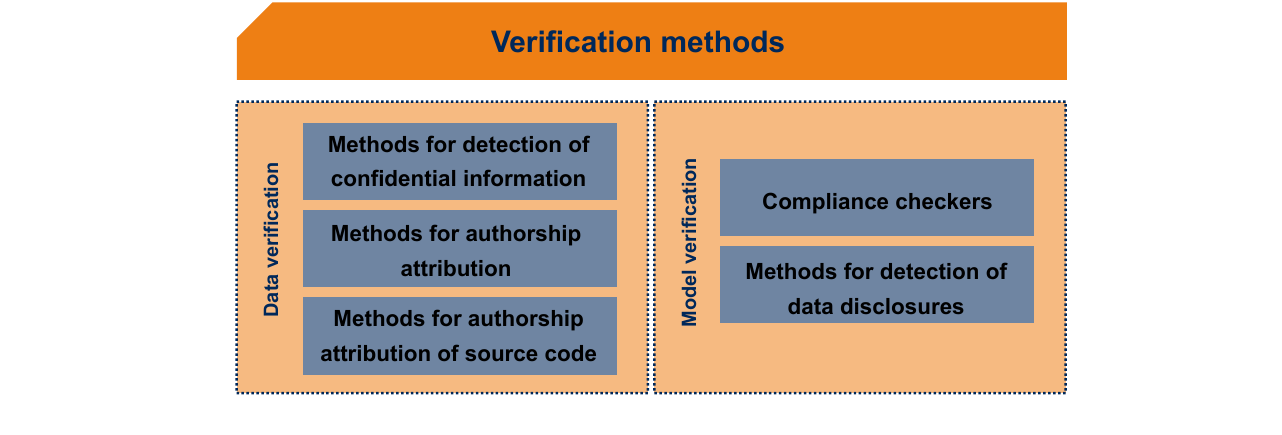}
    \caption{Sub-categories and groups for verification methods}
    \label{fig:verification-methods}
\end{figure}

\subsection{Verification methods}\label{subsub:verification-methods}

Verification methods encompass approaches for verifying data and model susceptibility to privacy threats.
These methods are mostly applied during post-processing evaluations.
Figure~\ref{fig:verification-methods} depicts the sub-categories and groups for verification methods.
Furthermore, Table~\ref{tab:summary-of-verification-methods} summarizes each group of methods separately alongside the works these groups are composed of.

\subsubsection{Methods for detection of confidential information}\label{subsub:methods-conf-information}

\begin{sidewaystable}
\sidewaystablefn%
\begin{center}
\begin{minipage}{\textheight}
\begin{threeparttable}
    \caption{Summary of verification methods}\label{tab:summary-of-verification-methods}
    \begin{tabular}{l|lllll}
    \toprule
    \textbf{Group} & \textbf{Work} & \textbf{Neural models} & \textbf{PET} & \boldmath{$T$} & \boldmath{$S$} \\
    \toprule
    \multirow{11}{*}{Methods for DCI} & \cite{neerbek2018detecting} & RecNN, GloVe & Relative weighting & $W$ & SC\\
    {} & \cite{battaglia2020towards} & MLP, GloVe & CSA & $A$ & SC \\
    {} & \multirow{2}{*}{\cite{zhao2020gender}} &
         ELMO, BERT, & Bias analysis & $W$ & DC \\
    {} & {} & XL-Net, fasttext & \\
    {} & \multirow{3}{*}{\cite{tan2019assessing}} & GloVe, ELMO, & Bias analysis & $W$ & LE \\
    {} & {} & BERT, GPT, & \\
    {} & {} &  GPT-2 & \\
    {} & \cite{hutchinson2020social} & BERT & Bias analysis & $W$ & LE \\
    {} & \multirow{2}{*}{\cite{gonen2019lipstick}} & Hard-Debiased, & Bias analysis & $W$ & TL \\ 
    {} & {} & GN-GloVe, Glove  & \\
    {} & \cite{nissim2020fair} & word2vec & Analogy detection & $W$ & LE \\ 
    {} & \multirow{2}{*}{\cite{sweeney2019transparent}} & word2vec, GloVe, & Bias analysis by SS & $W$ & LE \\
    {} & {} & Conceptnet & \\
    \hline
    \multirow{ 5}{*}{Methods for AA} & 
   \cite{shrestha2017convolutional} & CNN & Feature extraction & $A$ & DC \\
    {} & \multirow{2}{*}{\cite{boumber2018experiments}} & CNN, word2vec, & Feature extraction & $A$ & DC\\
    {} & & GloVe & & \\
    {} & \multirow{2}{*}{\cite{barlas2020cross}} &
         RNN, BERT, ELMO, & PTM & $A$ & DC\\
    {} & {} & GPT-2, ULMFiT & \\
    {} & \cite{caragea2019myth} & CNN, word2vec & Feature extraction & $A$ & DC\\
    \hline
    Methods for AA & \cite{abuhamad2018large} & LSTM+GRU & Feature extraction & $A$ & OSC\\ 
    of source code & \cite{abuhamad2019code} & CNN, C\&W & Feature extraction & $A$ & OSC\\
    \bottomrule
    \end{tabular}
    \textit{Continues on next page...}
    \end{threeparttable}
    \end{minipage}
    \end{center}
    \end{sidewaystable}

\begin{sidewaystable}
\sidewaystablefn%
\begin{center}
\begin{minipage}{\textheight}
\begin{threeparttable}   
\ContinuedFloat  
\caption{Summary of verification methods (Continued)}
\begin{tabular}{l|lllll}
    \toprule
    \textbf{Group} & \textbf{Work} & \textbf{Neural models} & \textbf{PET} & \boldmath{$T$} & \boldmath{$S$} \\
    \toprule
    \multirow{6}{*}{Compliance checkers} & \cite{song2019auditing} & LSTM, seq2seq & Membership inference & $A$ & DC \\ 
    {} & \multirow{3}{*}{\cite{may2019measuring}} &
         GloVe, InferSent,
        & Bias analysis & $W$ & WCE \\ 
    {} & & GenSent, USE, & \\
    {} & {} & ELMO, GPT, & \\
    {} & {} & BERT & \\
    {} & \cite{Basta2020ExtensiveSO} &
    	word2vec, ELMO & Bias analysis & $W$ & WCE \\
	{} & \cite{vig2020causal} & GPT-2 & Bias analysis & $W$ & WCE \\
    \hline
    \multirow{ 9}{*}{Methods for DDD} &   \multirow{4}{*}{\cite{song2020information}} & 
         word2vec, GloVe,
         & Disclosure attacks & $A$ & PA \\ 
    {} & {} & Fasttext, LSTM, & \\
    {} & {} & Transformer, BERT, & \\
    {} & {} & ALBERT & \\
    {} & \multirow{4}{*}{\cite{pan2020privacy}} & 
    BERT, transformer-XL, & Disclosure attacks & $W$ &  MR  \\
    {} & {} & XLNet, GPT, & {} \\
    {} & {} & GPT-2, RoBERTa, & {} \\
    {} & {} & XLM, ERNIE 2.0 & \\
	{} & \cite{carlini2019secret} & LSTM & Exposure metric & $W$ & PA \\
	{} & \multirow{2}{*}{\cite{akiti-etal-2020-semantics}} & BiLSTM, GloVe, & SRL & $A$ & SM \\
	{} & {} & BERT, ELMO & \\
    \bottomrule
    \end{tabular}
    \footnotetext{\small
      \boldmath{$A$} stands for a set of protected attributes, 
      \textbf{AA} stands for `authorship attribution',
      \textbf{CSA} stands for `content sensitivity analysis', 
      \textbf{DC} stands for `document collection',
      \textbf{DDD} stands for 'detection of data disclosures',
      \textbf{DCI} stands for `detection of confidential information',
      \textbf{GRU} stands for `Gated recurrent units',
      \textbf{LE} stands for `language encoding',
      \textbf{MR} stands for `medical records',
      \textbf{OSC} stands for `open source contributors',
      \textbf{PA} stands for `private attributes',
      \textbf{PET} stands for `privacy-enhancing technology',
      \textbf{PTM} stands for `pre-trained model',
      \boldmath{$S$} stands for `scenario', 
      \textbf{SC} stands for `sensitive content',
      \textbf{SM} stands for `social media',
      \textbf{SRL} stands for `sentence role labeling',
      \textbf{SS} stands for `sentiment scoring',
      \boldmath{$T$} stands for `target', 
      \textbf{TL} stands for `transfer learning',
      \boldmath{$W$} stands for a set of target words,
      \textbf{WCE} stands for `word or context embeddings'.    }
  \end{threeparttable}
  \end{minipage}
\end{center}
\end{sidewaystable}

Documents compose a source of information that must remain secret, under organizational ethics or privacy preservation laws, in order to prevent data leakages. 
Human experts have manually performed most of the efforts to tag sensitive or confidential information, hence being prone to workforce and time overheads, depending on the volume of documents to be analyzed.  
So deep neural networks have recently been employed in detecting these kinds of information automatically and quickly on large document collections.
\cite{neerbek2018detecting} propose to learn phase structures discriminating between documents with sensitive and non-sensitive content using a recursive neural network~\citep{irsoy2014deep} trained on labeled documents with no need to label every sentence itself. 
The authors argue that current keyword-based approaches for detecting sensitive information may fail to find complex sensitive information since these models do not take into account the sentences' context in the document. 
This architecture recursively receives a part of the input structure as input in each step. 
From sentences modeled as parse-trees, the neural network preserves the grammatical order of the sentences in a bottom-up manner, which consists of processing a node in the parse-tree in each step, ending at the root-node. 
A relative weighting strategy is implemented to distinguish between sensitive and non-sensitive information by higher weights to the kind of information detected in the sentence while computing a cross-entropy loss.
 
Measuring the harmfulness of information is a challenging task. 
In order to address it, \cite{battaglia2020towards} defined a new data mining task called content sensitivity analysis which aims at assigning scores to data files taking into account their degree of sensitivity as a function $\hat{s}: \mathscr{O} \rightarrow [-1,1]$, in which $\mathscr{O}$ is the domain of all user-generated contents. Therefore, given a user-generated object $o'_i$, $\hat{s}(o'_i) = 1$ if $o'_i$ is maximally privacy-sensitive, on the contrary $\hat{s}(o'_i) = -1$. 
Since the notion of sensitive information is subjective depending on use cases and data protection regulations, $\hat{s}: \mathscr{O} \rightarrow [-1,1]$ can be learned according to an annotated corpus of content objects satisfying:
\begin{equation}
	\min \sum_{i=1}^{N}(\hat{s}(o'_i)-\beta_i^2),
\end{equation}
in which $O' = \{(o'_i, \beta_i)\}$ is a set of $N$ annotated objects $o'_i \in \mathscr{O}$ with their related sensitivity score $\beta$. 
In the experimental evaluation, the authors used GloVe embeddings to represent words and an MLP network to distinguish whether an input text is sensitive or not as a binary classification setting.

Discriminatory biases towards individuals or groups in DL models frequently arise from demographic attributes in the datasets leading to discriminatory decisions towards individuals or groups. 
Many recent approaches in the literature introduced methodologies for the detection and measurement of human biases in word representations. 
\cite{zhao2020gender} quantify gender bias in language models for cross-lingual transfer, i.e., a scenario in which a model trained in one language is deployed to applications in another language. 
In the study, the authors evaluated three transfer learning methods, namely ELMO~\citep{Peters-2018}, BERT, and XL-Net~\citep{NEURIPS2019_dc6a7e65}, in addition to a modified fasttext model, applying a vector space alignment technique to reduce bias and a metric to quantify it. 
\cite{tan2019assessing} analyze the extent to which contextualized word representation models (ELMO, BERT, GPT, and GPT-2) and GloVe embeddings can encode bias related to demographic attributes, such as gender, race, and intersectional identities.
\cite{hutchinson2020social} reveal evidence of bias encoded by language models towards mentions of disabilities, focusing on the BERT model. 
\cite{gonen2019lipstick} conduct experiments hypothesizing that gender bias information might still be reflected in the distances between gender-neutral words in debiased embedding models, hence presenting the risk of recovery. 
Two embedding models are used by the authors, namely Hard-Debiased~\citep{bolukbasi2016man} and GN-GloVe, and then compared against the standard GloVe. 
\cite{nissim2020fair} investigate the role analogies play in bias detection by considering these language structures as an inaccurate diagnosis for bias. 
In fact, the authors claim that analogies may have been overused, so possibly non-existing biases have been exacerbated, and others have been hidden. 
The word2vec model is used in the experiments alongside measures to detect analogies related to gender and profession names, such as gynecologist, nurse, and doctor. 

Most of the approaches for measuring bias in word embedding models rely on distances in vector spaces. 
Claiming that insights based on those metrics are geometrically rich but limited with regard to model interpretability, \cite{sweeney2019transparent} present a framework for evaluating discriminatory biases on embedding models towards protected groups, such as national origin and religion. 
The authors introduce a novel metric for measuring fairness in word embedding models named RNSB, which takes into account the negative sentiment associated with terms related to demographic groups. 
In the study, three pre-trained word embedding models are evaluated, namely GloVe, word2vec, and ConceptNet~\citep{speer2017conceptnet}.
The last one was proved to be the least biased among the analyzed models.

\subsubsection{Methods for authorship attribution (AA)}\label{subsub:methods-aa}

Determining the identity of written text's authors is a challenging problem with many real-world applications, such as plagiarism detection, spam filtering, phishing recognition, identification of harassers, text authenticity check, and detection of bot users on social media. 
AA is the task of distinguishing texts written by different authors based on stylometric features~\citep{stamatatos2009survey}. 
Unveiling the authorship of written texts also poses privacy threats since both text content and the author's identity may be subject to data protection regulations. 
Furthermore, the author's willingness to share their identity is also a key factor to bear in mind during the development of AA applications which can be used, for instance, to re-identify anonymous texts or comments which are widely available on the internet as happened to the Netflix prize's dataset~\citep{narayanan2008robust}. 

Addressing AA for short texts is an even more complex task compared to longer texts, according to \cite{shrestha2017convolutional}. 
The authors apply a CNN model over character n-grams, which capture patterns at the character level to help model the style of different authors. 
Firstly, the CNN architecture receives a sequence of character n-grams as input and forwards it across three modules: an embedding module, a convolutional module, and a fully connected softmax module. 
Secondly, the embedding module yields a vector representation for the character n-grams, passed over the convolutional layer to capture a feature map $\omega$. 
Therefore, $\omega$ is pooled by max-over-pooling to produce $\tilde{v}_{k'}$ in the form
\begin{equation}
	\tilde{v}_{k'} = \max_{{i}} \omega_{k'}[i],{k'}=1,\dots,m',
\end{equation}
in which $\tilde{v}_{k'}$ is the maximum value in the $k'$-th feature map, and $m'$ is the number of feature maps. 
Finally, after concatenating the pooled maps, the model generates a compacted text representation with the most important text features regardless of their original position. 
This representation is then input to a softmax layer that discriminates the text's author. 
The authors come up with the hypothesis that their model is able to learn features at morphological, lexical, and syntactical levels simultaneously.

Some scenarios impose further hurdles for AA models, as noticed in documents written by multiple authors whose detection resembles a multi-label classification problem. 
\cite{boumber2018experiments} design a CNN architecture for multi-label AA handling documents as sets of sentences that may present many labels. 
This model implements a strategy called collaborative section attribution, which consists of taking two possibilities into account concurrently. 
The first regards continuous sections written by a single author, while the second refers to the influence coauthors play on each other's writing style or, yet, editing passages written by others. 
Two word embedding models (word2vec and GloVe) are used on the multi-label AA architecture to represent the words in the documents. 
Like \cite{shrestha2017convolutional}, feature extraction steps have produced a vector to be input into a classification layer with a softmax activation function. 

Divergences between training and test sets of texts represent another realistic and recurrent barrier for the task of AA. 
\cite{barlas2020cross} take differences between textual genre and topic by dismissing information related to these two factors and solely concentrating on stylistic properties of texts associated with the personal writing styles of authors. 
In the work, the authors applied four pre-trained language models (i.e., ULMFiT, ELMO, GPT-2, and BERT) to cross-genre and cross-topic AA. 
Besides holding state-of-the-art results across an extensive range of NLP tasks, pre-trained models do not pose privacy for the text corpora used in their training steps since those corpora are typically composed of texts publicly available on the internet, such as news or Wikipedia articles. 
However, it is still unclear if the writing styles of the training data may affect the model behavior while performing AA.

Anonymity is occasionally preferred over explicit authorship information when a fair decision about a text must be made. 
For instance, during scientific papers' review, the author's name information in the manuscript up to evaluation would eventually bias the reviewers towards world-class authors over less renowned names. 
\cite{caragea2019myth} investigate the effectiveness of deep neural networks for inferring the authorship information of scientific papers submitted to two top-tier NLP conferences (ACL and EMNLP).
In order to perform their study, the authors implement a CNN model trained on scientific paper collections from the two conferences. 
The word2vec embedding model produces the representations for the words in the documents prior to inputting these vectors to the convolutional layers. 
Separate sub-components of the CNN are responsible for feature extraction from paper content, stylometric features, and the entries on each paper's references. 
On the network top, a fully connected layer and a classification layer with a softmax activation function predict the class for each article after receiving the outputs from the three sub-components for feature extraction.

\subsubsection{Methods for authorship attribution of source code}\label{subsub:methods-aic}

AA is an NLP task regarding the assignment of correct authors to contentious samples of work whose writing is anonymous or disputable, including source code in programming languages~\citep{burrows2014comparing}. 
\cite{abuhamad2018large} argue that AA of source code poses privacy threats regarding developers working on open source projects when they refuse to reveal their identities.
However, it can enhance applications for digital forensics, plagiarism detection, and identification of malware code developers. 
So the authors propose a DL-based system for AA of source code relying on an LSTM model with gated recurrent units for yielding representations of TF-IDF vectors of code files. 
A classifier receives the features learned by the LSTM architecture and performs AA, and technical analyses are conducted. 
In a similar fashion, \cite{abuhamad2019code} come up with CNN models for this task. 
Both TF-IDF and C\&W~\citep{collobert2011natural} vectors generate representations for source code, as well as are compared with respect to the final classification performance. 
Therefore, the neural networks learn features and classify the input vectors by assigning their authorship. 

\subsubsection{Compliance checkers}\label{subsub:compliance-checkers} 
Data privacy has been assured by laws in many countries, such as the EU's GDPR, hence some approaches in the literature of privacy-preserving NLP verify whether DL models complied or not with such regulations or with concepts of fairness. 
It is a common practice to use user-generated data to input language models for applications like word prediction, automatic question-answering, and dialogue generation. However, there should be transparency in the data collection and utilization. 
\cite{song2019auditing} develop an auditing technique that is able to inform users if their data was used during the training step of text generation models built on LSTMs. 
This study also analyzes to what extent language models memorize their training data since it can be considered a problem for both NLP and privacy aspects. 
According to the authors, auditing can be thought of as a membership inference towards a model at the user level.

The detection of implicit human biases encoded by word representation models is an extensive field of research in NLP.
Such biases can be propagated into downstream applications and lead to discriminatory decisions, hence breaching good practice protocols and regulations that enforce fairness.
Thus, we include in the group of compliance checkers works that propose new methods for detecting bias in embeddings.
Here, the major focus consists of bias detection, contrary to Section~\ref{subsub:debiasing-m}, which focuses on the bias reduction and mitigation.
Common means for bias identification are tests like the Word Embedding Association Test (WEAT)~\citep{caliskan2017semantics}, which measures the association between two equally sized sets of target word embeddings $\tilde{V}$ and $\tilde{W}$, and two sets of attribute word embeddings $\tilde{A}$ and $\tilde{B}$. 
WEAT's test statistic is computed in the form:
\begin{equation}
	\rho(\tilde{V},\tilde{W},\tilde{A},\tilde{B}) = [\sum_{\tilde{v} \in \tilde{V}} \rho(\tilde{v},\tilde{A},\tilde{B}) - \sum_{\tilde{w} \in \tilde{W}} \rho(\tilde{w},\tilde{A},\tilde{B})],
\end{equation}
in which $\rho(\tilde{t},\tilde{A},\tilde{B})$ is the difference between the mean cosine similarity of the attribute word embeddings $\tilde{a}$ and $\tilde{b}$, respectively in $\tilde{A}$ and $\tilde{B}$,~\citep{may2019measuring} computed as
\begin{equation}
	\rho(\tilde{t},\tilde{A},\tilde{B}) = [mean_{\tilde{a} \in \tilde{A}} cos(\tilde{t},\tilde{a}) - mean_{\tilde{b} \in \tilde{B}} cos(\tilde{t},\tilde{b})]
\end{equation}
Specifically, $\rho(\tilde{t},\tilde{A},\tilde{B})$ is a measure of how associated a target word embedding $\tilde{t}$ and a attribute word embeddings are, while $\rho(\tilde{V},\tilde{W},\tilde{A},\tilde{B})$ is a measure of the association between the sets of target word embeddings and the attribute ones~\citep{caliskan2017semantics}. 
Additionally, permutation test on $\rho(\tilde{V},\tilde{W},\tilde{A},\tilde{B})$ computes the significance of the association between both pairs of target and attribute word embeddings~\citep{may2019measuring}, as
\begin{equation}
	\tilde{p} = Pr_i[\rho(\tilde{V}_i,\tilde{W}_i,\tilde{A},\tilde{B})>\rho(\tilde{V},\tilde{W},\tilde{A},\tilde{B})],
\end{equation}
in which $(\tilde{V}_i,\tilde{W}_i)$ stand for all partitions in $\tilde{V} \cup \tilde{W}$, and $\tilde{p}$ is one sided $p$-value of the test. Finally, the magnitude of the association~\citep{caliskan2017semantics,may2019measuring} is calculated by
\begin{equation}
	\mu = \frac{mean_{\tilde{v}\in \tilde{V}}\rho(\tilde{v},\tilde{A},\tilde{B})-mean_{\tilde{w}\in \tilde{W}}\rho(\tilde{w},\tilde{A},\tilde{B})}{std\_dev_{\tilde{t} \in \tilde{V} \cup \tilde{W}}\rho(\tilde{t},\tilde{A},\tilde{B})}.
\end{equation}

Since WEAT is meant to detect biases at the word level, \cite{may2019measuring} propose a generalized version of this test titled Sentence Encoder Association Test (SEAT) to uncover biases at the phrase and sentence levels. 
To do so, SEAT is inputted with sets of contextualized embeddings for sentences with similar structure, focusing on specific words, like ethnic names, and attributes that are related to biases towards the specific words. 
The experiments performed by the authors targeted two types of bias, namely the black woman stereotype~\citep{harris2011sister} and double binds defined as antagonistic expectations of femininity and masculinity~\citep{harris2011sister,may2019measuring}. 
A total of seven sentence encoders (Table~\ref{tab:summary-of-verification-methods}) are analyzed with SEAT, drawing evidence that such encoders evince less bias than prior word embedding models.

\cite{Basta2020ExtensiveSO} evaluate language models for both English and Spanish languages with regard to gender bias. 
While the former language does not present distinct gendered forms for most nouns, the latter heavily does. 
Furthermore, the authors do not use WEAT or SEAT as methods for detecting biases on ELMO and word2vec embeddings. 
Their experimental evaluation takes into account metrics based on principal component analysis, gender direction, clustering, and supervised classifiers, namely support vector machines and $K$-NN. 
Similarly, \cite{vig2020causal} present a methodology relying on causal mediation analysis~\citep{pearl2001direct} to analyze which internal components of pre-trained GPT-2 models concentrate most of the gender bias learned from the training data. 
Causal mediation analysis gauges the extent network neurons mediate gender bias individually and jointly. 
The study concludes that gender bias is concentrated in a few language model components, and the individual effects of some components may be amplified by interactions. 
Last, the authors also argue that the total gender bias effect approximates the sum of both direct and indirect effects related to the information flows from input to output variables.

\subsubsection{Methods for detection of data disclosures}\label{subsub:methods-dd}

Language models often handle private or sensitive attributes, such as demographic information or stylometric features, which should remain unveiled for parties using or accessing these neural networks. 
In order to verify the susceptibility of such models to breaching private data, recent studies have come up with attacking approaches or information tracking methodologies. 
\cite{song2020information} develop three classes of attacks to study which kinds of information could be leaked from word embedding models. 
First, embedding inversion attacks aim to invert existing vector representations to their raw text formats, including private content. 
Second, attribute inference attacks check how likely embedding models are to reveal sensitive attributes from the training data. 
Finally, membership inference attacks demonstrate the ability to recover training samples when an adversarial has access to both the language model and its outputs. 

\cite{pan2020privacy} investigate how 8 general-purpose language models (Table~\ref{tab:summary-of-verification-methods}) capture sensitive information which can later be disclosed by adversaries for harassment afterward.
The authors design two classes of attacks aiming at disclosing sensitive information from the tested models. 
First, pattern reconstruction attacks enable adversaries to recover sensitive segments of sequences, such as genome sequences, used for model training. 
Second, keyword inference attacks target sensitive keywords in unknown texts, like medical descriptions. 
The study has found that leaked embeddings present a high potential to allow adversaries to disclose sensitive information from users.

Unintended memorization is a noticeable drawback of neural networks that may put training instances' privacy at risk of disclosure in case these instances hold unique or rare values, such as IDs, addresses, or credit card numbers. 
Aiming to assess this problem, \cite{carlini2019secret} describe a testing methodology that limits data exposure by minimizing its memorization. 
An exposure metric is proposed by the authors to quantify the propensity of data disclosure by neural models. 
Consequently, the study finds that memorization may not be due to excessive model training but a side-effect that appears at the early stages of training and prevails on several models and training frameworks.
 
On social media platforms, such as Twitter, Facebook, and Instagram, people share huge amounts of content every day, often including their own private information, unaware of the privacy risks these actions may bring about. 
\cite{akiti-etal-2020-semantics} detect emotional and informational self-disclosure on Reddit data using semantic role labeling, which is a technique for recognizing predicate-argument pairs in sentences. 
According to the authors, emotional self-disclosures regard the user's feelings about people or things, while the informational ones are related to revealing personal information, such as age, career status, address, and location. 
As a result, the authors overtake state-of-the-art methods, demonstrating the approach's efficiency in detecting personal disclosures.   

\section{Applications and datasets}\label{sec:applications}

\begin{sidewaystable}
\begin{center}
\begin{minipage}{\textheight}
\caption{Summary of tasks and benchmark datasets for privacy-preserving NLP}~\label{tab:summary-of-datasets}
\centering
\begin{tabular}{l|l}
    \toprule
    \textbf{NLP task} & \textbf{Datasets \& works} \\
    \toprule
      Anonymization of document images & Invoice images dataset~\citep{sanchez2018automatic}. \\
      \midrule
      Authorship attribution & 20-author set, 50-author set~~\citep{fernandes2019generalised},\\
      {} & Twitter~\citep{shrestha2017convolutional}, \\
      {} & MLPA-400, PAN-2012~\citep{boumber2018experiments}, \\
      {} & ACL papers, EMNLP papers~\citep{caragea2019myth}, \\
	  {} & Enron~\citep{feyisetan2020privacy},\\
	  {} & CMCC~\citep{barlas2020cross}. \\
	  \midrule
	  AA of source code & GitHub, Google Code Jam (2008-2016)~\citep{abuhamad2018large,abuhamad2019code}.\\
	  \midrule
	  Author obfuscation & PAN11, PAN12~\citep{feyisetan2019leveraging}.\\
	  \midrule
      Bias assessment & Winobias~\citep{vig2020causal,tan2019assessing},\\
	  {} & Winogender~\citep{vig2020causal},\\
	  {} & MIBs~\citep{zhao2020gender},
	  1BWord, BookCorpus~\citep{tan2019assessing}, \\
	  {} & Wikipedia, WebText~\citep{tan2019assessing}, \\
	  {} & Reddit~\citep{hutchinson2020social}, \\
	  {} & Professions~\citep{gonen2019lipstick}, \\
	  {} & Word2vec test set~\citep{nissim2020fair}, \\
	  {} & Sentiment training~\citep{speer2017conceptnet}.\\
      \midrule
      Biomedical translation & Biomed~\citep{Basta2020ExtensiveSO}. \\
      \midrule
      Capitalization & MedText~\citep{melamud2019towards}.\\
      \bottomrule
      \multicolumn{2}{l}{\textit{Continues on next page...}} \\
    \end{tabular}
    \end{minipage}
    \end{center}
    \end{sidewaystable}

\begin{sidewaystable}
\sidewaystablefn%
\begin{center}
\begin{minipage}{\textheight}
\ContinuedFloat  
\caption{Summary of tasks and benchmark datasets for privacy-preserving NLP (Continued)}
\centering
\begin{tabular}{l|l}
    \toprule
    \textbf{NLP task} & \textbf{Datasets \& works} \\
    \toprule
      Clinical notes generation & MedText-2~\cite{melamud2019towards},\\
      {} & MedText-103, WikiText-2, WikiText-103~\citep{melamud2019towards}.\\
      \midrule
      Content sensitive analysis & Dataset created by the authors~\citep{battaglia2020towards}.\\
      \midrule
      Co-reference resolution & Ontonotes 5.0, WinoBias~\citep{zhao2018learning}.\\
      \midrule
      De-identification of medical records & 2014 i2b2~\citep{liu2017identification,dernoncourt2017identification,friedrich-etal-2019-adversarial},\\
      {} & 2016 N-GRID~\citep{liu2017identification},\\
      {} & MIMIC~\citep{dernoncourt2017identification}.\\
      \midrule
	  Detection of altered mental status &  AMS~\citep{obeid2019impact}. \\
	  \midrule
	  Detection of demographic biases & CF Twitter, 100 Authors Twitter~\citep{barrett-etal-2019-adversarial},\\
	  {} & PAN14 Blogs, PAN14 Reviews~\citep{barrett-etal-2019-adversarial},\\
	  {} & PAN14 SoMe, PAN16 Rand~\citep{barrett-etal-2019-adversarial},\\
	  {} & German Facebook comments~\citep{papakyriakopoulos2020bias},\\
	  {} & Jigsaw dataset, Multilingual Tweets~\citep{gencoglu2020cyberbullying}, \\
	  {} & WikiDetox dataset, Gab Hate Corpus~\citep{gencoglu2020cyberbullying}.\\
	  \midrule
	  Detection of self-disclosures & Reddit~\citep{akiti-etal-2020-semantics}. \\
	  \midrule
	  Detection of sensitive information & Justice corpus, Healthcare corpus~\citep{martinelli2020nlpijcnn},\\
      {} & Enron~\citep{neerbek2018detecting}. \\
     \bottomrule
    \multicolumn{2}{l}{\textit{Continues on next page...}} \\
    \end{tabular}\\
    \end{minipage}
    \end{center}
    \end{sidewaystable}

\begin{sidewaystable}
\sidewaystablefn%
\begin{center}
\begin{minipage}{\textheight}
\ContinuedFloat  
\caption{Summary of tasks and benchmark datasets for privacy-preserving NLP (Continued)}
\centering
\begin{tabular}{l|l}
    \toprule
    \textbf{NLP task} & \textbf{Datasets \& works} \\
    \toprule
      Dialog generation & Cornell movie dialogs, Ubuntu dialogs~\citep{song2019auditing}.\\
      \midrule
      Disentangling latent spaces & Yelp, Amazon reviews~\citep{john2019disentangled}.\\
      \midrule
      Evaluation of privacy guarantees & GloVe vocabulary~\citep{feyisetan2019leveraging}.\\
      \midrule
      Evaluation of word vector distortion & WordSim353~\citep{sweeney2020reducing}. \\
      \midrule
      Extraction of sensitive information & Randomly generated citizen IDs, Genome~\citep{pan2020privacy}.\\
      \midrule
      Feature extraction & Louisiana Tumor Registry~\citep{alawad2020privacy},\\
      {} & Kentucky Cancer Registry~\citep{alawad2020privacy}. \\
      \midrule
      Keyword inference attacks & Airline reviews, CMS public healthcare records~\citep{pan2020privacy}.\\
      \midrule
      Keyword search & Amazon reviews, Enron, Science dataset~\citep{liu2020mitigating},\\
      {} & 20 news groups~\citep{dai2019efficient}.\\
      \midrule
      Medical document anonymization & MEDDOCAN, NUBES~\citep{pablos2020sensitive}.\\
      \midrule
      Memorization detection & Enron~\citep{carlini2019secret}. \\
      \midrule
      Model training & Penn Treebank (PTB), WikiText-103~\citep{carlini2019secret}. \\
      \midrule
      Native language detection & L2-Reddit, TOEFL17~\citep{kumar-etal-2019-topics}. \\
      \bottomrule
      \multicolumn{2}{l}{\textit{Continues on next page...}} \\
      \end{tabular}\\
    \end{minipage}
    \end{center}
    \end{sidewaystable}

\begin{sidewaystable}
\sidewaystablefn%
\begin{center}
\begin{minipage}{\textheight}
\ContinuedFloat  
\caption{Summary of tasks and benchmark datasets for privacy-preserving NLP (Continued)}
\centering
\begin{tabular}{l|l}
    \toprule
    \textbf{NLP task} & \textbf{Datasets \& works} \\
    \toprule
      Natural language inference & MedNLI~\citep{melamud2019towards},\\
      {} & SICK-E~\citep{feyisetan2019leveraging}.\\
      \midrule
      Natural language understanding & GLUE~\citep{huang2020texthide}.\\
      \midrule
      Neural machine translation & WMT~\citep{feng2020securenlp,Basta2020ExtensiveSO},\\
      {} & SATED~\citep{song2019auditing},\\
      {} & Europarl~\citep{Basta2020ExtensiveSO,song2019auditing},\\
      {} & TEDx, WMT13 (Spanish)~\citep{Basta2020ExtensiveSO},\\
	  {} & Newstest2012, Newstest2013~\citep{font2019equalizing}.\\
	  \midrule
      News recommendation & Adressa, MSN-News~\citep{qi-etal-2020-privacy}. \\
      \midrule
      Next word prediction & en\_US keyboard clients~\citep{chen2019federated},\\
      {} & pt\_BR keyboard clients~\citep{chen2019federated},\\
	  {} & Reddit posts~\citep{mcmahan2018learning},\\
	  {} & Logs data, Cache data~\citep{hard2018federated}.\\
	  \midrule
	  Opinion polarity classification & MPQA~\citep{feyisetan2019leveraging}. \\
	  \midrule
      Paraphrase detection & MRPC~\citep{feyisetan2019leveraging}. \\
      \midrule
      Pos-tagging & TrustPilot, WebEng, AAVE~\citep{li2018towards}.\\
      \midrule
      Prediction adaptation & Amazon reviews, 20 news groups~\citep{clinchant2016transductive}.\\
      \midrule
      Privacy audit & Search logs~\citep{feyisetan2020privacy}.\\
      \bottomrule
      \multicolumn{2}{l}{\textit{Continues on next page...}} \\
      \end{tabular}\\
    \end{minipage}
    \end{center}
    \end{sidewaystable}

\begin{sidewaystable}
\sidewaystablefn%
\begin{center}
\begin{minipage}{\textheight}
\ContinuedFloat  
\caption{Summary of tasks and benchmark datasets for privacy-preserving NLP (Continued)}
\centering
\begin{tabular}{l|l}
    \toprule
    \textbf{NLP task} & \textbf{Datasets \& works} \\
    \toprule
      Privacy-aware text rewriting & Yelp, Facebook comments, DIAL~\citep{xu-etal-2019-privacy}. \\
      \midrule
	  Question answering & InsuranceQA~\citep{feyisetan2020privacy}.\\
	  \midrule
	  Question type classification & TREC-6~\citep{feyisetan2019leveraging}.\\
	  \midrule
	  Recognition of private information & CODE ALLTAG$_{S+d}$, CODE ALLTAG$_{XL}$~\citep{eder2020code}. \\
	  \midrule
	  Sentence intent classification & TREC~\citep{zhuetalal2020empiricalstudies}.\\
	  \midrule
	  Sentiment analysis & IMDB movie reviews~\citep{feyisetan2020privacy},\\
	  {} & Trustpilot~\citep{coavoux-etal-2018-privacy,mosallanezhad-etal-2019-deep},\\
	  {} & Trustpilot~\citep{lyu2020differentially,li2018towards}. \\ 
	  \midrule
	  Sentiment bias evaluation & Identity terms~\citep{sweeney2020reducing}. \\
	  \midrule
	  Sentiment prediction & Dialectal tweets (DIAL)~\citep{elazar-goldberg-2018-adversarial},\\
	  {} & MR, CR, SST-5~\citep{feyisetan2019leveraging}. \\
	  \midrule
	  Sentiment valence regression & SemEval-2018 Task 1~\citep{sweeney2020reducing}. \\
	  \midrule
	  Stylometrics & BookCorpus~\citep{song2020information}. \\
	  \midrule
	  Synthesized language generation & Twitter posts~\citep{oak2016generating}. \\
	  \midrule
	  Text style transfer & Yelp, Amazon reviews~\citep{john2019disentangled}.\\
	  \bottomrule
	  \multicolumn{2}{l}{\textit{Continues on next page...}} \\
      \end{tabular}\\
    \end{minipage}
    \end{center}
    \end{sidewaystable}

\begin{sidewaystable}
\sidewaystablefn%
\begin{center}
\begin{minipage}{\textheight}
\ContinuedFloat  
\caption{Summary of tasks and benchmark datasets for privacy-preserving NLP (Continued)}
\centering
\begin{tabular}{l|l}
    \toprule
    \textbf{NLP task} & \textbf{Datasets \& works} \\
    \toprule
	  Topic classification & AG news~\citep{coavoux-etal-2018-privacy,lyu2020differentially},\\
	  {} & Deutsche Welle~\citep{coavoux-etal-2018-privacy},\\
	  {} & Blog authorship~\citep{coavoux-etal-2018-privacy,lyu2020differentially}, \\
	  {} & Document set~\citep{fernandes2019generalised}.\\
	  \midrule
	  Toxicity classification & Wikipedia Talk~\cite{sweeney2020reducing}. \\
	  \midrule
	  Tweet-mention prediction & Dialectal tweets (DIAL)~\citep{elazar-goldberg-2018-adversarial},\\
	  {} & PAN16~\citep{elazar-goldberg-2018-adversarial,barrett-etal-2019-adversarial}. \\
	  \midrule
	  Word analogy & SemBias, Google Analogy, MSR~\citep{zhao2018learning,kaneko2019gender},\\
	  {} & SemEval~\citep{kaneko2019gender}. \\
	  \midrule
	  Word embedding model training & Wikipedia~\citep{song2020information,zhao2018learning}\\
	  {} & Wikipedia, Word lists~\citep{kaneko2019gender},\\
	  {} & Google News~\citep{bolukbasi2016man}, \\
	  {} & Mixed source sentences~\citep{font2019equalizing},\\
	  {} & Wikipedia, German Tweets, German Facebook~\citep{papakyriakopoulos2020bias}.\\
	  \midrule
	  Word extraction & Sentiment Lexicon~\citep{sweeney2020reducing}. \\
	  \midrule
	  Word prediction & Reddit, Wikitext-103~\citep{song2019auditing}. \\
	  \midrule
	  Word similarity & WS353, RG-65, MTurk, RW, MEN~\citep{zhao2018learning,kaneko2019gender},\\
      {} & SimLex~\citep{kaneko2019gender},\\ 
      {} & STS14~\citep{feyisetan2019leveraging}. \\
      \bottomrule
\end{tabular}
\end{minipage}
\end{center}
\end{sidewaystable}

Due to the growing literature on privacy-preserving NLP, we list the benchmark data used for the experiments in the reviewed works in Table~\ref{tab:summary-of-datasets}.
We also include NLP tasks in the table to point out the utility of the proposed privacy-preserving methods.

Many NLP tasks are performed over private data, as sentiment analysis~\citep{feyisetan2020privacy,lyu2020differentially,li2018towards}, AA~\citep{boumber2018experiments}, and neural machine translation~\citep{feng2020securenlp}, hence putting privacy at risk from disclosures or attacks.
PETs, such as FHE, DP, adversarial learning, and FL, are potential solutions for such privacy issues.
However, one should find a balance for the privacy-utility tradeoffs for each one of these technologies in a different manner.
So the performance requirements of an NLP task, alongside the computational power of the devices in the learning scenario, will play a significant role in choosing suitable PETs.
For instance, FHE may lead to memory overheads that are hard to manage for small devices like smartphones, slowing computations down~\citep{huang2020texthide}.
Therefore, a lighter encryption scheme or even controlled noise by DP would be preferred in such a scenario.

Although Table~\ref{tab:summary-of-datasets} presents a large number of entries, we can notice that the lack of data for privacy-preserving NLP tasks is still an open problem since the number of datasets per task is small for most of the tasks.
Firstly, this data availability issue is mainly related to the hardness of annotating sensitive text content~\citep{eder2020code} and human biases~\citep{zhao2020gender}.
Secondly, another hurdle to generating data for privacy purposes in NLP regards the safe release of PHI documents~\citep{melamud2019towards}, such as clinical notes and electronic health records, which is sometimes prohibited by legal terms.
Finally, there are further problems regarding the availability of datasets for languages other than English, yet a frequent situation across NLP tasks.

\section{Privacy metrics}\label{sec:privacy-metrics}

\begin{table}[h!]
\begin{center}
\begin{minipage}{\textwidth}
\caption{Summary of metrics for privacy in NLP}
\label{tab:measures}
\centering
\begin{tabular}{lll}
    \toprule
    \textbf{Metric} & \textbf{Privacy-related target} & \textbf{Work}\\
    \toprule
     $\epsilon$-DP & Privacy budget & \cite{fernandes2019generalised}\\
     {} & & \cite{feyisetan2020privacy}\\
     {} & & \cite{zhuetalal2020empiricalstudies}\\
     {} & & \cite{lyu2020differentially} \\
     inBias & Bias in multi-lingual embeddings & \cite{zhao2020gender} \\
     RNSB & Unintended demographic bias & \cite{sweeney2019transparent}\\
     $Exposure$ & Propensity for revealing data & \cite{carlini2019secret} \\
     Entropy & Sensitive information leakage & \cite{xu-etal-2019-privacy} \\
     P-Acc & Prediction of sensitive attributes & \cite{xu-etal-2019-privacy} \\
     M-Acc & Label probabilities for sentences & \cite{xu-etal-2019-privacy} \\
     FNED & Unintended biases & \cite{gencoglu2020cyberbullying} \\
     FPED & Unintended biases & \cite{gencoglu2020cyberbullying} \\
     S-PDTP & Prediction of private records & \cite{melamud2019towards} \\
\bottomrule
\end{tabular}
\end{minipage}
\end{center}
\end{table}

Privacy metrics aim at gauging the amount of privacy that users of a system experience and the extent of protection privacy-preserving methods provide~\citep{wagner2018technical}.
In the privacy-preserving NLP domain, many privacy metrics have been proposed in recent years.
Table~\ref{tab:measures} shows ten privacy metrics we have identified in the surveyed works.
The table presents metric names, alongside their privacy-related target that represents the privacy-related guarantee to be measured, as the privacy budget for DP or the amount of bias in word representations.
Therefore, we summarize these metrics as follows.
For detailed mathematical notation, we advise the reader to refer to the papers cited along column `Work'.

\begin{itemize}
    \item \textit{$\epsilon$-DP}. In DP~\citep{dwork2008differential}, the parameter $\epsilon$ is the upper bound to the probability of output to be changed by the addition or removal of a data instance~\citep{dwork2006calibrating,andrew2019tensorflow}. 
    This parameter is related to the privacy budget, which regards the amount of privacy to be protected.
    The closer the values of $\epsilon$ are to zero, the better the privacy protection.
    DP is broadly used for privacy-preserving NLP methods, such as FL~\citep{zhuetalal2020empiricalstudies}, authorship obfuscation~\citep{fernandes2019generalised}, and representation learning~\citep{feyisetan2020privacy,lyu2020differentially}.
    \item \textit{inBIAS}. 
    Gender bias is a broadly researched privacy issue in word and language representations. 
    Thus, metrics to quantify gender bias in embeddings have been proposed. 
    InBias~\citep{zhao2020gender} measures the intrinsic gender bias in multilingual word embeddings from a word-level perspective. 
    It is based on distances between gendered words, like occupations, and gendered seed words, like gender pronouns.
    For instance, if a feminine occupation presents a larger distance to the feminine seed gendered word when compared to the distance between its equivalent masculine words, it can be seen as a sign of bias. 
    Furthermore, this metric presents the advantage of enabling bias evaluation for embeddings of words in languages other than English.
    \item \textit{RNSB}. Bias towards demographic attributes (e.g., national origin and religion) can be gauged by this measure~\citep{sweeney2019transparent}. 
    It works by measuring the association of positive and negative sentiments and the protected demographic attributes in word embedding models.
    \item \textit{$Exposure$}. DL models may memorize training data, thereby putting data privacy at risk if an attacker tries to recover the original training samples. Therefore, the metric of $exposure$ measures the unintended memorization of unique or rare data instances~\citep{carlini2019secret}.
    It can be used during the model training to evaluate the risks of a successful attack.
    \item \textit{Entropy}. In the task of privacy-aware text rewriting, the generated text, which does not contain sensitive information, can be evaluated by the metric of entropy~\citep{xu-etal-2019-privacy}.
    Given a classifier that predicts the probability of sensitive attributes in the generated sentences in this task, higher entropy in the predictions is a sign of low risk of information leakage.
    \item \textit{P-Acc}. 
    This metric measures the ratio of correct predictions of the sensitive attribute obfuscated in the task of privacy-aware text rewriting~\citep{xu-etal-2019-privacy}. 
    Therefore, lower values of this metric suggest better obfuscation.
    \item \textit{M-Acc}.
    This is another metric to evaluate the privacy of a privacy-preserving text re-writing task.
    It compares the accuracies of predicting the labels for both original and generated sentences.
    Therefore, the approval for the generated sentence is based on the drop in the probability of the protected attributed after the re-writing task~\citep{xu-etal-2019-privacy}.
    \item \textit{FNED} and \textit{FPED}. These two metrics evaluate unintended biases towards demographic attributes, such as gender, religion, nationality, and language~\citep{gencoglu2020cyberbullying}. 
    These two measures are, based on the false negative and false positive ratios of classification tasks like cyberbullying detection.
    The total amount of unintended bias of a model is assumed to be the sum of both measures~\citep{gencoglu2020cyberbullying}. 
    \item \textit{S-PDTP}. 
    This metric measures the privacy protection for documents, like clinical notes, which should not be shared without PETs, e.g., DP or de-identification.
    S-PDTP~\citep{melamud2019towards} estimates the privacy risks for synthetic clinical notes generated from private ones.
    Therefore, the lower the scores of S-PDTP are, the less private information from the original clinical notes is predicted.
\end{itemize}

\section{Discussion and open challenges}\label{sec:discussion}

Protecting NLP data against privacy-related issues presents hurdles as to utility task goals, computation scenarios, DL architectures, and the properties of the datasets to be used.
For instance, the removal of protected terms from the vocabulary of a language model may disrupt the semantics of the remaining word representations and compromise the utility of such a privacy-preserving method.
In this section, we discuss the challenges of integrating privacy protection into NLP models regarding five aspects: traceability, computational overhead, dataset size, bias prevalence in embeddings, and privacy-utility tradeoff.
We also point out suggestions for future directions in privacy-preserving NLP.

\subsection{Traceability}\label{sub:traceability}

Obfuscating private text attributes is a tricky problem since these attributes may not be explicitly presented in the text.
Taking the problem of gender disclosure as an example, the gender of an author can be inferred from sentences, even though no explicit mention of their actual gender is made in the text.
Other private attributes can also be easily inferred from the text, such as location, age, native language, and nationality.
This issue of getting rid of private information is hardened by more complex text data, such as e-mails, documents, clinical notes, or electronic medical records, which are subject to data protection regulations, such as the HIPAA or the EU's GDPR.
Such regulations hinder personal or medical data sharing, or public release, without safeguarding methods, such as sanitization, de-identification, DP, or encryption.
However, when it comes to privacy guarantees, these methods present pitfalls whose solution still has room for new contributions, like diminishing the traceability of the original data.

Given an adversary that accesses a de-identified or anonymized document, it should not be able to trace the private document version.
First, data sanitization fails at bringing formal guarantees that traceability is unpractical for an adversary.
Second, de-identification replaces the private attributes with generic surrogate terms. 
However, in case these terms do not belong to the same semantic category as the original ones, the document context is disrupted.
Third, traceability is also a problem for representation learning since private attributes can be recovered from learned latent representations of words and documents, and implicit private information can be found on the gradients of DL models~\citep{qi-etal-2020-privacy}.
Further, searchable encryption is an efficient method against traceability since it forces adversaries to expend high computing power to beak ciphertexts, despite threats of file injection attacks.
Finally, DP introduces formal privacy guarantees against this issue~\citep{obeid2019impact}, enforcing the principle of indistinguishability between computation outputs.
We believe that more efforts toward DP will push the boundaries against data traceability in the future.

\subsection{Computation overhead}\label{sub:computational-overhead}

Privacy preservation often comes at the cost of computation overheads related to model run-time, memory footprint, and bandwidth consumption.
Infrastructure requirements represent a key point for the development of privacy-preserving models since the computation scenario is usually constrained.
For instance, some mobile devices may feature poor memory and processing power in a distributed setting.
So implementing memory-consuming FHE or MPC schemes may be prohibitive.
MPC overhead related to bandwidth absorption~\citep{feng2020securenlp} is a promising problem to work towards a solution in the future since distributed computation scenarios are increasing at the pace that mobile devices become increasingly popular.
Additionally, noise control for FHE on DL networks is another challenging issue.
FHE works by adding noise terms to ciphertexts, which will grow with mathematical operations of addition and multiplication.
When a DL model (e.g., LSTM, BERT, and ELMO) presents recurrence functions, the level of multiplicative depth can increase the level of noise, corrupting the model outputs once they are decrypted~\citep{al2020privft}.
Furthermore, MPC and FHE may slow down computations on massive databases.
DP noise also influences the model run-time but with the advantage of trading privacy protection with computation overhead.
So the amount of protected privacy can be accounted for as a budget.
Finally, an important open challenge for FL regards coming up with models that do not slow down the application run-time.
The importance of addressing this challenge is evinced by real-time mobile applications, like virtual smartphone keyboards, that should ideally provide instant results for users.
More than privacy protection, the solution for this problem influences the application's usability and rating by users.

\subsection{Dataset size}\label{sub:dataset-size}

Dataset size plays an important role in the decision for a PET in real-world NLP applications.
The importance of this factor is based on its influence on the generalization power of adversarial learning and FL and the privacy budget of DP.
For adversarial learning, the efficiency of the adversarial classifier that unlearns a private attribute can be negatively affected if the dataset is overly small.
Similarly, the dataset size can be a burden for FL since it may not be homogeneously distributed across the devices in the federated setting.
For instance, the storage size of distributed devices, such as smartphones or internet of things sensors, can present different storage sizes.
Therefore, an FL method must consider this point before designing the aggregation function on the global model and integrating additional PETs, like FHE or DP, into the learning setting.

In FL, sometimes the small number of distributed devices is also a burden to the model performance, e.g., data of a small number of users is insufficient to train accurate news recommendation models~\citep{qi-etal-2020-privacy}.
Although DP guarantees hold on smaller datasets, the level of noise injected into the data may occasionally vary.
Further, DP noise can be input into model gradients instead of data instances themselves~\citep{abadi2016deep}.
When it comes to combining DP with FL, it is a good practice to define the privacy budget locally based on each distributed device.
Establishing this parameter in the global model without considering the differences in data storage between the devices in the learning setting may lead to a scenario in which privacy is protected for devices with smaller amounts of data. 
In contrast, devices storing large amounts of data would not enjoy enough protection.
We believe many advances can still be made with respect to the influence of dataset sizes in privacy-preserving NLP, especially in backdrops with many devices featuring different data storage capacities.

\subsection{Bias prevalence in embeddings}\label{sub:bias-prevalence}

Bias and fairness are two critical privacy-related topics in NLP due to the ethical consequences they cause on automatic decision-making.
These topics thereby influence the daily lives of a large number of people since many real-word systems are based on word and sentence embedding models, such as word2vec, GloVe, fasttext, and BERT, which were found encoding and propagating discriminatory human biases towards demographic attributes, especially gender, race, nationality, and religion, on to downstream applications.
As an example, the machine translation task was found to output masculine words for non-gender-specific English words, like friend, when a feminine word was expected in the translated sentence~\citep{font2019equalizing}.
Reducing or removing human biases from embedding models mainly relies on adversarial training.
However, adversarial debiasing is not always able to remove biases completely~\citep{elazar-goldberg-2018-adversarial,barrett-etal-2019-adversarial}.
Therefore, bias prevalence in debiased embeddings is an open challenge.

Additionally, the evaluation step for debiasing and fairness solutions often does not include a human evaluation round.
We believe that inviting people for such an evaluation, especially those individuals from groups to which biases are often directed, would enhance the quality of the debiased embeddings and their successive applications.
Approaches for reducing discriminatory biases based on statistical dependency, namely causality, are also potential solutions for this problem~\citep{vig2020causal}.
This research direction may also explore the cases when biases arise from data sampling methods, such as sampling data instances mostly related to a specific sensitive attribute, whereas others are left out.
Finally, the gap between PETs and debiased embedding models must still be bridged.
PETs protect the data content during model computations, but the predictions on these protected datasets may still be biased.
Thus, there is an ever-increasing need for privacy-preserving fair NLP methods.

\subsection{Privacy-utility tradeoff}\label{sub:privacy-utility-tradeoff}

In the reviewed works, it is a recurrent statement that privacy preservation in NLP is exchanged for performance on NLP tasks.
Many PETs, like DP, AL, and data de-identification, reduce the model utility, whereas data privacy is protected.
For instance, the usability of medical text is frequently reduced by de-identification techniques~\citep{obeid2019impact}.
Similarly, the context of documents can be lost by replacing sensitive terms with surrogate words~\citep{eder2019identification}.
In embedding models, the problem of semantic disruption may be caused by DP noise, which is injected into the representations, aiming to obfuscate the text's private attributes.
Therefore, the final results of downstream applications can be compromised.

When a PET is implemented alongside the DL model for NLP, the model itself may face drawbacks.
For instance, searches on encrypted documents can be less efficient for the users than searches on the plaintext.
DP enables finding a good balance for the privacy-utility tradeoff~\citep{abadi2016deep,huang2020texthide} by reducing the accuracy losses but still preserving privacy to some extent.
FL is an effective privacy-preserving method, but as long as no data is exchanged between the distributed devices and the global model, the aggregation strategy to update the global model plays an important role in this tradeoff.
Some distributed devices in a federated setting may influence the global model to a larger degree based on their larger locally stored data.
Additionally, devices with poor bandwidth can be left out of model updates.
So the local performances can drop as well.
Therefore, balancing this tradeoff is certainly a key factor in designing privacy-preserving methods for NLP.

\section{Conclusion}\label{sec:conclusions}

This article presents the first extensive systematic review of DL methods for privacy-preserving NLP, covering a large number of tasks, PETs, privacy issues, and metrics for evaluating data privacy.
We propose a taxonomy structure to classify the existing works in this field as a guide to readers, with the advantage of enabling this structure to be extended to future works and regular machine learning methods.
Our review covers over sixty DL methods for privacy-preserving NLP published between 2016 and 2020, which safeguard privacy from the perspectives of data, models, computation scenarios, and PETs.
These methods explore the privacy-utility tradeoff to balance privacy protection and NLP task performance based on the assumption that privacy guarantees often come at a performance cost.

To sum up, we investigated the question of how to keep text private.
Firstly, we approached this endeavor by considering the data properties, such as type, size, and content.
Secondly, we describe the factors that influence choosing a suitable PET to be implemented with the DL model.
Thirdly, our extensive benchmark datasets and privacy measures tables can assist researchers and practitioners in successive works. 
Finally, the reviewed works show that PETs will play an ever-increasing role in future NLP applications since standard solutions may not hinder privacy threats when processing text data.

\section{Acknowledgments}

This work was supported by the EU’s Horizon 2020 project TRUSTS under grant agreement No. 871481.
Know‐Center is funded within the Austrian COMET Program – Competence Centers for Excellent Technologies – under the auspices of the Austrian Federal Ministry of Transport, Innovation and Technology, the Austrian Federal Ministry of Economy, Family and Youth, and the State of Styria. COMET is managed by the Austrian Research Promotion Agency (FFG).

\section*{Statements and declarations}

All authors certify that they have no affiliations with or involvement in any organization or entity with any financial interest or non-financial interest in the subject matter or materials discussed in this manuscript.

\bibliographystyle{unsrtnat}
\bibliography{sample-manuscript}

\end{document}